\documentclass{article}

\PassOptionsToPackage{numbers}{natbib}
\usepackage[preprint]{neurips_2025}

\usepackage[utf8]{inputenc} % allow utf-8 input
\usepackage[T1]{fontenc}    % use 8-bit T1 fonts
\usepackage{hyperref}       % hyperlinks
\usepackage{url}            % simple URL typesetting
\usepackage{booktabs}       % professional-quality tables
\usepackage{amsfonts}       % blackboard math symbols
\usepackage{nicefrac}       % compact symbols for 1/2, etc.
\usepackage{microtype}      % microtypography
\usepackage{xcolor}         % colors
\usepackage{amsmath}
\usepackage{enumitem}
\usepackage{graphicx}
\usepackage{caption}
\usepackage{arydshln}
\usepackage{wrapfig}
\usepackage{tikz}
\usepackage{subcaption}
\usepackage{enumitem}
\usepackage{wrapfig}
\usepackage{float} 
\usetikzlibrary{shapes.geometric, arrows.meta, positioning, calc, backgrounds}
\usepackage{amssymb}% http://ctan.org/pkg/amssymb
\usepackage{pifont}% http://ctan.org/pkg/pifont
\usepackage[most]{tcolorbox}

\newcommand{\cmark}{\ding{51}}%
\newcommand{\xmark}{\ding{55}}%
\usepackage{booktabs}
\usepackage[table]{xcolor}
\usepackage{colortbl}
\usepackage{arydshln}
\usepackage{siunitx}
\definecolor{wmHeaderBg}{HTML}{f5eef3}
\definecolor{wmGroupBg}{HTML}{fcf8fb}
\definecolor{wmGap}{HTML}{8B95A5}      % muted grey for gap subscripts
\definecolor{wmBest}{HTML}{662E7D}     % deep purple for 
\definecolor{wmSecond}{HTML}{6B7280}   % grey for runner-up
\definecolor{wmGoodCell}{HTML}{D9EAD3}
\definecolor{wmBadCell}{HTML}{F4CCCC}

\definecolor{cVAE}{HTML}{CC6677}
\definecolor{cVAVAE}{HTML}{882255}
\definecolor{cCOSMOS}{HTML}{DDCC77}
\definecolor{cVJEPA2}{HTML}{332288}
\definecolor{cWEBSSL}{HTML}{117733}
\definecolor{cSIGLIP}{HTML}{88CCEE}
\definecolor{cQWEN}{HTML}{AA4499}
\definecolor{cDiT}{RGB}{0,123,255} % Placeholder color for cDiT

\newcommand{\best}[1]{\textbf{\textcolor{wmBest}{#1}}}
\newcommand{\second}[1]{\underline{\textcolor{wmSecond}{#1}}}
\newcommand{\gap}[1]{{\scriptsize\textcolor{wmGap}{(\,#1\,)}}}

\newcommand{\famtag}[2]{{\scriptsize\textcolor{c#1}{#2}}}
\newcommand{\rowhead}[1]{\textsc{#1}}
\newcommand{\uncertainty}[1]{{\scriptsize\textcolor{wmGap}{#1}}}

\newtcolorbox{recipebox}{
  colback=wmGroupBg,
  colframe=wmGroupBg,
  boxrule=0pt,
  arc=2mm,
  left=2mm,
  right=2mm,
  top=1mm,
  bottom=1mm,
  breakable
}

\newtcolorbox{wmPromptBox}[2]{
  breakable,
  enhanced,
  colback=white,
  colframe=wmBest,
  coltitle=white,
  boxrule=0.8pt,
  arc=1.5mm,
  left=1.5mm,
  right=1.5mm,
  top=1mm,
  bottom=1mm,
  title={#1},
  fonttitle=\bfseries,
  before upper={
    \textbf{Use:} #2\par\medskip
  },
}

\DeclareMathOperator*{\argmin}{arg\,min}
\title{Reconstruction or Semantics? What Makes a Latent Space Useful for Robotic World Models
}
\def\huggingface{\raisebox{-1.5pt}{\includegraphics[height=1.05em]{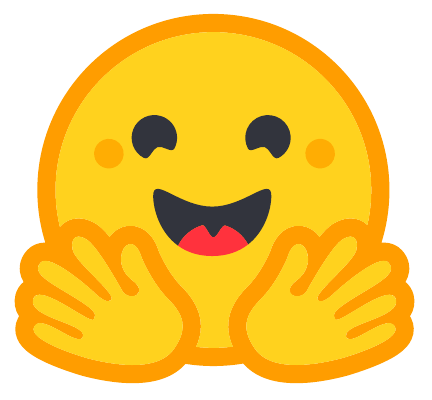}}}
\def\github{\raisebox{-1.5pt}{\includegraphics[height=1.05em]{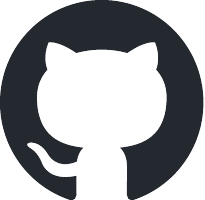}}}

\author{
  Nilaksh$^{\textbf{*}1,2,3}$ \:\:
  Saurav Jha$^{\textbf{*}1,2,3}$ \:\:
  Artem Zholus$^{\textbf{*}1,2,3}$ \:\:
  Sarath Chandar$^{1,2,3,4}$ \\
  $^{1}$Chandar Research Lab \quad 
  $^{2}$Mila -- Quebec AI Institute \quad 
  $^{3}$Polytechnique Montréal \quad 
  $^{4}$Canada CIFAR AI Chair \\
  \vspace{1.5em}
  \scriptsize $^*$Equal Contribution \\
  \scriptsize Correspondence: \texttt{[nilaksh.nilaksh, saurav.jha]@mila.quebec} \\
  \vspace{3mm}
  {\github \: \footnotesize \url{https://hskalin.github.io/semantic-wm/}}\\
  {\huggingface \: \footnotesize \url{https://huggingface.co/Nilaksh404/semantic-wm}}
}

\begin{document}

\maketitle
\begin{abstract}
World model-based policy evaluation is a practical proxy for testing real-world robot control by rolling out candidate actions in action-conditioned video diffusion models. As these models increasingly adopt latent diffusion modeling (LDM), choosing the \textit{right latent space} becomes critical. While the status quo uses autoencoding latent spaces like VAEs that are primarily trained for pixel \textit{reconstruction}, recent work suggests benefits from pretrained  encoders with representation-aligned \textit{semantic} latent spaces. We systematically evaluate these latent spaces for action-conditioned LDM by comparing six reconstruction and semantic encoders to train world model variants under a fixed protocol on BridgeV2 dataset, and show effective world model training in high-dimensional representation spaces with and without dimension compression. We then propose three axes to assess robotic world model performance: visual fidelity, planning and downstream policy performance, and latent representation quality. Our results show visual fidelity alone is insufficient for world model selection. While reconstruction encoders like VAE and Cosmos achieve strong pixel-level scores,  semantic encoders such as V-JEPA 2.1 (strongest overall on policy), Web-DINO, and SigLIP 2 generally excel across the other two axes at all model scales. Our study advocates semantic latent space as stronger foundation for policy-relevant robotics diffusion world models.
\end{abstract}

\section{Introduction}
Action-conditioned video world models are emerging as a practical interface between generative modeling and robotics \citep{ha2018world, yang2023learning, brooks2024video}. Given observation and action histories, they predict future observations and serve as learned proxies for robot-environment interaction when handcrafted simulators are difficult to build \citep{todorov2012mujoco, erez2015simulation}. Recent works show that such models can support policy evaluation with good correlation to real-world outcomes \citep{tseng2025scalable}, and policy improvement \citep{zhu2025wmpo, zhang2025reinforcing, sharma2026world}. Yet current evaluations say little about which representation makes a world model faithful to robotic dynamics.

% Central paper figure: action-conditioned latent diffusion WM + encoder ablation panel.
%
% Required preamble in your paper (add once to your main .tex):
%   \usepackage{tikz}
%   \usetikzlibrary{shapes.geometric, arrows.meta, positioning, calc, backgrounds}
%   \input{theme/latex_macros.tex}   % defines cVAE, cVAVAE, ..., cQWEN
%
% Then use:
%   \input{figures/out/architecture_overview.tex}
%
% This file is self-contained: it re-defines accent colors at the top so it works even
% without the theme macros (it falls back to grey if the cFAM colors are missing).
%
% NOTE: The top panel includes \includegraphics calls for frame_0_t0.png ...
% frame_3_t33.png. Make sure those files are on the LaTeX graphics path (or
% comment out the four \draw nodes that include them if you don't need the
% rendered frame thumbnails).

% Accent colors. \definecolor silently overwrites if already defined in your preamble,
% so re-running these is harmless.
\definecolor{wmAccentBrown}{HTML}{8C5A3C}
\definecolor{wmBlueAccent}{HTML}{1F4E79}
\definecolor{wmEvalGrey}{HTML}{6F7480}
% Encoder family colors (warm_cool palette; will be overwritten if you load
% theme/latex_macros.tex which defines them too).
\definecolor{cVAE}{HTML}{D08770}
\definecolor{cVAVAE}{HTML}{BF616A}
\definecolor{cCOSMOS}{HTML}{EBCB8B}
\definecolor{cVJEPA2}{HTML}{5E81AC}
\definecolor{cWEBSSL}{HTML}{88C0A0}
\definecolor{cSIGLIP}{HTML}{81A1C1}
\definecolor{cQWEN}{HTML}{B48EAD}

\begin{figure*}[t!]
\centering
\resizebox{\textwidth}{!}{%
\begin{tikzpicture}[
  font=\footnotesize,
  >={Stealth[length=1.8mm, width=1.6mm]},
  node distance=4mm and 5mm,
  every node/.append style={align=center, line width=0.4pt},
]

% ---- Block styles (used by bottom panel) ---------------------------------
\tikzset{
  card/.style={
    rounded corners=2.5pt, draw=#1, fill=white, line width=0.8pt,
    minimum width=25mm, minimum height=13mm, inner sep=0pt,
  },
}

% ============================================================================
% TOP PANEL — pipeline (imported from second TikZ figure)
% Wrapped in a scope so its absolute coordinates / yscale=-1 don't affect the
% bottom panel layout.
% ============================================================================
\begin{scope}[local bounding box=toppanel,
              x=0.75pt, y=0.75pt, yscale=-0.55, xscale=0.75,
              line width=0.75pt]

%uncomment if require: \path (0,427); %set diagram left start at 0, and has height of 427

%Rounded Rect [id:dp19717433913292381] 
\draw  [draw opacity=0][fill={rgb, 255:red, 70; green, 119; blue, 222 }  ,fill opacity=0.1 ] (399.5,107.48) .. controls (399.5,104.73) and (401.73,102.5) .. (404.48,102.5) -- (616.52,102.5) .. controls (619.27,102.5) and (621.5,104.73) .. (621.5,107.48) -- (621.5,248.52) .. controls (621.5,251.27) and (619.27,253.5) .. (616.52,253.5) -- (404.48,253.5) .. controls (401.73,253.5) and (399.5,251.27) .. (399.5,248.52) -- cycle ;
%Shape: Trapezoid [id:dp4627835917763471] 
\draw  [draw opacity=0][fill={rgb, 255:red, 206; green, 179; blue, 211 }  ,fill opacity=0.1 ] (196.18,43.5) -- (387.01,107.56) -- (387.01,197.44) -- (196.18,261.5) -- cycle ;
%Rounded Rect [id:dp9447519005786312] 
\draw  [color={rgb, 255:red, 102; green, 46; blue, 125 }  ,draw opacity=1 ][fill=white  ,fill opacity=1 ] (140.5,52.25) .. controls (140.5,51.92) and (140.77,51.65) .. (141.1,51.65) -- (158.15,51.65) .. controls (158.48,51.65) and (158.75,51.92) .. (158.75,52.25) -- (158.75,249.23) .. controls (158.75,249.56) and (158.48,249.83) .. (158.15,249.83) -- (141.1,249.83) .. controls (140.77,249.83) and (140.5,249.56) .. (140.5,249.23) -- cycle ;
%Shape: Trapezoid [id:dp15760065435463266] 
\draw  [color={rgb, 255:red, 102; green, 46; blue, 125 }  ,draw opacity=1 ][fill={rgb, 255:red, 237; green, 222; blue, 238 }  ,fill opacity=1 ][dash pattern={on 4.5pt off 4.5pt}] (201.8,54.17) -- (285.99,79.43) -- (285.99,225.57) -- (201.8,250.83) -- cycle ;
%Shape: Trapezoid [id:dp17835248673921644] 
\draw   (310.44,92.28) -- (382.78,114.73) -- (382.78,155.05) -- (310.44,177.5) -- cycle ;
%Straight Lines [id:da4222912865247961] 
\draw [color={rgb, 255:red, 102; green, 46; blue, 125 }  ,draw opacity=1 ]   (165,149.74) -- (191.94,149.74) ;
\draw [shift={(194.94,149.74)}, rotate = 180] [fill={rgb, 255:red, 102; green, 46; blue, 125 }  ,fill opacity=1 ][line width=0.08]  [draw opacity=0] (8.04,-3.86) -- (0,0) -- (8.04,3.86) -- (5.34,0) -- cycle    ;
%Straight Lines [id:da5054830863123779] 
\draw [color={rgb, 255:red, 102; green, 46; blue, 125 }  ,draw opacity=1 ]   (291.6,134.5) -- (300.5,134.5) -- (304.19,134.5) ;
\draw [shift={(307.19,134.5)}, rotate = 180] [fill={rgb, 255:red, 102; green, 46; blue, 125 }  ,fill opacity=1 ][line width=0.08]  [draw opacity=0] (8.04,-3.86) -- (0,0) -- (8.04,3.86) -- (5.34,0) -- cycle    ;
%Rounded Rect [id:dp9768883990758472] 
\draw  [color={rgb, 255:red, 39; green, 91; blue, 130 }  ,draw opacity=1 ] (412.96,113) .. controls (412.96,111.42) and (414.24,110.14) .. (415.82,110.14) -- (567.25,110.14) .. controls (568.83,110.14) and (570.11,111.42) .. (570.11,113) -- (570.11,194.17) .. controls (570.11,195.75) and (568.83,197.03) .. (567.25,197.03) -- (415.82,197.03) .. controls (414.24,197.03) and (412.96,195.75) .. (412.96,194.17) -- cycle ;
%Rounded Rect [id:dp7465103167848897] 
\draw  [color={rgb, 255:red, 39; green, 91; blue, 130 }  ,draw opacity=1 ] (595.54,108.27) .. controls (595.54,107.91) and (595.84,107.61) .. (596.2,107.61) -- (614.84,107.61) .. controls (615.21,107.61) and (615.5,107.91) .. (615.5,108.27) -- (615.5,247.21) .. controls (615.5,247.57) and (615.21,247.86) .. (614.84,247.86) -- (596.2,247.86) .. controls (595.84,247.86) and (595.54,247.57) .. (595.54,247.21) -- cycle ;
%Straight Lines [id:da576805913174678] 
\draw [color={rgb, 255:red, 102; green, 46; blue, 125 }  ,draw opacity=1 ]   (388.51,135.92) -- (404.22,135.92) ;
\draw [shift={(407.22,135.92)}, rotate = 180] [fill={rgb, 255:red, 102; green, 46; blue, 125 }  ,fill opacity=1 ][line width=0.08]  [draw opacity=0] (8.04,-3.86) -- (0,0) -- (8.04,3.86) -- (5.34,0) -- cycle    ;
%Rounded Rect [id:dp7298148497259694] 
\draw  [color={rgb, 255:red, 39; green, 91; blue, 130 }  ,draw opacity=1 ] (427.5,122.46) .. controls (427.5,121.93) and (427.93,121.5) .. (428.46,121.5) -- (455.54,121.5) .. controls (456.07,121.5) and (456.5,121.93) .. (456.5,122.46) -- (456.5,186.54) .. controls (456.5,187.07) and (456.07,187.5) .. (455.54,187.5) -- (428.46,187.5) .. controls (427.93,187.5) and (427.5,187.07) .. (427.5,186.54) -- cycle ;
%Rounded Rect [id:dp9015162532436996] 
\draw  [color={rgb, 255:red, 39; green, 91; blue, 130 }  ,draw opacity=1 ] (469.5,121.46) .. controls (469.5,120.93) and (469.93,120.5) .. (470.46,120.5) -- (497.54,120.5) .. controls (498.07,120.5) and (498.5,120.93) .. (498.5,121.46) -- (498.5,187.54) .. controls (498.5,188.07) and (498.07,188.5) .. (497.54,188.5) -- (470.46,188.5) .. controls (469.93,188.5) and (469.5,188.07) .. (469.5,187.54) -- cycle ;
%Straight Lines [id:da23087223124507572] 
\draw [color={rgb, 255:red, 102; green, 46; blue, 125 }  ,draw opacity=1 ]   (530.5,218.92) -- (589,218.92) ;
\draw [shift={(592,218.92)}, rotate = 180] [fill={rgb, 255:red, 102; green, 46; blue, 125 }  ,fill opacity=1 ][line width=0.08]  [draw opacity=0] (8.04,-3.86) -- (0,0) -- (8.04,3.86) -- (5.34,0) -- cycle    ;
%Straight Lines [id:da920167167542513] 
\draw [color={rgb, 255:red, 102; green, 46; blue, 125 }  ,draw opacity=1 ]   (295.5,218.92) -- (517.5,218.92) ;
%Straight Lines [id:da13185553792598637] 
\draw [color={rgb, 255:red, 39; green, 91; blue, 130 }  ,draw opacity=1 ]   (574.51,157.92) -- (590.22,157.92) ;
\draw [shift={(593.22,157.92)}, rotate = 180] [fill={rgb, 255:red, 39; green, 91; blue, 130 }  ,fill opacity=1 ][line width=0.08]  [draw opacity=0] (8.04,-3.86) -- (0,0) -- (8.04,3.86) -- (5.34,0) -- cycle    ;
%Straight Lines [id:da17252438506446044] 
\draw  [dash pattern={on 4.5pt off 4.5pt}]  (524.5,258.5) -- (524.5,203.5) ;
\draw [shift={(524.5,200.5)}, rotate = 90] [fill={rgb, 255:red, 0; green, 0; blue, 0 }  ][line width=0.08]  [draw opacity=0] (10.72,-5.15) -- (0,0) -- (10.72,5.15) -- (7.12,0) -- cycle    ;
%Rounded Rect [id:dp9161375688952148] 
\draw  [color={rgb, 255:red, 102; green, 46; blue, 125 }  ,draw opacity=1 ] (71.5,52.25) .. controls (71.5,51.92) and (71.77,51.65) .. (72.1,51.65) -- (89.15,51.65) .. controls (89.48,51.65) and (89.75,51.92) .. (89.75,52.25) -- (89.75,249.23) .. controls (89.75,249.56) and (89.48,249.83) .. (89.15,249.83) -- (72.1,249.83) .. controls (71.77,249.83) and (71.5,249.56) .. (71.5,249.23) -- cycle ;
%Rounded Rect [id:dp9706890425296356] 
\draw  [color={rgb, 255:red, 102; green, 46; blue, 125 }  ,draw opacity=1 ] (101.5,52.25) .. controls (101.5,51.92) and (101.77,51.65) .. (102.1,51.65) -- (119.15,51.65) .. controls (119.48,51.65) and (119.75,51.92) .. (119.75,52.25) -- (119.75,249.23) .. controls (119.75,249.56) and (119.48,249.83) .. (119.15,249.83) -- (102.1,249.83) .. controls (101.77,249.83) and (101.5,249.56) .. (101.5,249.23) -- cycle ;
%Shape: Trapezoid [id:dp18859902702423614] 
\draw  [color={rgb, 255:red, 102; green, 46; blue, 125 }  ,draw opacity=1 ][fill={rgb, 255:red, 237; green, 222; blue, 238 }  ,fill opacity=1 ] (716.5,252.83) -- (647.8,232.22) -- (647.8,76.78) -- (716.5,56.17) -- cycle ;
%Straight Lines [id:da9861218936378091] 
\draw [color={rgb, 255:red, 102; green, 46; blue, 125 }  ,draw opacity=1 ]   (624.51,158.92) -- (640.22,158.92) ;
\draw [shift={(643.22,158.92)}, rotate = 180] [fill={rgb, 255:red, 102; green, 46; blue, 125 }  ,fill opacity=1 ][line width=0.08]  [draw opacity=0] (8.04,-3.86) -- (0,0) -- (8.04,3.86) -- (5.34,0) -- cycle    ;
%Rounded Rect [id:dp12157417049398822] 
\draw  [color={rgb, 255:red, 102; green, 46; blue, 125 }  ,draw opacity=1 ] (743.5,54.25) .. controls (743.5,53.92) and (743.77,53.65) .. (744.1,53.65) -- (761.15,53.65) .. controls (761.48,53.65) and (761.75,53.92) .. (761.75,54.25) -- (761.75,251.23) .. controls (761.75,251.56) and (761.48,251.83) .. (761.15,251.83) -- (744.1,251.83) .. controls (743.77,251.83) and (743.5,251.56) .. (743.5,251.23) -- cycle ;
%Straight Lines [id:da3566244278380646] 
\draw [color={rgb, 255:red, 102; green, 46; blue, 125 }  ,draw opacity=1 ]   (721.51,152.92) -- (737.22,152.92) ;
\draw [shift={(740.22,152.92)}, rotate = 180] [fill={rgb, 255:red, 102; green, 46; blue, 125 }  ,fill opacity=1 ][line width=0.08]  [draw opacity=0] (8.04,-3.86) -- (0,0) -- (8.04,3.86) -- (5.34,0) -- cycle    ;
%Image [id:dp507257645648219] 
\draw (840,263) node  {\includegraphics[width=35pt,height=35pt]{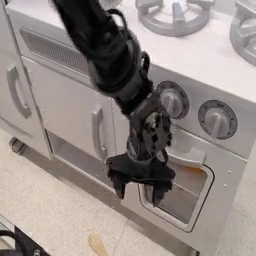}};
%Image [id:dp9839310464171072] 
\draw (840,175) node  {\includegraphics[width=35pt,height=35pt]{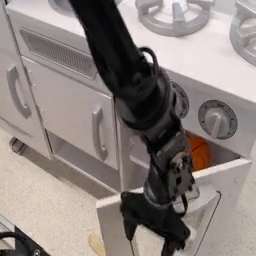}};
%Image [id:dp5091390263461476]
\draw (840,87) node  {\includegraphics[width=35pt,height=35pt]{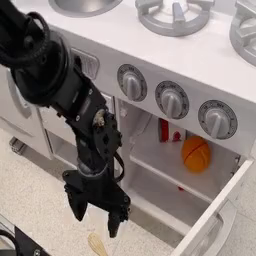}};

% Text Node
\draw (442,154.5) node  [font=\scriptsize,rotate=-270]  {$\text{spatial}$};
% Text Node
\draw (484,154.5) node  [font=\scriptsize,rotate=-270]  {$\text{temporal}$};
% Text Node
\draw (502,146.4) node [anchor=north west][inner sep=0.75pt]    {$\times$ Blocks};
% Text Node
\draw (440,79.4) node [anchor=north west][inner sep=0.75pt]    {$\text{\textbf{DiT}} \ \text{\textit{latent diffusion}}$};
% Text Node
\draw (400,262.4) node [anchor=north west][inner sep=0.75pt]    {$\text{action} \ a_{0:t} +\text{\textcolor{wmBlueAccent}{\textit{(optional) text}}}$};
% Text Node
\draw (605.52,177.74) node  [rotate=-270]  {$\text{wide head}$};
% Text Node
\draw (120,145.4) node [anchor=north west][inner sep=0.75pt]  [color={rgb, 255:red, 102; green, 46; blue, 125 }  ,opacity=1 ]  {$\cdots $};
% Text Node
\draw (73.5,264.63) node [anchor=north west][inner sep=0.75pt]    {$\text{\textit{history}}$};
% Text Node
% \draw (130,264.4) node [anchor=north west][inner sep=0.75pt]    {$\text{\textit{future}}$};
% Text Node
\draw (243.89,152.5) node    {$ \begin{array}{l}
\text{\textbf{Encoder}} \ f_\phi \\
\ \ \ \text{frozen}
\end{array}$};
% Text Node
\draw (346.61,134.89) node    {$ \begin{array}{l}
\text{adapter}\\
D \rightarrow d
\end{array}$};
% Text Node
\draw (682.15,154.5) node    {$\text{decoder}$};
% Text Node
\draw (754.75,279.03) node [anchor=south] [inner sep=0.75pt]    {$\text{\textit{predicted}}$};
\end{scope}

% ============================================================================
% Bridge header (between panels)
% ============================================================================
\node[below=0mm of toppanel.south, font=\small] (zooheader)
  {$\Phi \in \{\,$ candidate latent representations $\,\}$};

% ============================================================================
% BOTTOM PANEL — encoder zoo (6 cards, Qwen-VL removed)
% ============================================================================
\def\cardSep{5mm}

\foreach \i/\fam/\backbone/\dim/\colour/\adapter in {%
  0/VAE/SD3 VAE/16/cVAE/,
  1/VA-VAE/VA-VAE/32/cVAVAE/,
  2/Cosmos/Cosmos CI encoder/16/cCOSMOS/,
  3/V-JEPA 2/V-JEPA 2.1 ViT-L/1024/cVJEPA2/{$+$ adp.\ $d{=}96$},
  4/Web-SSL/Web-DINO ViT-L/1024/cWEBSSL/{$+$ adp.\ $d{=}96$},
  5/SigLIP 2/SigLIP 2 ViT-L/1152/cSIGLIP/{$+$ adp.\ $d{=}96$}%
}{%
  \pgfmathsetmacro{\cardX}{(\i-2.5)*27}%
  \node[card=\colour, anchor=north]
       at ([xshift=\cardX mm, yshift=-1mm] zooheader.south) (card\i) {};
  % Top coloured stripe with family name (drawn on background so card border stays clean)
  \begin{scope}[on background layer]
    \node[fill=\colour, draw=none, rounded corners=1pt,
          minimum width=23mm, minimum height=4mm,
          anchor=north, inner sep=0pt, yshift=-1mm] at (card\i.north) {};
  \end{scope}
  \node[anchor=north, yshift=-1.6mm, font=\bfseries\scriptsize,
        text=white] at (card\i.north) {\fam};
  % Body text — packed tight, no big vertical gap
  \node[font=\scriptsize, anchor=north, yshift=-2mm,
        text width=24mm, align=center, inner sep=1pt] at (card\i.north)
       {\backbone\\[1pt]native $D\!=\!\dim$\\[2pt]\textit{\adapter}};
}

% Group captions (positioned under the geometric centre of each group)
\node[font=\scriptsize\itshape, color=wmAccentBrown,
      below=2.5mm of card1.south]
     {VAE-like (frozen encoder, no adapter)};
\node[font=\scriptsize\itshape, color=wmBlueAccent,
      below=2.5mm of $(card3.south)!0.5!(card5.south)$]
     {Representation encoders (frozen $+$ optional learned adapter)};

\end{tikzpicture}}
\caption{%
   \textbf{Which latent space makes a better robotic world model?}\,
  For a latent diffusion model, we fix the Diffusion Transformer (DiT) transition model, action conditioning, and training data. We vary only the encoder $f_\phi$ defined latent interface: encoder, optional compression adapter, and the associated decoder path. This isolates how reconstruction-aligned and semantic representations affect action-faithful dynamics, generated rollouts, and downstream policy performance for robot control. We show the encoder families compared in the bottom panels.%
}
\vspace{-1.8em}
\label{fig:pipeline-overview}
\end{figure*}
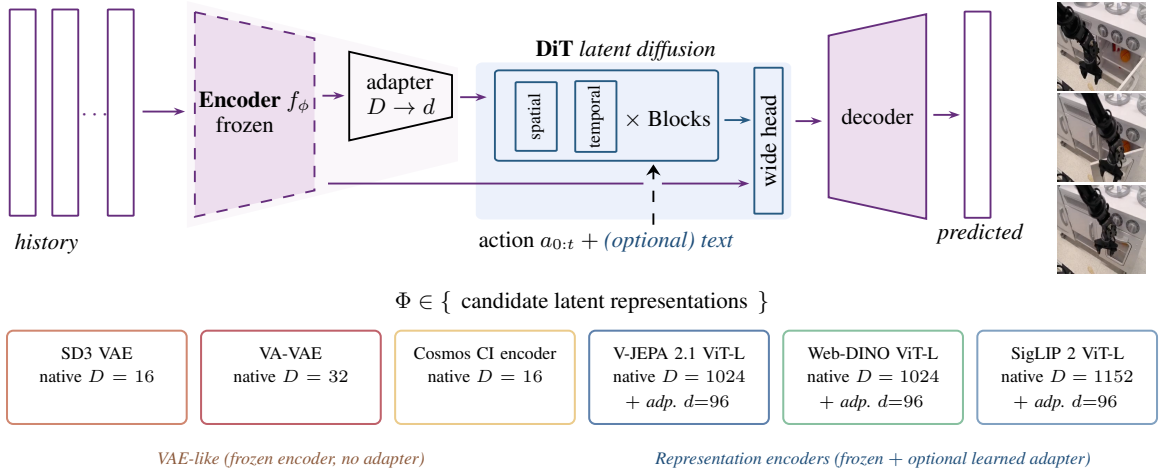

This question is increasingly important because many video world models are latent diffusion models (LDMs) \citep{vahdat2021score, rombach2022high} that learn dynamics in an encoder-defined latent space. The standard choice is a reconstruction-aligned autoencoder, such as a VAE \citep{kingma2013auto} or recent variants \citep{esser2024scaling, yao2025vavae, agarwal2025cosmos}, whose latents are optimized for pixel fidelity and stable decoding. But robotic world models are more than video generators, where planning and evaluations require predictions that preserve physical, spatial, and task dynamics. This motivates using the semantic spaces of self-supervised and vision-language encoders as latents for robot world modeling \citep{caron2021emerging, oquab2023dinov2, he2020momentum, he2022masked, midovjepa2, radford2021learning, tschannen2025siglip}. These spaces expose object layout and task structure more directly than pixel-trained autoencoders \citep{shi2026latent}. However, they are hard to use for diffusion due to their higher dimensionality yielding off-manifold latent generation with poor object structures \citep{zhang2025both}. RAE \cite{zheng2025diffusion} makes them more tractable with a dimension-dependent noise-schedule shift and a wide DDT head \citep{wang2025ddt}, while S-VAE  \citep{zhang2025both} learns a compact, KL-regularized latent space using an autoencoder as an adapter over the frozen semantic features.

Still, the effect of semantic latents on action-conditioned LDM for robotics remains open. DINO-WM \citep{zhouDINOWMWorldModels2025} and V-JEPA 2-AC \citep{midovjepa2} show that pretrained feature spaces support planning, but they are not diffusion models: DINO-WM is an autoregressive feature-prediction world model, while V-JEPA 2-AC is a JEPA predictor \citep{assran2023self}. RAE-NWM \citep{zhang2026rae} shows that DINOv2 \citep{oquab2023dinov2} spaces  support diffusion-based navigation world modeling. Yet navigation differs from contact-rich manipulation, where gripper motion, object state, geometry, and policy rollouts all matter. This leads to our question: \textbf{what effects does latent space choice have for LDM-based robotic world modeling?}

We answer this with a controlled evaluation study that varies only the representation space in which the transition model operates (see Fig.~\ref{fig:pipeline-overview}). For effective semantic space LDM training, we adapt  RAE's wide-head and schedule-shift recipe \citep{zheng2025diffusion} alongside the compact S-VAE adapter \citep{zhang2025both}, and train on the Bridge V2 dataset \citep{pmlr-v229-walke23a} with the same DiT transition model \citep{peebles2023scalable} and action-conditioning scheme. We then propose an evaluation suite spanning three axes: visual fidelity, planning and downstream policy performance, and latent quality. Our findings show that semantic latents improve action recoverability, task-success classification, CEM planning, and policy-in-the-loop success, while reconstruction latents mainly retain photometric advantages. Our key contributions are three-fold:

\begin{enumerate}[leftmargin=*]
    \item Our primary contribution is the \textit{evaluation} of representation spaces for latent diffusion world modeling. We do controlled analyses of how latent space choice affects not only visual generation, but also robotics tasks and robustness through our proposed three evaluation axes.

    \item We propose an effective recipe for \textit{training diffusion world models in high dimensional semantic spaces}, by leveraging the recent advances in semantic space diffusion and extending them to action-conditioned world modeling. We also study the effects of different design choices.

    \item We show that semantic latent spaces are consistently more useful for policy evaluation and planning, even when reconstruction latents match or exceed them on low-level pixel fidelity, establishing that the best robotic world model latent space is the one that preserves action-relevant structure, not merely the one that reconstructs images the best.
\end{enumerate}

\section{Problem Formulation}\label{sec:formulation}

We consider multi-task robot manipulation from partial observations. The offline dataset is
$\mathcal{D}=\{(o_{0:T},\,a_{0:T-1},\,\ell,\,y)\}$, where $o_t \in \mathcal{O}$ is an RGB observation, $a_t \in \mathbb{R}^{d_a}$ is a continuous robot action, $\ell$ is an optional language instruction, and $y \in \{0,1\}$ denotes episode success. Tasks vary in object configurations and instructions, but share a robot embodiment; we therefore view the data as samples from related partially observed Markov Decision Processes with shared dynamics and task-dependent goals. Because a single observation does not generally determine the next observation under an action, we condition on a finite visual-action history of length $H$ and model the action-conditioned predictive distribution over a rollout horizon $K$: $p(o_{t+1:t+K} \mid o_{t-H:t},\, a_{t-H:t+K-1})$.

\begin{wrapfigure}{r}{0.31\textwidth}
  \centering
  \vspace{-2.3em}
\includegraphics[width=0.3\textwidth]{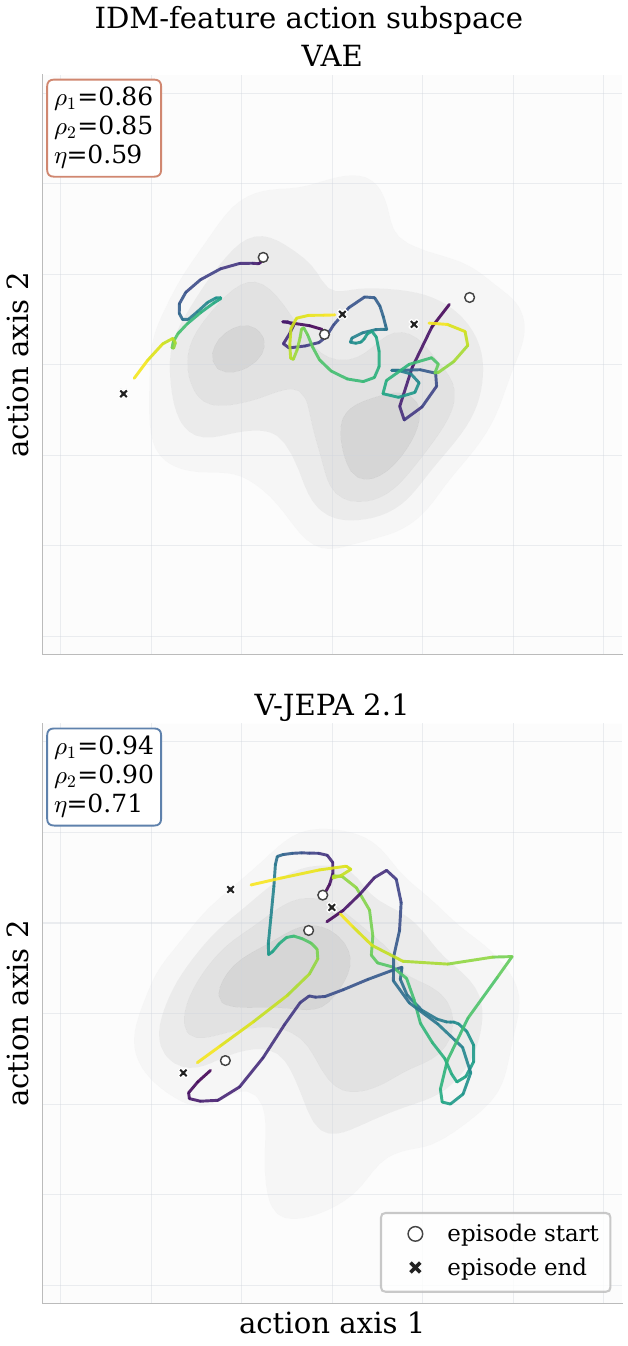}
  \caption{\textbf{Action trajectories induced by encoder spaces:} episode rollouts projected onto the top-2 canonical-correlation directions between IDM features and ground-truth actions. $(\rho_1, \rho_2)$ are the leading canonical correlations, $\eta$ summarizes the aggregate action alignment. Colored curves are episodes.}
  \label{fig:action-traj}
  \vspace{-5.1em}
\end{wrapfigure}
\subsection{Latent Space World Models}\label{sec:latent_wm}

Rather than predicting future frames directly in pixel space, latent world models learn predictive dynamics in a representation space. Each model consists of a frozen encoder, an optional frozen adapter, an action-conditioned transition model, and a decoder.

\paragraph{Encoder and adapter.}
A pretrained image encoder maps each observation to a spatial latent $z_t=f_\phi(o_t)\in\mathbb{R}^{N\times D}$, where $N=h\times w$ is the number of patches and $D$ is the encoder's native channel dimension. The encoder is frozen, so $f_\phi$ fixes the representation space in which dynamics are learned. For high-dimensional semantic representation encoders, we optionally    use a frozen adapter $\alpha_\psi$ to obtain compact diffusion-friendly latents $\tilde{z}_t=\alpha_\psi(z_t)\in\mathbb{R}^{N\times d}$ \citep{zhang2025both}. For compressed reconstruction-aligned latent spaces, the adapter is  the identity map.

\paragraph{Transition model.}
  An action-conditioned DiT \citep{peebles2023scalable} predicts future latent trajectories:
  $\tilde{z}_{t+1:t+K}\sim p_\theta(\cdot\mid \tilde{z}_{t-H:t},a_{t-H:t+K-1})$.
  Only the transition model is updated during world model training; the encoder, adapter, and decoder remain fixed. For semantic encoders without adapters,
  we add a lightweight wide DDT  head  ~\citep{wang2025ddt}, which adds few parameters but addresses the width bottleneck of DiT for high-dimensional latent spaces \citep{zheng2025diffusion}. Otherwise, variants share  the same transition backbone and differ only in representation and decoding path. Table \ref{tab:arch-param-compute} (Appx. \ref{app:archi-and-training}) shows that the DiT backbone with adapter does not incur an increase in parameter count or GFLOPs. Compute parity is explained in Appx. \ref{app:faq}.

\paragraph{Decoder.}
Predicted latents are mapped back to pixels as $\hat{o}_{t+1:t+K}=\mathrm{Dec}(\tilde{z}_{t+1:t+K})$. The decoder is needed for visual rollouts and pixel-level evaluation, but decoded image quality alone does not determine world model quality: a model may render plausible frames while missing action-relevant dynamics, or preserve control-relevant structure despite minor photometric errors.

\subsection{The Role of the Latent Space in Robotics}\label{sec:latent_spaces}
The encoder-defined latent space determines the state representation on which the transition model $p_\theta$ learns dynamics. In LDM, reconstruction-aligned latents $z_t^{\mathrm{pix}}=f_\phi^{\textsc{pix}}(o_t)\in\mathbb{R}^{N\times D_{\mathrm{pix}}}$ are commonly used because they preserve pixel-level information and provide reliable decoders \citep{child2021very}. For robotic world models, however, the relevant state is not only what an image looks like, but how it changes under actions and whether those changes preserve task progress, object state, contact, and geometry. This creates a multi-objective problem where useful latents should be action-controllable, task-informative, visually decodable, and useful for planning or policy evaluation.

As an initial diagnostic, we use  inverse dynamics model (IDM) to probe whether an encoder makes action-relevant change explicit in latent space (see Appx. \ref{app:latent-rep} for details). Figure~\ref{fig:action-traj} shows that different encoders induce markedly different action-aligned trajectory geometries, suggesting that encoder choice changes which aspects of robot dynamics are easy for a transition model to learn. This motivates us to treat the latent space $f_\phi$ as the experimental variable, and evaluate its effect beyond visual fidelity and on axes spanning controllability, task semantics, and policy performance.

We thus compare reconstruction-aligned latents with semantic latents from pretrained vision foundation models \citep{oquab2023dinov2, midovjepa2, tschannen2025siglip}, denoted as $z_t^{\mathrm{rep}}=f_\phi^{\textsc{rep}}(o_t)\in\mathbb{R}^{N\times D_{\mathrm{rep}}}$. Since $D_{\mathrm{rep}}$ is typically large, we evaluate both native features and compact adapter latents $\tilde{z}_t=\alpha_\psi(z_t^{\mathrm{rep}})$. We train one world model per candidate in $\Phi=\{f_\phi^{(1)},\ldots,f_\phi^{(m)}\}$ while fixing the data, history, action conditioning, optimizer, and transition backbone, so that each model learns a different latent transition
$p_{\theta}^{(\phi)}(\tilde{z}_{t+1:t+K}\mid \tilde{z}_{t-H:t},a_{t-H:t+K-1})$. The decoder differences are controlled through reconstruction gap metrics, latent-space metrics, and planning metrics.

\section{Experiments}

\subsection{Dataset and Training}

\paragraph{Benchmark protocol.}
We isolate the effect of the encoder-defined latent space by fixing the dataset, history length, action conditioning, transition architecture, optimizer, and training schedule, and varying only the encoder $f_\phi$, optional adapter $\alpha_\psi$, and decoder path. For each encoder--adapter pair, we train an LDM from scratch and evaluate the resulting world model for visual fidelity, representation quality, and downstream policy performance (see Appx. \ref{app:archi-and-training}).

\paragraph{Dataset.}
We train and evaluate on Bridge~V2~\citep{pmlr-v229-walke23a}, a real-robot manipulation dataset with $\approx$60K WidowX~250 demonstrations across 13 task families. Each episode includes RGB observations, 7 Degrees of Freedom (DoF) end-effector actions covering position, rotation, and gripper state, and a language instruction.  For trajectory success classification, we use SOAR~\citep{zhou2024autonomous} which contains roughly 30.5K success/failure class episodes for WidowX~250 with a 1:2 class split.

\paragraph{Encoder variants.}
We compare two encoder families. reconstruction-aligned encoders $f_\phi^{\textsc{pix}}$ include: Stable Diffusion 3 (SD3) VAE~\citep{esser2024scaling}  with $D{=}16$, VA-VAE~\citep{yao2025vavae} with $D{=}32$, and Cosmos~\citep{agarwal2025cosmos} with $D{=}16$; for these, $\alpha_\psi \equiv \mathbb{I}$. Semantics-aligned encoders $f_\phi^{\textsc{rep}}$ include: V-JEPA\,2.1~\citep{MurLabadia2026VJEPA2U} with $D{=}1024$, Web-DINO~\citep{fan2025scaling}, adapted from DINOv2~\citep{oquab2023dinov2}, with $D{=}1024$, and SigLIP\,2~\citep{tschannen2025siglip} with $D{=}1152$. For semantic encoders, we evaluate both native latents and compact latents from a pretrained S-VAE adapter~\citep{zhang2025both}, which maps $D{\to}d$ with $d{=}96$.

\paragraph{Adapter, decoder, and transition model.}
The S-VAE adapter~\citep{zhang2025both} is pretrained to reconstruct frozen encoder features with a KL-regularized  loss, and is paired with a lightweight pixel decoder. All transition models are DiTs trained on Bridge~V2 \citep{pmlr-v229-walke23a}  with flow matching~\citep{lipman2023flow}. Each DiT layer factorizes attention into a spatial block within each frame and a causal temporal block across frames. We sample every second frame, condition on $H{=}2$ history frames, and predict 8 future frames. We do not make use of language instruction conditioning while training the DiT. For all non-VAE encoders, we apply a dimension-dependent noise-schedule shift ~\citep{esser2024scaling}. At inference, models roll out autoregressively one frame at a time using a 10-frame sliding context; VAE variants use their native pixel decoders, while semantic variants use the learned adapter decoder (see Appx. \ref{app:archi-and-training} for details).

\subsection{Evaluation Metrics} 
\label{sec:eval-metrics}

To study how the choice of latent representation propagates through to downstream tasks, we propose an evaluation suite that segregates this effect across three axes. See Appx. \ref{app:eval_metrics} for details.

\begin{enumerate}[label=\textbf{\arabic*.}, leftmargin=*, align=left, labelsep=0.5em]

\item \textbf{Planning and downstream policy performance.} For robotics applications, a latent world model should enable planning, \textit{i.e.},\ searching for the optimal action sequence given a goal state \citep{zhouDINOWMWorldModels2025, midovjepa2}. Evaluating planning helps separate the latent world modeling performance from the pixel decoder performance, which visual metrics conflate together. Given a real $k$-step transition, we use the cross-entropy method (CEM)~\citep{rubinstein2004cross} to recover the action sequence whose predicted latent best matches the target, and report CEM error at single-step  $(k=1)$  and multi-step $(k=4)$ horizons.

We also test whether the world model can serve as a policy-evaluation environment. We roll out OpenVLA-7B~\citep{kim2025openvla} inside each world model on 20 Bridge~V2 test episodes with 8 trials per episode, and a subset of 10 of these were used for Out-Of-Distribution (OOD) evaluations. We use two Vision-Language Models (VLMs): InternVL~3.5 \citep{wang2025internvl3} and Qwen~3.6 \citep{qwen3.6-27b}, to judge the tasks' success. We report consensus success rate, Borda rank, and robustness under distractor-object and OOD-instruction perturbations. See Appx. \ref{app:eval_metrics} for metrics definitions, Appx. \ref{app:faq} regarding fairness of VLM ratings, and Appx. \ref{app:vla-eval-ood} \& \ref{app:vlm-judge} for exact details about OOD frame and OOD instruction generation, as well as details about tasks.

\item \textbf{Pixel fidelity and scene geometry.}
Decoded rollouts must remain visually coherent  to support visual policies. We report image/video metrics: FID, SSIM, LPIPS, FVD, temporal LPIPS, and point-track consistency, together with perceptual and geometric scores from WorldArena~\citep{shang2026worldarena}.
This family measures generation and motion quality, temporal consistency,  and scene geometry. 

\item \textbf{Latent representation quality.}
Because the transition model operates in latent space, we directly probe whether generated latents preserve action and task-relevant structure. We train an inverse dynamics model (IDM) \citep{tian2025predictive} on frozen encoder latents to recover action chunks for horizon $k{\in}\{1,4\}$, and apply the IDM to world model latents to measure generation-induced degradation. We  train a classifier on latent trajectories of SOAR \citep{zhou2024autonomous}, a language and success label annotated dataset of trajectories, to classify whether a trajectory was a success given the text instruction. We again measure the degradation in accuracy induced by evaluating on generated latents. 

\end{enumerate}
\begin{figure}[t]
  \centering
  % First figure
  \begin{subfigure}[b]{0.32\textwidth}
    \centering
    \includegraphics[width=\textwidth]{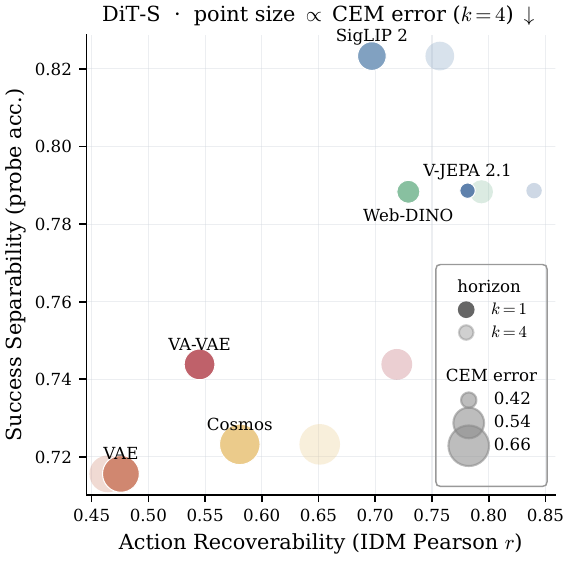}
        \vspace{-1em}
    \caption{Latent space utility}
        \label{fig:dits-idm-metrics}
  \end{subfigure}
  \hfill
  % Second figure
  \begin{subfigure}[b]{0.32\textwidth}
    \centering
    \includegraphics[width=\textwidth]{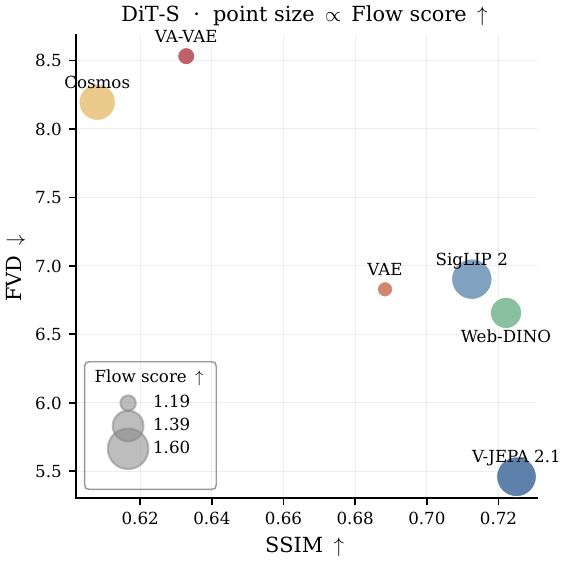}
        \vspace{-1em}
    \caption{Visual utility}
        \label{fig:dits-vis-metrics}
  \end{subfigure}
  \hfill
  % Second figure
  \begin{subfigure}[b]{0.32\textwidth}
    \centering
    \includegraphics[width=\textwidth]{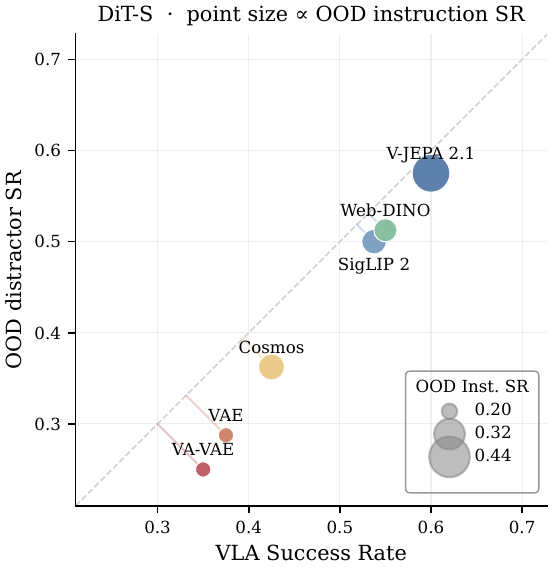}
    \vspace{-1em}
    \caption{Policy performance}
        \label{fig:dits-policy-metrics}
  \end{subfigure}
  \caption{\textbf{Latent space effect overview:}
each point is a DiT-S world model trained by varying only the encoder and the associated decoder path. 
(a)  \best{Upper-right is favorable.} Latent space metrics show that semantic encoders improve action recoverability, task-success separability, and action planning error (CEM) relative to reconstruction-aligned encoders. 
(b)  \best{Lower-right  is favorable.} Visual utility metrics show that pixel fidelity alone does not explain downstream performance: reconstruction-aligned spaces remain competitive on low-level image quality, while semantic spaces often improve video and motion quality. 
(c)  \best{Upper-right  is favorable.} Closed-loop evaluations show that semantic spaces generally yield higher VLA success and stronger robustness to OOD objects and instructions. Details about all metrics are in Sec. \ref{sec:eval-metrics} and Appx. \ref{app:eval_metrics}.}
  \label{fig:main_findings}
\end{figure}

\section{Findings}
\subsection{Does the choice of latent space affect planning and policy performance?}
\label{sec:policy-perf}

\paragraph{Semantic latents offer better policy-in-the-loop performance.}
Table~\ref{tab:policy-perf} shows that encoder choice strongly affects downstream VLA policy rollouts at DiT-S. Reconstruction-aligned spaces perform worst: VAE and VA-VAE have the lowest consensus success rates and weakest Borda ranks, while semantic encoders improve policy success, interaction quality, and robustness. V-JEPA~2.1 and SigLIP~2 variants give the strongest DiT-S results. Semantic-family VLA SR and CEM outperform reconstruction-family under paired bootstrap over tasks as shown in our analysis in Appx. \ref{app:stat-analysis}.

\begin{table}[t]
\centering
\vspace{0em}
\caption{\textbf{DiT-S policy and behavioral metrics.} \best{Best} and \second{runner-up} per column. In-distribution (ID) SR and Out-of-Distribution (OOD) SR are calculated on a subset of 10 episodes with InternVL 3.5. Consenus SR and Borda rank  aggregate InternVL3.5-14B and Qwen3.6-27B rankings.  Interaction quality measures the plausibility of robot-object contact. PCK coverage measures point tracking recall (Appx. \ref{app:eval_metrics}). Muted $\pm$ terms show one standard deviation error averaged over episodes.}
\label{tab:policy}
\resizebox{\textwidth}{!}{%
\begin{tabular}{l|cc|cc|c|ccc|cc}
\toprule
  & \multicolumn{2}{c|}{VLA SR} & \multicolumn{2}{c|}{Interaction quality} & \multicolumn{1}{c|}{PCK} & \multicolumn{3}{c|}{OOD robustness} & \multicolumn{2}{c}{CEM error} \\
\cmidrule(lr){2-3}\cmidrule(lr){4-5}\cmidrule(lr){6-6}\cmidrule(lr){7-9}\cmidrule(lr){10-11}
Encoder & \shortstack{Consensus\\SR} & \shortstack{Borda\\rank} & \shortstack{IQ\\score} & \shortstack{Instr.\\follow} & \shortstack{PCK\\coverage} & \shortstack{ID\\SR} & \shortstack{OOD SR\\distractor} & \shortstack{OOD SR\\instruction} & k=1 & k=4 \\
  & $\uparrow$ & $\downarrow$ & $\uparrow$ & $\uparrow$ & $\uparrow$ & $\uparrow$ & $\uparrow$ & $\uparrow$ & $\downarrow$ & $\downarrow$ \\
\midrule
\famtag{VAE}{$\bullet$}~VAE & 0.169~\uncertainty{$\pm$\,0.030} & 25 & 3.26 & 3.48 & 0.719 & 0.375~\uncertainty{$\pm$\,0.054} & 0.287~\uncertainty{$\pm$\,0.051} & 0.200~\uncertainty{$\pm$\,0.045} & 0.111~\uncertainty{$\pm$\,0.009} & 0.612~\uncertainty{$\pm$\,0.023} \\
\rowcolor{wmGroupBg}\famtag{VAVAE}{$\bullet$}~VA-VAE & 0.175~\uncertainty{$\pm$\,0.030} & 23 & 3.22 & 3.42 & 0.715 & 0.350~\uncertainty{$\pm$\,0.053} & 0.250~\uncertainty{$\pm$\,0.048} & 0.200~\uncertainty{$\pm$\,0.045} & 0.097~\uncertainty{$\pm$\,0.005} & 0.543~\uncertainty{$\pm$\,0.023} \\
\famtag{COSMOS}{$\bullet$}~Cosmos & 0.244~\uncertainty{$\pm$\,0.034} & 16 & 3.32 & 3.51 & 0.707 & 0.425~\uncertainty{$\pm$\,0.055} & 0.362~\uncertainty{$\pm$\,0.054} & 0.275~\uncertainty{$\pm$\,0.050} & 0.112~\uncertainty{$\pm$\,0.009} & 0.661~\uncertainty{$\pm$\,0.033} \\
\cmidrule{1-11}
\rowcolor{wmGroupBg}\famtag{VJEPA2}{$\bullet$}~V-JEPA 2.1 & \second{0.344}~\uncertainty{$\pm$\,0.038} & \best{6} & 3.43 & \second{3.78} & \best{0.735} & \second{0.600}~\uncertainty{$\pm$\,0.055} & \second{0.575}~\uncertainty{$\pm$\,0.055} & \best{0.400}~\uncertainty{$\pm$\,0.055} & \second{0.084}~\uncertainty{$\pm$\,0.008} & \best{0.424}~\uncertainty{$\pm$\,0.014} \\
\famtag{VJEPA2}{$\bullet$}~V-JEPA 2.1$_{96}$ & \best{0.362}~\uncertainty{$\pm$\,0.038} & \second{8} & \best{3.52} & \best{3.84} & 0.735 & 0.600~\uncertainty{$\pm$\,0.055} & 0.537~\uncertainty{$\pm$\,0.056} & 0.250~\uncertainty{$\pm$\,0.048} & 0.089~\uncertainty{$\pm$\,0.007} & 0.548~\uncertainty{$\pm$\,0.017} \\
\rowcolor{wmGroupBg}\famtag{WEBSSL}{$\bullet$}~Web-DINO & 0.212~\uncertainty{$\pm$\,0.032} & 21 & 3.34 & 3.58 & \second{0.735} & 0.550~\uncertainty{$\pm$\,0.056} & 0.512~\uncertainty{$\pm$\,0.056} & 0.250~\uncertainty{$\pm$\,0.048} & 0.090~\uncertainty{$\pm$\,0.007} & \second{0.474}~\uncertainty{$\pm$\,0.026} \\
\famtag{WEBSSL}{$\bullet$}~Web-DINO$_{96}$ & 0.300~\uncertainty{$\pm$\,0.036} & 11 & \second{3.44} & 3.77 & 0.732 & 0.600~\uncertainty{$\pm$\,0.055} & 0.512~\uncertainty{$\pm$\,0.056} & 0.275~\uncertainty{$\pm$\,0.050} & 0.090~\uncertainty{$\pm$\,0.007} & 0.531~\uncertainty{$\pm$\,0.025} \\
\rowcolor{wmGroupBg}\famtag{SIGLIP}{$\bullet$}~SigLIP 2 & 0.325~\uncertainty{$\pm$\,0.037} & 9 & 3.43 & 3.58 & 0.730 & 0.537~\uncertainty{$\pm$\,0.056} & 0.500~\uncertainty{$\pm$\,0.056} & 0.263~\uncertainty{$\pm$\,0.049} & \best{0.082}~\uncertainty{$\pm$\,0.006} & 0.523~\uncertainty{$\pm$\,0.030} \\
\famtag{SIGLIP}{$\bullet$}~SigLIP 2$_{96}$ & 0.331~\uncertainty{$\pm$\,0.037} & 15 & 3.42 & 3.71 & 0.731 & \best{0.625}~\uncertainty{$\pm$\,0.054} & \best{0.588}~\uncertainty{$\pm$\,0.055} & \second{0.312}~\uncertainty{$\pm$\,0.052} & 0.086~\uncertainty{$\pm$\,0.005} & 0.537~\uncertainty{$\pm$\,0.026} \\
\bottomrule
\end{tabular}
}
\label{tab:policy-perf}
\vspace{0em}
\end{table}

\paragraph{Native semantic spaces preserve action geometry for planning.}
 Representation aligned spaces have the lowest action-recovery errors across all DiT backbone sizes (Table \ref{tab:policy-perf}, and Table \ref{tab:policy-full} in Appx \ref{app:additional-res}). For example, at DiT-S V-JEPA~2.1 is best at $k{=}4$ and SigLIP~2 is best at $k{=}1$. Fig.~\ref{fig:dits-policy-metrics} likewise shows semantic encoders closer to the upper-right diagonal in the VLA--OOD plane, while VAE-family models fall lower and suffer larger distractor-induced drops.

\paragraph{Scaling narrows policy gaps but not action-centric gaps.}
Appx. Table~\ref{tab:policy-full} shows that For DiT-L, the gaps in VLA success and OOD robustness for  VAE and Cosmos narrow relative to semantic encoders. We attribute this to improved visual fidelity at larger model size, which benefits the VLA policy. However, both still lag on CEM action recovery, which depends directly on latent transition structure rather than rendered visual quality; at DiT-L, VAE and Cosmos have larger $k{=}1$ CEM errors than all semantic encoders. They also lag on IDM $r$ and classifier accuracy (Table \ref{tab:idm-pearson-k1-k4} and \ref{tab:success-probe-full}).

\subsection{Does the latent space affect action recoverability and preservation of task semantics?}
\begin{wraptable}{r}{0.5\textwidth}
\centering
\vspace{-1em}
\caption{\textbf{IDM Pearson} $r$ (horizons $k\in\{1,4\}$) and Success classifier for DiT-S, reported on encoder (Enc.) and  world model (WM) latents. 
% \best{Best} and \second{runner-up} results are highlighted per column.
}
\resizebox{\linewidth}{!}{%
\begin{tabular}{l|cc|cc|cc}
\toprule
  & \multicolumn{4}{c|}{Pearson $r$} & \multicolumn{2}{c}{Classifier Acc.} \\
\cmidrule(lr){2-5}\cmidrule(lr){6-7}
Encoder & \multicolumn{2}{c|}{Enc.$\uparrow$} & \multicolumn{2}{c|}{WM$\uparrow$} & \multicolumn{2}{c}{Whole-video} \\
\cmidrule(lr){2-3}\cmidrule(lr){4-5}\cmidrule(lr){6-7}
  & $k{=}1$ & $k{=}4$ & $k{=}1$ & $k{=}4$ & Enc.$\uparrow$ & WM$\uparrow$ \\
\midrule
\famtag{VAE}{$\bullet$}~VAE & 0.507 & 0.478 & 0.476 & 0.464 & 0.835 & 0.716 \\
\rowcolor{wmGroupBg}\famtag{VAVAE}{$\bullet$}~VA-VAE & 0.549 & 0.744 & 0.545 & 0.719 & 0.868 & 0.744 \\
\famtag{COSMOS}{$\bullet$}~Cosmos & 0.626 & 0.673 & 0.581 & 0.651 & 0.851 & 0.723 \\
\cmidrule{1-7}
\rowcolor{wmGroupBg}\famtag{VJEPA2}{$\bullet$}~V-JEPA 2.1 & \best{0.829} & \best{0.865} & \best{0.781} & \best{0.840} & \second{0.905} & \second{0.789} \\
\famtag{WEBSSL}{$\bullet$}~Web-DINO & \second{0.820} & \second{0.845} & \second{0.729} & \second{0.794} & \best{0.906} & 0.788 \\
\rowcolor{wmGroupBg}\famtag{SIGLIP}{$\bullet$}~SigLIP 2 & 0.772 & 0.793 & 0.697 & 0.757 & 0.903 & \best{0.823} \\
\bottomrule
\end{tabular}
}
\label{tab:idm-suc-dits}
\vspace{-1.5em}
\end{wraptable}

\paragraph{Semantic latents make action-relevant changes more recoverable.}
Table~\ref{tab:idm-suc-dits} shows that semantic encoders retain substantially more action information than reconstruction-aligned ones. On encoder latents, V-JEPA~2.1 and Web-DINO achieve the strongest IDM Pearson $r$ across both horizons, and this advantage largely persists after world model (WM) generation.  The trends also hold with DiT scaling (Tables \ref{tab:idm-pearson-k1-k4} and \ref{tab:success-probe-full} in Appx. \ref{app:latent-rep}). %These results suggest that semantic latent spaces make action-conditioned state changes easier for the transition model to represent and preserve.

\paragraph{Semantic latents better preserve task-success information.}
From Table~\ref{tab:idm-suc-dits}, we also see that success classifiers trained on frozen encoder latents achieve higher accuracy for semantic encoders, and their performance degrades less when evaluated on generated WM latents, with SigLIP 2 having best WM latent accuracy.  
This indicates that semantic spaces not only encode local action effects, but also retain higher-level task progress signals useful for policy evaluation.

\subsection{How does the latent space affect visual fidelity?}

\begin{wrapfigure}{l}{0.32\textwidth}
  \centering
  \vspace{-1.5em}
  \includegraphics[width=0.32\textwidth]{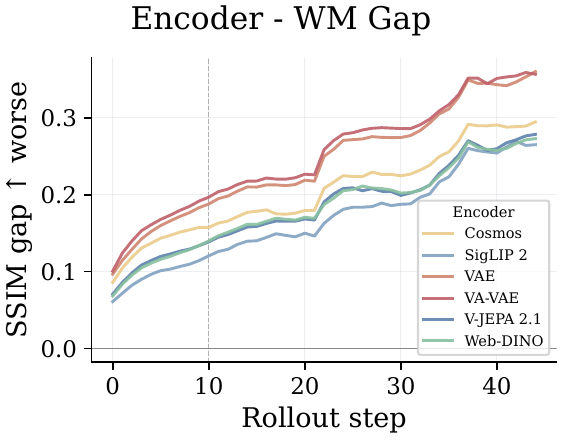}
  \caption{\textbf{SSIM gap} over steps.}
  \vspace{-3em}
  \label{fig:rollout-gap-trend}
\end{wrapfigure}

\paragraph{Semantic latent spaces remain visually competitive.}
Table~\ref{tab:visual-quality} shows that the policy gains from semantic encoders do not come at the cost of decoded visual quality. At DiT-S scale, these encoders dominate most perceptual, structural, and video-level metrics, particularly when used with adapters $d_{96}$: SigLIP~2$_{96}$ gives the best SSIM, V-JEPA~2.1$_{96}$ gives the best FVD, and Web-DINO variants are strongest on JEPA similarity, subject consistency, depth error, and temporal LPIPS. 

VAE-style spaces remain competitive on image quality, and qualitatively tend to preserve sharper local appearance details, but they lag behind semantic spaces on global structure and temporal generation quality. Figures \ref{fig:rollout-gap-trend} and \ref{fig:rollout-gap-full} (Appx. \ref{app:additional-res})  show  semantic space models have lower gap for pixel reconstruction, particularly while extrapolating beyond the 10-frame horizon length seen during training.

\paragraph{Large DiTs help recover much of the visual advantage of reconstruction latents.}
Increasing transition model capacity benefits reconstruction latents the most. For DiT-L, VAE becomes highly competitive, achieving the best FID, image quality, aesthetic quality, JEPA similarity, depth error, dynamic degree, and FVD, while also ranking second on LPIPS and flow score. Here, semantic encoders still remain strong: V-JEPA~2.1$_{96}$ gives the best SSIM and LPIPS, and SigLIP~2$_{96}$ remains competitive on structure and temporal metrics, but their gains from scaling are less uniform. Overall, visual fidelity alone does not explain the downstream policy advantages observed in Sec.~\ref{sec:policy-perf}.

\subsection{Does scaling along input views and model size help?}
\begin{wrapfigure}{l}{0.62\textwidth}
  \centering
  \vspace{-1.1em}
  \begin{minipage}[t]{0.31\textwidth}
    \centering
    \includegraphics[width=\linewidth]{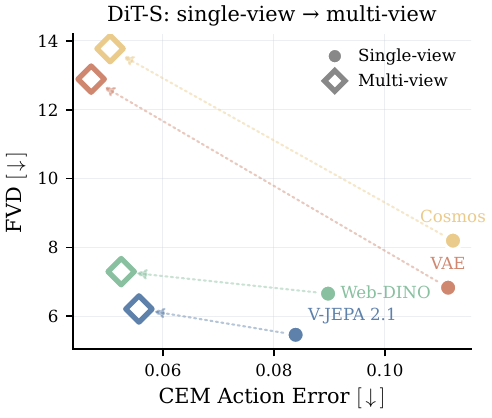}
  \end{minipage}%
  \hfill
  \begin{minipage}[t]{0.31\textwidth}
    \centering
    \includegraphics[width=\linewidth]{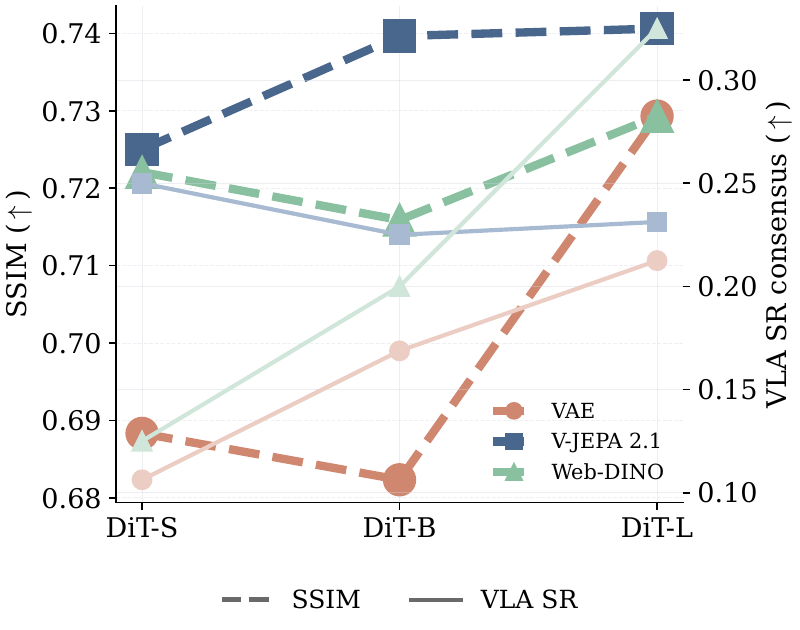}
  \end{minipage}
  \caption{\textbf{Scaling} camera views (left) and DiT sizes (right).}
  \label{fig:scaling}
  \vspace{-1.05em}
\end{wrapfigure}

\paragraph{Multi-view training improves action recovery but can hurt video quality under limited data.}
We take the trained DiT-S models and finetune them for 20 epochs on the BridgeV2 episodes that contain three camera views. Fig.~\ref{fig:scaling} (left) shows that while this does lead to superior CEM action prediction, it also degrades generation quality, possibly due to smaller number of training episodes. However, the semantic encoders are more robust to this degradation. 
\textbf{Model scaling improves both visual quality and policy success, with larger gains for reconstruction latents:}
in Fig.~\ref{fig:scaling} (right), we see that both generation (SSIM) and policy performance (VLA-SR) generally scale with the DiT size. Here, VAE scales notably well on visual metrics and approaches semantic encoders, which already perform strongly at DiT-S.

\begin{wrapfigure}{r}{0.4\textwidth}
  \centering
  \vspace{-2em}  \includegraphics[width=\linewidth]{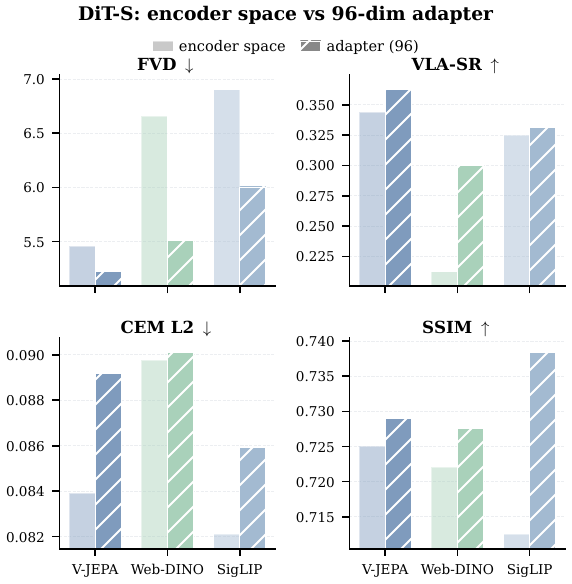}
  \caption{\textbf{Adapter} ablation results.}
  \label{fig:adapt-vs-noadapt}
  \vspace{0em}
\end{wrapfigure}

\subsection{Do reconstruction-aligned and semantic encoders fail differently?}

\begin{figure}[t]
\vspace{-1em}
    \centering
\includegraphics[width=\textwidth]{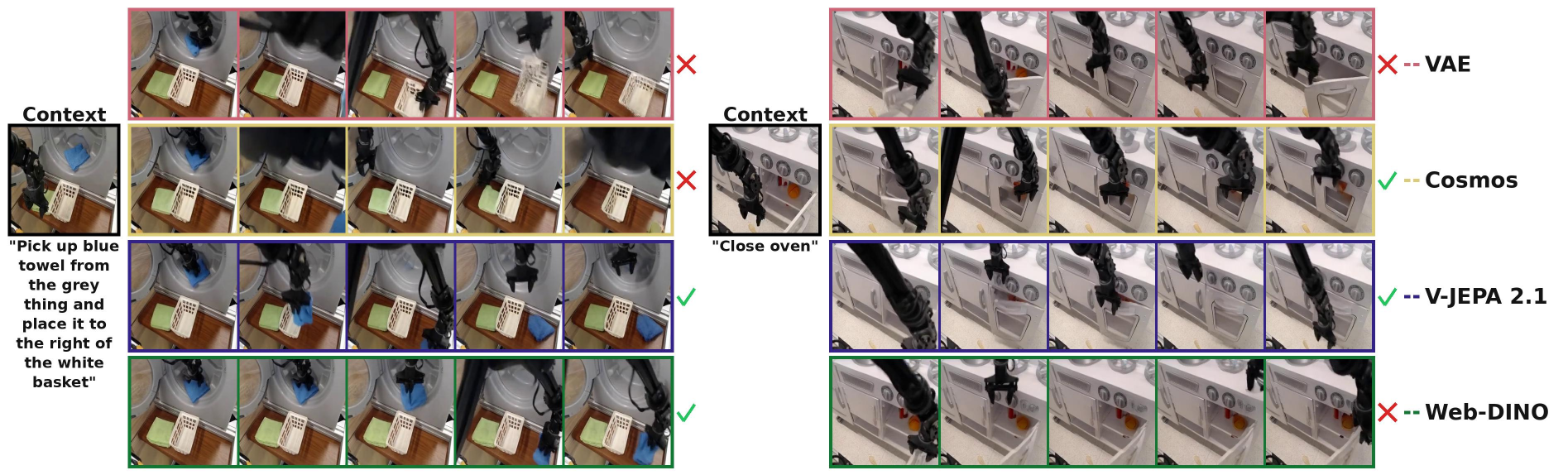}
    \caption{\textbf{Open-VLA success rate comparison on two random episodes:} four frames are sampled at even intervals. \textcolor{green}{\cmark} and \textcolor{red}{\xmark}\;
 show trajectories marked as success and failure by InternVL 3.5 VLM.}
    \label{fig:success-rate-comp}
    \vspace{-1.4em}
\end{figure}

\paragraph{The main failure modes differ: reconstruction latents hallucinate task semantics, while semantic latents miss geometry and contact.}
Our qualitative rollouts in Appx. fig. \ref{fig:hallucinated-rollout-pixels2} show that all encoder families share a common failure mode where  static scene elements are faithfully preserved while manipulation-relevant details  hallucinate. 
Beyond this universal pattern, encoder families show distinct hallucinations. Reconstruction encoders tend to fail at the object-semantic level: VAE and Cosmos hallucinate the white basket and green towel respectively in Fig.~\ref{fig:success-rate-comp} producing coherent looking but task-incorrect states, and under OOD instructions (Appx. Fig.~\ref{fig:ood-instruction-same-ep}), both maintain the prior action pattern rather than updating to the new goal. Semantic encoders preserve task-level intent at the cost of geometric precision (e.g., VJEPA2.1 under-opens the drawer in Appx.~Fig.~\ref{fig:hallucinated-rollout-pixels2}). We find the latter to better capture semantic distinctions even under instruction shift (e.g., the fold-unfold task in Appx. Fig.~\ref{fig:ood-comparison-same-ep}).

\subsection{Do compressed adapter latents aid semantic encoders further for world modeling?}

\paragraph{Adapters improve diffusion ease but can distort control geometry.}
Fig.~\ref{fig:adapt-vs-noadapt}, Table~\ref{tab:policy-perf}, and Table~\ref{tab:visual-quality} show that the compressed space $d_{96}$ of adapters helps the latent diffusion model, as also observed by \citet{zhang2025both} and \citet{baiSemanticGenVideoGeneration2025}. This leads to generally stronger performance than the native variants on most metrics except latent CEM action error, OOD robustness, and PCK coverage.  These findings hint towards the adapter compressing the latent space in a way that is useful for high-level task completion such as diffusion denoising but hurtful for fine-grained tasks like trajectory optimization, where precise action information is needed.

\subsection{Do high-dimensional semantic latents and adapter add computational overhead?}

  \textbf{High-dimensional semantic latents do not substantially increase DiT compute in our setup.} The DiT always receives the same number of tokens per frame $N{=}256$, hence larger
  channel dimensions only affect the input/output projections (see Appx. \ref{app_subsec:action-conditioned-diffusion} for discussion). The main compute differences instead come from the frozen encoder and decoder
  architectures. In particular, ViT-based semantic encoders paired with the adapter pixel decoder  remain competitive in total GFLOPs, while native high-dimensional semantic spaces require only a lightweight wide DDT head \citep{wang2025ddt}. We report parameter counts
  and GFLOPs split by encoder, adapter, DiT, and decoder in Appx. Table ~\ref{tab:arch-param-compute}.

\begin{tcolorbox}[
title=\textbf{Key Empirical Takeaways},
colback=white,
colframe=wmBest,
boxrule=0.6pt,
arc=2pt,
left=4pt,
right=4pt,
top=4pt,
bottom=4pt,
fonttitle=\bfseries,
float, floatplacement=h
]
\begin{itemize}[leftmargin=1.1em,itemsep=1pt,topsep=2pt]

    \item \textbf{Visual fidelity does not always imply downstream performance.} Reconstruction latents can match or exceed semantic latents on pixel-level metrics, especially at larger DiT scale, yet lag on action recovery, task-success probes, CEM planning, and policy-in-the-loop evaluation.

    \item \textbf{Semantic latents scale better with multiple views.} 

    Under limited data, adding multiple views improves planning but can hurt visual rollouts; semantic encoders retain the action recoverability benefit with substantially less degradation than reconstruction latents.

    \item \textbf{Adapters trade control geometry for diffusion ease.}
    Adapters ease diffusion and decoding, but can distort fine-grained action geometry compared with native semantic features.

    \item \textbf{World models in semantic spaces lower reconstruction and generation ceiling gap.} Training decoders with the same budget for semantic world models is more effective.
    \item \textbf{High-dimensional semantic latents are practical in DiTs.} With a fixed patch-token count, semantic width adds little to the transition-model cost.

\end{itemize}
\end{tcolorbox}

\begin{recipebox}
\section{A Recipe for Semantic Latent Diffusion Robotics World Modeling}

Our findings suggest a practical recipe for building robotic latent diffusion world models. \best{Do not begin by optimizing for visual realism alone}. Instead, choose a latent space that makes \best{action and task progress} explicit, make that space easy for diffusion to model, and evaluate the resulting world model with control- and policy-based metrics. Visual realism can often be improved through better decoder training, but transition quality and latent fidelity remain important. Use robot demonstration datasets with preferably \best{multi-view trajectories} and, when available, success/failure labels to unlock diverse evaluations. \best{Choose pretrained semantic encoders} as the default latent state space, since they preserve action geometry and task progress better than reconstruction latents. \best{Pair them with adapter compression} when decoded rollout quality or VLA-in-the-loop evaluation matters. For transition model, a robust default for high-dimensional semantic spaces is: a \best{spatial-temporal DiT} with causal temporal blocks,  a shallow-wide DDT head \citep{wang2025ddt}, and a dimension-aware noise shifting \citep{zheng2025diffusion}. The spatial blocks stay non-causal since per-frame patches  are denoised jointly. 
For training,  diffusion forcing \citep{chen2024diffusion} can be used for  autoregressive next-frame rollout. Finally, evaluate world models  on \best{multiple axes} covering both visual, latent, and downstream task performance. 
\end{recipebox}

\begin{table}[t]
\centering
\vspace{-1em}
\caption{\textbf{Visual realism quality} for DiT-S and L. \best{Best} and \second{runner-up} within each size group.}
\resizebox{\textwidth}{!}{%
\begin{tabular}{l|cccccc|cc|cccc}
\toprule
  & \multicolumn{6}{c|}{Visual quality} & \multicolumn{2}{c|}{Content consistency} & \multicolumn{4}{c}{Motion quality} \\
\cmidrule(lr){2-7}\cmidrule(lr){8-9}\cmidrule(lr){10-13}
Encoder & SSIM$\uparrow$ & LPIPS$\downarrow$ & FID$\downarrow$ & \shortstack{Image\\quality$\uparrow$} & \shortstack{Aesthetic\\quality$\uparrow$} & \shortstack{JEPA\\sim.$\uparrow$} & \shortstack{Subject\\consist.$\uparrow$} & \shortstack{Depth\\AbsRel$\downarrow$} & \shortstack{Dyn.\\degree$\uparrow$} & \shortstack{Flow\\score$\uparrow$} & FVD$\downarrow$ & t-LPIPS$\downarrow$ \\
  % & $\uparrow$ & $\downarrow$ & $\downarrow$ & $\uparrow$ & $\uparrow$ & $\uparrow$ & $\uparrow$ & $\downarrow$ & $\uparrow$ & $\uparrow$ & $\downarrow$ & $\downarrow$ \\
\midrule
\multicolumn{13}{c}{\normalfont DiT-S} \\
\midrule
\famtag{VAE}{$\bullet$}~VAE & 0.688 & 0.218 & 17.428 & \best{0.592} & 0.467 & 0.871 & 0.810 & 0.390 & 0.767 & 1.186 & 6.829 & 0.0264 \\
\rowcolor{wmGroupBg}\famtag{VAVAE}{$\bullet$}~VA-VAE & 0.633 & 0.226 & 15.488 & \second{0.585} & 0.464 & 0.783 & 0.817 & 0.455 & 0.765 & 1.204 & 8.531 & 0.0253 \\
\famtag{COSMOS}{$\bullet$}~Cosmos & 0.608 & 0.245 & 16.947 & 0.558 & 0.463 & 0.517 & 0.793 & 0.638 & 0.813 & 1.511 & 8.195 & 0.0223 \\
\cmidrule{1-13}
\rowcolor{wmGroupBg}\famtag{VJEPA2}{$\bullet$}~V-JEPA 2.1 & 0.725 & \best{0.176} & 6.771 & 0.578 & \second{0.473} & 0.929 & 0.841 & 0.404 & 0.832 & 1.587 & \second{5.459} & \second{0.0197} \\
\famtag{VJEPA2}{$\bullet$}~V-JEPA 2.1$_{96}$ & \second{0.729} & \second{0.179} & \second{6.302} & 0.579 & \best{0.474} & 0.928 & 0.841 & \second{0.363} & \best{0.843} & \best{1.653} & \best{5.224} & 0.0212 \\
\rowcolor{wmGroupBg}\famtag{WEBSSL}{$\bullet$}~Web-DINO & 0.722 & 0.199 & 7.626 & 0.576 & 0.472 & \second{0.938} & \best{0.849} & \best{0.350} & 0.794 & 1.408 & 6.656 & 0.0234 \\
\famtag{WEBSSL}{$\bullet$}~Web-DINO$_{96}$ & 0.728 & 0.181 & \best{5.998} & 0.574 & 0.473 & \best{0.944} & 0.841 & 0.375 & \second{0.835} & \second{1.634} & 5.510 & \best{0.0195} \\
\rowcolor{wmGroupBg}\famtag{SIGLIP}{$\bullet$}~SigLIP 2 & 0.713 & 0.205 & 7.858 & 0.566 & 0.471 & 0.931 & 0.839 & 0.394 & 0.827 & 1.602 & 6.902 & 0.0228 \\
\famtag{SIGLIP}{$\bullet$}~SigLIP 2$_{96}$ & \best{0.738} & 0.179 & 6.881 & 0.573 & 0.472 & 0.938 & \second{0.843} & 0.372 & 0.827 & 1.547 & 6.005 & 0.0223 \\
\midrule
\multicolumn{13}{c}{\normalfont DiT-L} \\
\midrule
\famtag{VAE}{$\bullet$}~VAE & 0.729 & \second{0.168} & \best{5.351} & \best{0.598} & \best{0.475} & \best{0.980} & 0.827 & \best{0.281} & \best{0.844} & \second{1.635} & \best{3.495} & 0.0202 \\
\rowcolor{wmGroupBg}\famtag{COSMOS}{$\bullet$}~Cosmos & 0.657 & 0.186 & 9.234 & 0.578 & 0.469 & 0.760 & 0.817 & 0.465 & \second{0.843} & \best{1.650} & 6.536 & 0.0199 \\
\cmidrule{1-13}
\famtag{VJEPA2}{$\bullet$}~V-JEPA 2.1 & 0.741 & 0.172 & 6.944 & 0.578 & 0.474 & 0.926 & 0.844 & 0.330 & 0.832 & 1.573 & 5.371 & 0.0195 \\
\rowcolor{wmGroupBg}\famtag{VJEPA2}{$\bullet$}~V-JEPA 2.1$_{96}$ & \best{0.743} & \best{0.165} & \second{6.186} & \second{0.581} & \second{0.474} & 0.929 & 0.842 & 0.346 & 0.831 & 1.558 & \second{5.223} & 0.0201 \\
\famtag{WEBSSL}{$\bullet$}~Web-DINO & 0.729 & 0.192 & 6.918 & 0.573 & 0.472 & \second{0.945} & \second{0.847} & 0.343 & 0.823 & 1.557 & 6.014 & 0.0219 \\
\rowcolor{wmGroupBg}\famtag{WEBSSL}{$\bullet$}~Web-DINO$_{96}$ & 0.741 & 0.189 & 14.259 & 0.578 & 0.466 & 0.709 & \best{0.852} & 0.352 & 0.833 & 1.568 & 13.107 & \best{0.0189} \\
\famtag{SIGLIP}{$\bullet$}~SigLIP 2 & 0.730 & 0.188 & 7.574 & 0.569 & 0.472 & 0.937 & 0.845 & 0.344 & 0.822 & 1.562 & 6.688 & 0.0207 \\
\rowcolor{wmGroupBg}\famtag{SIGLIP}{$\bullet$}~SigLIP 2$_{96}$ & \second{0.743} & 0.171 & 6.740 & 0.573 & 0.472 & 0.937 & 0.844 & \second{0.326} & 0.830 & 1.580 & 5.780 & \second{0.0193} \\
\bottomrule
\end{tabular}
}
\label{tab:visual-quality}
\vspace{0em}
\end{table}

%%%%%%%%%%%%%%%%%%%%%%%%%%%%%%%%%%%%%%%%%%%%%%%%%%%%%%%%%%%%

\section{Related work}

\textbf{Robotic world models} can be seen to span three related objectives. One line treats world models as policy-evaluation environments: WorldGym \citep{quevedo2025worldgym} and WorldEval \citep{li2025worldeval} roll out policies in learned video models; \cite{tseng2025scalable} studies how pretraining, data diversity, and failure modes affect evaluation. A second line adapts pretrained generators into interactive simulators: UniSim \citep{yang2023learning} learns interactive real-world simulators from broad data; Vid2World \citep{huang2025vid2world} causalizes video diffusion with action guidance; Ctrl-World \cite{guo2025ctrl} studies multi-view, long-horizon, policy-in-the-loop manipulation. A third line moves prediction and planning into semantic feature space: DINO-WM \citep{zhouDINOWMWorldModels2025}, DINO-world \citep{baldassarre2025back}, and V-JEPA 2-AC \citep{midovjepa2} show that pretrained representations can support latent space forecasting and zero-shot or few-shot planning. These works establish the utility of both video generation and semantic representations, but do not isolate the encoder-defined latent space within a unified action-conditioned framework.

\textbf{World model evaluation} has moved beyond rollout plausibility and policy ranking toward physics, semantics, and embodied utility~\citep{li2025evaluating,MurLabadia2026VJEPA2U}.
RBench~\citep{li2024rbench} measures task correctness and structural realism. WorldModelBench~\citep{li2025worldmodelbench} highlights instruction-following and physics-adherence failures missed by generic video metrics.  EWMBench~\citep{yue2025ewmbench} evaluates scene consistency, motion correctness, and semantic alignment.
World-in-World~\citep{zhang2025world} prioritizes closed-loop task success, WoW-World-Eval~\citep{fan2026wow} adds inverse-dynamics-based action plausibility, and WorldArena~\citep{shang2026worldarena} exposes the gap between perceptual quality and downstream functionality.
These benchmarks evaluate world models at system-level while we seek to evaluate them at model-level. See Appx. \ref{app:ldm-lit} for a review of LDM.

\section{Future Work and Limitations}
\label{sec:future_work}
Our study isolates the effect of encoder-defined latent spaces within a controlled action-conditioned LDM protocol. The conclusions are therefore scoped to the Bridge V2 manipulation setting and a shared robot embodiment. Evaluating broader embodiments, domains, and data regimes is an important next step. Our policy-in-the-loop experiments also focus on evaluating a fixed VLA policy inside generated rollouts, while policy improvement and sim-to-real transfer would test a complementary use of the same world models.
Lastly, our evaluation partially relies on VLM-based success judgments, which may introduce evaluator bias. We reduce this dependence by aggregating multiple VLMs and pairing them with non-VLM diagnostics, including CEM planning, inverse dynamics, latent success classification, and visual/geometric metrics. 
\section{Conclusion}

Our study shows that the encoder-defined latent space is a central design choice for action-conditioned latent diffusion world models in robotics. Across visual, latent, planning, and policy-in-the-loop evaluations, semantic representation spaces such as that of V-JEPA 2.1, Web-DINO, and SigLIP 2 generally provide stronger action recoverability, task-success classification accuracy, robustness, and downstream policy performance than reconstruction-aligned VAE-style latents, even when the latter remains competitive or superior on low-level photometric metrics. These results support the view that robotic world models should not be selected solely by visual realism, but by whether their latent dynamics preserve action-relevant structure and policy evaluation accuracy.

\begin{ack}
Nilaksh is partly supported by a grant (\url{https://doi.org/10.69777/2009238}) from the Fonds de recherche du Québec (FRQNT).
Saurav Jha is supported by the IVADO postdoctoral fellowship and the Canada First Research Excellence Fund. Sarath Chandar is supported by the Canada CIFAR AI
Chairs program, the Canada Research Chair in Lifelong
Machine Learning, and the NSERC Discovery Grant.
This research was enabled in part by compute resources
provided by Mila (\url{mila.quebec}) and the Digital Research
Alliance of Canada (\url{www.alliancecan.ca}).
\end{ack}

% \newpage
\bibliographystyle{plainnat}
\bibliography{neurips_2026}

\newpage
\appendix

\section{Frequently Asked Questions (FAQs)}
\label{app:faq}
\begin{enumerate}[leftmargin=*, itemsep=2pt]
    \item \textbf{What are the parameter counts and GFLOPs of the full diffusion pipelines for each of the encoder families? How is the parameter/compute parity ensured with adapters and wide heads?}

    We show in Table \ref{tab:arch-param-compute} in Appx. \ref{app:archi-and-training} the summary of the parameter counts and compute required for inference of all semantic spaces, with and without adapters.

    Parameter and compute parity are ensured by keeping the same DiT backbone across all rows and giving every model the same 256 tokens per frame. For adapter-based semantic encoders, the S-VAE adapter compresses high-dimensional features to 96 channels, making the DiT almost identical to the VAE-latent case. For native semantic latents, only the shallow input/output projection or wide head changes, so the \textit{extra parameters do not increase DiT depth and add little compute}. Thus, the comparison is not driven by a larger diffusion model, it isolates the effect of using richer semantic representation spaces, which remain competitive in compute while providing stronger task-relevant structure.

    \item \textbf{How sensitive are the policy-in-the-loop results to the choice of VLM judges? Are inter-judge agreements available?}
    
    The policy-in-the-loop results do show sensitivity to the VLM judge, particularly on harder tasks: agreement is high on simple Level~1 tasks, while Level~2--4 tasks involve finer spatial, contact, deformable-object, and stacking judgments that naturally induce more judge variation; see Table~\ref{tab:vla-per-instruction-ditl} for detailed results. We therefore rate trajectories with three VLMs and select the two most correlated judges, InternVL3.5-14B and Qwen3.6-27B, based on inter-judge Cohen's $\kappa$ agreement (Fig. \ref{fig:vlm-kappa}). To further reduce single-judge dependence, Table~\ref{tab:policy-perf} reports both consensus success rates with variance and Borda ranks, which are less sensitive to absolute score calibration. Finally, our conclusions do not rely only on VLM ratings: we also report task-instruction-conditioned success-classifier metrics on generated latents in Table~\ref{tab:idm-suc-dits}, providing an independent task-conditioned signal that supports the same trends.

    \item \textbf{Why was CEM chosen for latent space planning instead of gradient based planners or differentiable MPC?}

    We use CEM because latent-space planning involves non-convex objectives and noisy gradients. As a derivative-free optimizer, CEM is robust to black-box dynamics and compounding errors \citep{rubinstein2004cross}. Its stochastic search avoids local minima better than gradient-based or differentiable Model Predictive Control (MPC), motivating its use in PlaNet \citep{hafner2019learning} and CEM-MPC \citep{pinneri2021sample}. While gradient planners are faster, they are sensitive to model inaccuracies and gradient instability \citep{bharadhwaj2020model}. Consequently, CEM provides a conservative, reliable baseline for evaluating world-model quality.

    \item \textbf{Is there evaluation on another manipulation dataset or embodiment (e.g., ALOHA, Franka) to test generalization? What are the expected transfer and potential pitfalls?}

    Evaluation on additional embodiments is an important direction but outside the scope of this study, whose controlled comparison is centered on BridgeV2; we did, however, use SOAR data for training the success classifier. We expect the main conclusion, that semantic latents are more policy-relevant than purely reconstruction-aligned latents, to transfer most directly when object-centric semantics and action-conditioned contact dynamics remain comparable. Cross-embodiment evaluation on ALOHA-style bimanual manipulation, Franka setups, or broader simulators such as RoboCasa \citep{robocasa2024, robocasa365} would introduce new challenges: different camera viewpoints, action spaces, gripper morphology, control frequencies, embodiment-specific failure modes, and sim-to-real gaps. These factors may require embodiment-specific action tokenization, calibration, or classifier re-training, making such benchmarks an excellent test of whether semantic latent world models generalize beyond a single robot-data distribution. We also mention this as a potential future work avenue in Section \ref{sec:future_work}.

    \item \textbf{What is the benefit of diffusion models over non-diffusion world models that use semantic features for manipulation like DINO-WM and V-JEPA 2 AC?}

    DINO-WM and V-JEPA 2-AC provide compelling evidence that pretrained semantic features are useful for robotic prediction and planning, and we view them as complementary to our study rather than direct competitors. Our central research question is specifically how the choice of latent space affects \emph{diffusion-based} action-conditioned world modeling, so comparing against non-diffusion architectures would conflate representation choice with model-family differences. Diffusion models are also a natural testbed for this question because they model a distribution over future sequences and can denoise an entire prediction horizon jointly, which may better capture multimodal futures and reduce the compounding errors associated with purely autoregressive one-step regression rollouts, although this is a mitigation rather than a guarantee. Thus, our experiments are intentionally scoped to isolate the effect of semantic versus reconstruction latents within a fixed LDM framework; broader comparisons to non-diffusion semantic world models are important future work.

    \item \textbf{How were the learning rates and other hyperparameters chosen for different encoder latent spaces?}

    We used the same optimizer and learning-rate recipe for all world models, rather than tuning separately for each latent space. Specifically, all DiTs were trained with AdamW, learning rate $10^{-4}$, betas $(0.9,0.99)$, weight decay $2\times 10^{-3}$, gradient clipping, EMA, linear warmup, and cosine decay. Our goal is to isolate the effect of the encoder-defined latent space, and per-encoder hyperparameter tuning would confound the comparison by giving different latent spaces different optimization budgets. For each model-size group, runs were trained under the same schedule and until losses had plateaued. Since each DiT-S run costs roughly 6--7 hours on 4 H100s, each DiT-L run about 34 hours, and adapter/pixel-decoder training about 55 hours, exhaustive sweeps over learning rate, weight decay, warmup, batch size, EMA, and noise schedule for every encoder would be prohibitively expensive. We therefore use a fixed standard recipe and report all models under the same optimization protocol.
    
\end{enumerate}

\section{Architecture and Training Details}
\label{app:archi-and-training}
\begin{table}[H]
\centering
\vspace{-0.5em}
\caption{\textbf{Architecture size and compute.} Adapter-based semantic encoders are marked with $_{96}$ and use the S-VAE adapter with $d{=}96$. Native semantic rows do not use adapter in the DiT and use a shallow-wide DDT head. All DiT parameter counts are for DiT-L. Note that the extra DiT parameters are due to the shallow-wide head, which does not contribute much to the depth of the DiT. For DiTs using high-dimensional latents of V-JEPA 2.1, Web-DINO, and SigLIP, decoding uses the adapter's pixel decoder as the surrogate.}
\resizebox{\textwidth}{!}{%
\begin{tabular}{l|c|c|rrrr|rrrrr}
\toprule
 & & & \multicolumn{4}{c|}{Params (M)} & \multicolumn{5}{c}{GFLOPs/frame} \\
\cmidrule(lr){4-7}\cmidrule(lr){8-12}
Encoder & DiT latent & Adapter & Enc. & Adapt. & DiT & Dec. & Enc. & Adapt. & DiT & Dec. & Total \\
\midrule
\famtag{VAE}{$\bullet$}~SD-VAE & 16 & -- & 34.3 & -- & 910.1 & 49.5 & 270.8 & -- & 316.5 & 620.2 & 1207.5 \\
\rowcolor{wmGroupBg}\famtag{COSMOS}{$\bullet$}~Cosmos-CI16x16 & 16 & -- & 33.5 & -- & 910.0 & 48.0 & 47.6 & -- & 316.5 & 101.7 & 465.9 \\
\famtag{VAVAE}{$\bullet$}~VA-VAE & 32 & -- & 28.4 & -- & 910.0 & 41.4 & 137.9 & -- & 316.5 & 252.2 & 706.6 \\
\cmidrule{1-12}
\rowcolor{wmGroupBg}\famtag{VJEPA2}{$\bullet$}~V-JEPA 2.1$_{96}$ & 96 & S-VAE & 304.7 & 38.1 & 910.1 & 177.0 & 154.7 & 10.6 & 316.5 & 428.3 & 910.1 \\
\famtag{VJEPA2}{$\bullet$}~V-JEPA 2.1 & 1024 & -- & 304.7 & -- & 921.5 & 177.0 & 154.7 & -- & 318.7 & 428.3 & 901.7 \\
\rowcolor{wmGroupBg}\famtag{WEBSSL}{$\bullet$}~Web-DINO$_{96}$ & 96 & S-VAE & 303.7 & 38.1 & 910.1 & 177.0 & 155.8 & 10.6 & 316.5 & 428.3 & 911.2 \\
\famtag{WEBSSL}{$\bullet$}~Web-DINO & 1024 & -- & 303.7 & -- & 921.5 & 177.0 & 155.8 & -- & 318.7 & 428.3 & 902.8 \\
\rowcolor{wmGroupBg}\famtag{SIGLIP}{$\bullet$}~SigLIP 2$_{96}$ & 96 & S-VAE & 427.7 & 46.4 & 910.1 & 177.0 & 211.9 & 12.8 & 316.5 & 428.3 & 969.5 \\
\famtag{SIGLIP}{$\bullet$}~SigLIP 2 & 1152 & -- & 427.7 & -- & 921.9 & 177.0 & 211.9 & -- & 318.8 & 428.3 & 959.0 \\
\bottomrule
\end{tabular}
}
\label{tab:arch-param-compute}
\end{table}

\subsection{Latent Diffusion Modeling (LDM)}
\label{app:ldm-lit}
LDM learns to denoise in compact reconstruction-aligned autoencoder spaces such as that of VAEs \citep{kingma2013auto}.
Recent VAE variants include: Stable Diffusion~3~\citep{esser2024scaling} adapting autoencoding to rectified flow models, VA-VAE~\citep{yao2025vavae} aligning autoencoders with vision foundation models, and Cosmos~\citep{agarwal2025cosmos} providing  tokenizers across flexible compression regimes.
In parallel, semantic-aligned encoders (DINOv2~\citep{oquab2023dinov2}, SigLIP~\citep{tschannen2025siglip}, Qwen-VL~\citep{Qwen2.5-VL, Qwen3-VL}, V-JEPA~2.1~\citep{MurLabadia2026VJEPA2U}) provide structured visual features, but their high dimensionality can make generative modeling unstable ~\citep{skorokhodov2025improving,yu2024image}.
Representation autoencoders (RAEs) address this by pairing frozen pretrained encoders with learned decoders~\citep{zheng2025diffusion,tong2026scaling}, enabling semantic latent spaces that support both visual understanding and generation~\citep{tong2026beyond}.
However, high-dimensional RAE features can still suffer from off-manifold sampling and weak fine-geometry reconstruction~\citep{zhang2025both}, suggesting that RAEs do not simply replace VAEs but instead expose a tradeoff between pixel faithfulness and semantic abstraction~\citep{zhang2026rae}.
For robotics, this tradeoff implies that the best latent space is not necessarily the one that reconstructs frames most faithfully, but the one that preserves action-relevant dynamics for prediction, planning, and policy evaluation.
\subsection{Action-Conditioned Diffusion Model}
\label{app_subsec:action-conditioned-diffusion}
The world model is trained in the latent space of a frozen visual encoder. Let
$o_{0:T-1}$ be a video clip, $a_{0:T-1}$ the corresponding action sequence,
and $f_{\phi}$ the frozen encoder. We first form latents
\begin{equation}
    z_{0:T-1} = f_{\phi}(o_{0:T-1}), \qquad
    z_t \in \mathbb{R}^{N \times D},
\end{equation}
where $N=h \times w$ is the number of spatial tokens and $D$ is the native
encoder channel dimension. In the code tensors are stored as
$h \times w \times D$, but the notation below flattens space to $N$ tokens.
For adapter-based semantic encoders, $z_t$ is further compressed by the adapter
$\alpha_{\psi}$ before being passed to the diffusion model,
\begin{equation}
    \tilde z_t = \alpha_{\psi}(z_t), \qquad
    \tilde z_t \in \mathbb{R}^{N \times d}, \qquad d=96.
\end{equation}
The adapter and encoder are frozen during world model training; only the DiT
parameters are optimized.

\begin{table}[H]
  \centering
  \vspace{-0.5em}
  \caption{\textbf{DiT size presets.} The hidden size, depth and the number of heads for each DiT size.}
  \resizebox{0.5\textwidth}{!}{%
  \begin{tabular}{l|rrrr}
  \toprule
  Preset & Hidden Size $d$ & Depth & Heads & Head Dim $d / h$ \\
  \midrule
  \famtag{DiT}{$\bullet$}~DiT-S  & 384  & 12 & 6  & 64 \\
  \rowcolor{wmGroupBg}\famtag{DiT}{$\bullet$}~DiT-B  & 768  & 12 & 12 & 64 \\
  \famtag{DiT}{$\bullet$}~DiT-L  & 1024 & 24 & 16 & 64 \\
  \bottomrule
  \end{tabular}
  }
  \label{tab:dit-size-presets}
  \end{table}

All DiT runs in Table~\ref{tab:arch-param-compute} use a DiT-L backbone with 24 layers, hidden size 1024, 16 attention heads, and $T{=}10$ frames. The context
length is $H{=}2$, so the model conditions on
$\tilde z_{0:H-1}$ and predicts the future block
$\tilde z_{H:T-1}$ under actions $a_{0:T-1}$ and optional language $\ell$. The
VAE latent has shape $32{\times}32{\times}16$ and is
patchified with DiT patch size $p{=}2$, while all semantic, Cosmos, and VA-VAE
latents use a $16{\times}16$ token grid with $p{=}1$. Thus every row gives the
DiT the same number of tokens per frame:
\begin{equation}
    N = (h/p)(w/p) = 16 \cdot 16 = 256 .
\end{equation}
\textbf{This is the main reason high-dimensional semantic latents do not substantially
increase DiT compute:} the transformer blocks operate on the same token count
and hidden width, and the latent channel dimension only appears in the input
patch projection and output prediction layer.

\paragraph{Shallow-wide DDT head.}
For high-dimensional representation latents, we also use a lightweight \textit{shallow-wide} DDT head \citep{wang2025ddt}. The DiT backbone remains unchanged. The shallow-wide head uses a 2048-dimensional readout width and keeps a
minimal spatial refinement stage before the final patch prediction layer. This adds local spatial processing capacity at the output while leaving the main
DiT backbone unchanged. As a result, the shallow head can improve the mapping from backbone features to high-dimensional representation with minimal increase in parameters.

\paragraph{World model training hyperparameters.}
All world models are trained on Bridge V2 clips resized to
$256{\times}256$, with $T{=}10$ frames, $H{=}2$ history frames, frame skip 2,
and 7-dimensional actions. Unless otherwise stated, the reported single-view runs use distributed data-parallel training on 4 H100 GPUs, per-GPU batch size
16 for DiT-S and 5 for DiT-L, bfloat16 autocast, and \texttt{torch.compile} \citep{ansel2024pytorch}.
The optimizer is AdamW \citep{loshchilov2018decoupled} with learning rate (LR) of $10^{-4}$, betas $(0.9,0.99)$,
weight decay $2{\times}10^{-3}$, $\epsilon{=}10^{-8}$, and gradient clipping at
global norm 1.0. We maintain an EMA copy of the DiT weights with decay
0.9995. The LR schedule is a linear warmup followed by cosine
decay to $0.7$ of the base LR. All runs use 3 LR warmup
epochs and 100 total epochs.

\paragraph{Flow matching.}
The model is trained with the optimal-transport flow-matching objective \citep{lipman2023flow}. For
future frames $i \in \{H,\dots,T-1\}$, we sample $\tau_i \sim p(\tau)$, draw
$\epsilon \sim \mathcal{N}(0,I)$, and linearly interpolate between data and
noise:
\begin{equation}
    \tilde z_{\tau_i,i} = (1-\tau_i)\tilde z_i + \tau_i \epsilon_i .
\end{equation}
The DiT predicts the velocity field
$v_{\theta}(\tilde z_{\tau}, \tau, a_{0:T-1}, \ell)$, and the target velocity is
\begin{equation}
    u_i = \epsilon_i - \tilde z_i .
\end{equation}
With clean history context, the training loss is
\begin{equation}
    \mathcal{L}_{\mathrm{FM}}
    =
    \mathbb{E}_{\tilde z,\epsilon,\tau}
    \left[
    \sum_{i=H}^{T-1}
    \left\|
    v_{\theta}(\tilde z_{\tau,i}, \tau_i, a_{0:T-1}, \ell) -
    (\epsilon_i - \tilde z_i)
    \right\|_2^2
    \right].
\end{equation}
We only apply this loss to future frames. History frames are
used as conditioning context with no diffusion noise $(\tau=0)$. During training, they however receive small Gaussian
augmentation,
\begin{equation}
    \tilde z^{\mathrm{ctx}}_{\mathrm{aug}}
    =
    \frac{\tilde z^{\mathrm{ctx}} + \sigma_h \eta}
         {\sqrt{1+\sigma_h^2}},
    \qquad
    \eta \sim \mathcal{N}(0,I),
\end{equation}
which prevents the model from overfitting to perfectly clean context latents.

\paragraph{Dimension-dependent noise schedule shift.}
For non-VAE latents, the timestep distribution is shifted as a function of the
latent dimensionality seen by the DiT. Following \citet{esser2024scaling} and \citet{zheng2025diffusion}, we use the shift:
\begin{equation}
    \gamma = \sqrt{\frac{(256/p^2)d}{4096}}, \qquad
    \tau' = \frac{\gamma \tau}{1 + (\gamma - 1)\tau}.
\end{equation}
Here $d$ is the DiT input channel count after any adapter. This makes the
noise level depend on the latent representation size rather than only on image
resolution.

\paragraph{Inference and causal attention.}
All our world models carry out autoregressive inference in latent space. Given encoded history
$\tilde z_{0:H-1}$, the sampler appends a Gaussian latent for the next frame and
integrates the learned velocity field backward from $\tau{=}1$ to $\tau{=}0$
with 10 Euler steps \citep{lipman2023flow, esser2024scaling}:
\begin{equation}
    \tilde z_{\tau_{j+1},t} =
    \tilde z_{\tau_j,t} - (\tau_j-\tau_{j+1})
    v_{\theta}(\tilde z_{\tau_j,0:t}, \tau_j, a_{0:t}, \ell)_t .
\end{equation}
The generated frame is then appended to the context and the process repeats for the
desired horizon. Our temporal attention blocks are causal where each
spatial token attends only to its own past states, following the causal video-transformer design used by
VDT~\citep{lu2024vdt}.

\subsection{Adapter}
High-dimensional semantic encoders produce per-patch features $z \in \mathbb{R}^{N \times D}$
that are prohibitively expensive for the diffusion model to operate on directly.
We pair them with an S-VAE adapter~\citep{zhang2025both} that compresses $z$ to
a compact latent $\tilde{z} \in \mathbb{R}^{N \times d}$ ($d \ll D$).
The adapter $a_\psi$ comprises a Transformer encoder $g_{\psi}^{\mathrm{enc}}$,
a per-token diagonal-Gaussian bottleneck, and a Transformer decoder $g_{\psi}^{\mathrm{dec}}$:
\begin{align}
    h &= g_{\psi}^{\mathrm{enc}}(z), \\
    (\mu,\log\sigma^2) &= W_{\mu,\sigma^2}\, h, \\
    \tilde{z} &= \mu + \sigma \odot \xi, \qquad \xi \sim \mathcal{N}(0,I), \\
    \hat{z} &= g_{\psi}^{\mathrm{dec}}(\tilde{z}).
\end{align}
Both $g_{\psi}^{\mathrm{enc}}$ and $g_{\psi}^{\mathrm{dec}}$ consist of 3 Transformer blocks at
dimension $D$, each followed by LayerNorm. The encoder appends a linear head
$D \to 2d$ and the decoder prepends a linear head $d \to D$.
We default to using 12 attention heads and FFN width of 3072.

The adapter training loss is:
\begin{equation}
\begin{split}
    \mathcal{L}_{\mathrm{adapter}}
    =
    \underbrace{\mathcal{L}_{\mathrm{MSE}}(z,\hat{z})
    + \lambda_{\mathrm{cos}}\mathcal{L}_{\mathrm{cos}}(z,\hat{z})
    + \lambda_{\mathrm{spec}}\mathcal{L}_{\mathrm{FFT}}(z,\hat{z})}_{\text{semantic reconstruction}}
    + \\ \lambda_{\mathrm{KL}}\,D_{\mathrm{KL}}\!\left(
      q_\psi(\tilde{z}\mid z)\,\|\,\mathcal{N}(0,I)
    \right)
    + \lambda_{\mathrm{pix}}\,\mathcal{L}_{\mathrm{pix}}(o,\hat{o}),
    \end{split}
    \label{eq:adapter_loss}
\end{equation}
where $\hat{o} = \mathrm{Dec}(\tilde{z})$ is the pixel-decoder reconstruction.
$\mathcal{L}_{\mathrm{MSE}}$ and $\mathcal{L}_{\mathrm{cos}} = 1 - \cos(z,\hat{z})$
jointly enforce feature-space fidelity: MSE penalises magnitude errors while the
cosine term preserves directional (semantic) structure.
$D_{\mathrm{KL}}$ regularizes the approximate posterior
$q_\psi(\tilde{z}\mid z) = \mathcal{N}(\mu, \sigma^2 I)$ toward a standard Gaussian prior.
$\mathcal{L}_{\mathrm{FFT}}$ is an $\ell_1$ loss on 1-D FFT magnitudes along the
spatial-token axis, penalizing loss of high-frequency structure through the bottleneck.
$\mathcal{L}_{\mathrm{pix}} =
  \mathcal{L}_{\mathrm{MSE}}(o,\hat{o})
  + \lambda_{\mathrm{LPIPS}}\mathcal{L}_{\mathrm{LPIPS}}
  + \lambda_{\mathrm{SSIM}}(1 - \mathrm{MS\text{-}SSIM})$
grounds the compact latent in pixel space.
Following \citet{zhang2025both}, we use $\lambda_{\mathrm{spec}}{=}0.01$,
$\lambda_{\mathrm{LPIPS}}{=}\lambda_{\mathrm{SSIM}}{=}0.5$.
During DiT training, $\alpha_\psi$ is frozen and applied deterministically
($\tilde{z} = \mu$) as a fixed projection into the compact latent space.

\paragraph{Adapter training hyperparameters.}
The encoder is frozen throughout adapter training. It is trained for 200 total epochs on Bridge V2, per-GPU
batch size 16 for single-view training, and bfloat16 autocast. The optimizer is
AdamW with betas $(0.9,0.99)$ and weight decay $10^{-4}$. The base adapter
learning rate is $10^{-4}$ for the single-view run; the pixel decoder uses a
3$\times$ learning-rate multiplier when trained jointly. Multi-view adapter
fine-tuning uses learning rate $5{\times}10^{-5}$ and lower per-GPU batch
sizes because each sample contains three camera views. The KL coefficient is
linearly warmed up for the first 20\% of optimizer steps to
$\lambda_{\mathrm{KL}}{=}10^{-4}$, while $\lambda_{\mathrm{cos}}{=}1$ and
$\lambda_{\mathrm{pix}}{=}1$. LPIPS, when enabled, is evaluated in float32
after a 50k-sample perceptual warmup. Gradients for both the adapter and pixel
decoder are clipped to norm 1.0. 

\subsection{Pixel Decoder}
The semantic encoders  use the adapter pixel decoder for reconstruction. The pixel
decoder maps compact latents $\tilde z \in \mathbb{R}^{N \times 96}$
to RGB observations:
\begin{equation}
    \hat o = \mathrm{Dec}(\tilde z) = D_{\omega}^{\mathrm{pix}}(\tilde z).
\end{equation}
Architecturally, it is an LDM-style convolutional decoder with two residual blocks per level, and a 4-head
self-attention block at $16{\times}16$ resolution. For the S-VAE setup, the pixel decoder is trained on detached adapter latents with the pixel loss $\mathcal{L}_{\text{pix}}$.
As such, the pixel loss does not backpropagate into the adapter. The pixel reconstruction
loss used in adapter training is
\begin{equation}
    \mathcal{L}_{\mathrm{pix}}
    =
    \|\hat o-o\|_2^2
    + \lambda_{\mathrm{LPIPS}}\mathcal{L}_{\mathrm{LPIPS}}(\hat o,o)
    + \lambda_{\mathrm{SSIM}}\left(1-\mathrm{MS\text{-}SSIM}(\hat o,o)\right).
\end{equation}
In the S-VAE stage, the pixel decoder is trained on detached adapter latents,
so pixel loss does not backpropagate into the adapter. The reported
experiments use this S-VAE path rather than the older PS-VAE mode. For native
semantic DiTs without an adapter in the diffusion model, visualization still
uses the same surrogate path: native latent $\rightarrow$ adapter encoder
$\rightarrow$ pixel decoder.

\subsection{Encoder-specific overhead}
Table~\ref{tab:arch-param-compute} summarizes parameter counts and compute.
We split the parameter counts  by frozen encoder, adapter, DiT, and decoder.
GFLOPs are reported per frame for a single $256{\times}256$ frame by
counting multiply-add as two separate floating-point operations. The total compute column
adds encoder, adapter projection when used, one DiT velocity evaluation, and
the decoder used for visualization/reconstruction.
The differences in total GFLOPs in Table~\ref{tab:arch-param-compute} are
therefore mostly due to the frozen encoder and decoder, and not the DiT backbone itself.
The DiT sees the same $N{=}256$ tokens per frame across all models, so increasing
the semantic latent channel dimension mainly changes the input/output
projections. In contrast, the encoders use different network families:
VAE and VA-VAE are convolutional autoencoders operating over high-resolution
feature maps,  V-JEPA 2.1 and Web-DINO are ViT-style patch encoders \citep{dosovitskiy2021an}, and
SigLIP 2 is a larger, higher-capacity ViT-style vision model. Decoder compute
also differs substantially: VAE uses its native convolutional decoder,
VA-VAE uses a lighter convolutional decoder, and the semantic encoders use the
adapter pixel decoder from a compact $16{\times}16$ latent grid. Thus the
native 1024--1152D semantic rows have nearly the same DiT GFLOPs as their
adapter-based $d_{96}$ counterparts.

Table~\ref{tab:training-compute} reports the measured training time and GPU
configuration for the adapter/pixel-decoder stage and the DiT scaling runs.

\begin{table}[t]
\centering
\caption{\textbf{Training wall-clock and compute resources.} Times are measured
wall-clock durations for the reported Bridge V2 training runs on 4 H100 GPUs and
exclude one-time dataset staging operation. Note that all latent spaces roughly take the same training time due to the fixed number of token count}
\resizebox{\textwidth}{!}{%
\begin{tabular}{lcccccc}
\toprule
Training run & Model size & Epochs & GPUs & Per-GPU batch & Precision & Wall-clock \\
\midrule
Adapter + pixel decoder & S-VAE + CNN decoder & 200 & 4$\times$H100 & 16 & bf16 & $\sim$55 h \\
World model & DiT-S & 100 & 4$\times$H100 & 16 & bf16 & 6--7 h \\
World model & DiT-L & 80 & 4$\times$H100 & 5 & bf16 & $\sim$34 h \\
\bottomrule
\end{tabular}
}
\label{tab:training-compute}
\vspace{-0.5em}
\end{table}

\subsection{Inverse Dynamics Model (IDM)}
\label{app:idm_subsec}
The Inverse Dynamics Model (IDM) \citep{tian2025predictive} is a patch-token Transformer trained to predict an action chunk $\hat{a}_{t:t+k-1} \in \mathbb{R}^{k \times d_a}$ from a window of $k+1$ consecutive encoder latents $(z_t, z_{t+1}, \ldots, z_{t+k})$, where each $z_t = f_\phi(o_t) \in \mathbb{R}^{N \times D}$ is the spatial patch grid produced by the frozen encoder $f_\phi$ directly, \textit{i.e.}, no adapter $a_\psi$ is applied, so the IDM always operates in the native encoder channel space of dimension $D$. Each frame's $N = h \times w$ patch tokens are projected by a shared linear layer into a model-width embedding, augmented with factored temporal and spatial positional embeddings, and then flattened into a joint sequence of $(k+1) \cdot N$ tokens. A set of $k$ learned per-step class token (CLS) readout queries is prepended to this sequence; all tokens attend jointly through $L$ pre-norm Transformer blocks with scaled dot-product self-attention \citep{vaswani2017attention}, and the final-layer representations of the $k$ CLS positions are decoded by a two-layer MLP head to the predicted action chunk $\hat{a}_{t:t+k-1}$. Following \citet{tian2025predictive}, we  train each encoder-specific IDM on real encoded trajectories from Bridge V2 with Smooth-L1 loss.

The IDM serves as a probe of action recoverability for each encoder space $f_\phi \in \Phi$. After training, it is evaluated at horizons $k \in \{1, 4\}$ using Pearson $r$ between the predicted action chunk $\hat{a}_{t:t+k-1}$ and the ground-truth $a^*_{t:t+k-1}$, averaged over the $d_a$ continuous action dimensions. Critically, the same frozen IDM head is then applied without retraining to world model-generated latent pairs $(\hat{z}_t, \hat{z}_{t+k})$ from DiT rollouts of the same episodes. The Pearson $r$ of the real-WM gap thus measures generation-induced erasure of the action-discriminative geometry in the latent space, a form of degradation invisible to pixel-level metrics such as SSIM or LPIPS.

\subsection{VLA success classifier probe}
The success classifier probe $s_\phi$ is a spatio-temporal Transformer trained on full latent
trajectories $z_{0:T}$ from the SOAR dataset \citep{zhou2024autonomous} to classify episode success $y \in \{0,1\}$
given the language instruction $\ell$. Each trajectory's spatial latent grid is first
downsampled to a $4{\times}4$ super-patch grid via adaptive average pooling, yielding
$P{=}16$ spatial tokens per frame, and linearly projected to a shared model width of 384.
Factored temporal and spatial positional embeddings are added in place, producing a token
tensor of shape $(T \times P)$; a learned \textsc{cls} token is then prepended.
Each of the six blocks of the success probe applies three sequential sub-operations with
pre-norm and residual connections: a) spatial self-attention within each frame independently
over the $P$ patch tokens, b) temporal self-attention across the $T$ frames independently
per patch position, and c) cross-attention from all video tokens to the frozen SigLIP\,2
token sequence encoding $\ell$, followed by a SwiGLU FFN\@.
After the final RMSNorm, the mean of the $T \times P$ patch token representations
is passed through a linear head to produce a binary logit $\hat{y}$.

The probe is trained with binary cross-entropy on SOAR episodes, with the encoder
$f_\phi$, adapter $a_\psi$, and SigLIP\,2 text encoder all frozen; only the parameters of $s_\phi$
 are updated. Instruction-mismatch negatives (episodes paired with a language
instruction drawn from a different task family) are mixed in to force the above cross-attention mechanism
to genuinely ground success in the video content rather than ignoring $\ell$.
Checkpoints are selected by balanced accuracy with ROC-AUC as the tie-breaker,
accounting for SOAR's 1:2 success-to-failure class imbalance.
At evaluation, the same frozen $s_\phi$ is applied without retraining to world model-generated latent trajectories $\hat{z}_{0:T}$ from DiT rollouts of the same SOAR
episodes. The drop in balanced accuracy from \textbf{Enc.\,Acc} to \textbf{WM\,Acc}
measures the semantic drift, \textit{i.e.}, the degree to which the transition model $p_\theta$ degrades
task-outcome separability in latent space over the full rollout horizon, a signal
invisible to per-step action metrics.

\section{Evaluation metrics}
\label{app:eval_metrics}
\subsection{Planning and downstream policy performance}
We evaluate planning and policy performance through three complementary sub-protocols: CEM-based latent controllability, VLA-in-the-loop closed-loop success, and robustness under distribution shift.
Throughout, $a_t \in \mathbb{R}^{d_a}$ is the action vector with seven degrees of freedom, $a^*_{t:t+k-1}$ is the ground-truth $k$-step action sequence, and $\tilde{z}_t$ is the compact latent on which the DiT $p_\theta$ operates.

\paragraph{A) CEM action controllability.}
We evaluate whether a trained world model preserves action information by asking whether actions can be recovered from its latent dynamics. Given a held-out transition window with two real context latents,
$(\tilde z_t,\tilde z_{t+1})$, ground-truth action sequence
$a^*_{t+1:t+k}$, and target future latents
$\tilde z^*_{t+2:t+k+1}$, we solve
\begin{equation}
  a^{\mathrm{plan}}_{t+1:t+k}
  =
  \argmin_{a_{t+1:t+k}}
  \frac{1}{k}
  \sum_{j=1}^{k}
  \left\|
    p_\theta^{(j)}(\tilde z_t,\tilde z_{t+1}, a_{t+1:t+k})
    - \tilde z^*_{t+1+j}
  \right\|_2^2 .
  \label{eq:cem_appendix}
\end{equation}
Here $p_\theta^{(j)}$ denotes the $j$th autoregressive latent prediction from the world model. We report results for $k \in \{1,4\}$ using 100 held-out windows per model.

The optimization in Eq.~\eqref{eq:cem_appendix} uses the cross-entropy method (CEM)~\citep{rubinstein2004cross}. For each transition window, CEM maintains a diagonal Gaussian over the optimized action coordinates for all $k$ steps. In the reported runs, we use a population of 400 candidate action sequences, 5 CEM
iterations, and 50 elites per iteration, i.e. an elite fraction of $0.125$. The sampling distribution is initialized with mean $a^*_{t+1:t+k}$ on the searched coordinates and standard deviation equal to one quarter of the action range for each searched coordinate. After each iteration, the Gaussian mean and standard
deviation are set to the empirical mean and standard deviation of the elite set. 

Each CEM candidate is evaluated with one latent rollout sample. The diffusion sampler uses the same inference setting as evaluation, with 10 flow-matching Euler steps per predicted latent frame. To make the CEM objective deterministic for a given transition, we sample one Gaussian rollout-noise tensor per transition window and reuse it for all candidates and all CEM iterations. Thus the world-model rollout is stochastic across evaluation windows through the
sampled diffusion noise, but the optimizer sees a fixed objective within each window. For $k>1$, candidates are evaluated by a joint autoregressive rollout: after the first predicted latent, the prediction is appended to the context and used to predict the next latent under the next candidate action.

We compute the \textbf{CEM error} from the recovered action sequences: $\frac{1}{k}\sum_{j=1}^{k}
  \|a^{\mathrm{plan}}_{t+j,S}-a^*_{t+j,S}\|_2$,
  averaged over transitions, where $S$ is the set of searched action
  dimensions. Lower error indicates that the world-model latent dynamics are
  more action-sensitive under the CEM inversion test.

\paragraph{B) VLA-in-the-loop closed-loop success.}
We roll out OpenVLA-7B~\citep{kim2025openvla} inside each world model for 50-step episodes across 20 Bridge~V2 test episodes with $8$ independent trials per episode (\textit{i.e.}, $N=80$ total rollouts ).
Each rollout video is scored by two VLMs, InternVL-3.5-14B~\citep{wang2025internvl3} and Qwen-3.6-27B~\citep{qwen3.6-27b} using 16 tail-biased frames sampled from the rollout. We use these to compute the following closed-loop success metrics:

\begin{itemize}[leftmargin=*, itemsep=2pt]

\item Consensus success rate (\textbf{Consensus SR}) reports the fraction of trials scored as a success by \emph{both} raters simultaneously:
$\mathrm{CSR} = \frac{1}{N}\sum_{i}\mathbf{1}[\mathrm{score}_i^{\mathrm{InternVL}} \geq 0.5 \wedge \mathrm{score}_i^{\mathrm{QwenVL}} \geq 0.5]$.
Requiring agreement from both raters reduces false positives from any single rater's miscalibration.

\item \textbf{Borda rank} is the sum of rank positions across both raters within each DiT-size group:
$\mathrm{Borda} = r_{\mathrm{InternVL}} + r_{\mathrm{QwenVL}}$,
where $r_{\mathrm{InternVL}}$ and $r_{\mathrm{QwenVL}}$ are the ordinal ranks of the model by SR-InternVL and SR-QwenVL respectively: rank 1 being the best.
This is an ordinal measure robust to rater calibration drift and a lower score is better.
\end{itemize}

\paragraph{C) VLM interaction-quality rubric.}
Each rollout is additionally scored by InternVL~3.5 \citep{wang2025internvl3} on a structured rubric with three independent sub-scores on a 1--5 integer scale, then averaged across the $N$ trials \citep{shang2026worldarena}. It is rated by a VLM using the prompt described in Sec. \ref{app:vlm-judge}.

\begin{itemize}[leftmargin=*, itemsep=2pt]
    \item Interaction quality score (\textbf{IQ score}$\uparrow$) measures the 
plausibility of robot--object contact, including whether grasps, pushes, and force transfers look realistic and avoid interpenetration artifacts.
This helps capture whether the world model renders credible manipulation dynamics without requiring pixel-level ground truth.
\item Instruction following (\textbf{Instr. follow}$\uparrow$) is the
degree to which the rollout visually executes the language instruction $\ell$ (e.g., grasping the correct object, moving in the specified direction).
Instruct follow is complementary to binary SR in the sense that it captures partial progress on episodes where neither judge counts the rollout as a full success.
\end{itemize}

\paragraph{D) Out-of-Distribution (OOD) robustness.}
We re-run a subset of 10 tasks from the 20 used for calculating VLA SR, $8$-trial setup under two independent perturbations.
Distractor-object (OOD distractor) rollouts add OOD objects to the scene as described in Sec. \ref{app:vlm-judge}, while  OOD-instruction rollouts replace the language instruction $\ell$ with a semantically unrelated instruction drawn from a different Bridge~V2 task family.
Success rates under perturbation use the mean of the two per-rater SRs:

\begin{itemize}[leftmargin=*, itemsep=2pt]
    \item OOD SR Distractor:
the per-rater mean SR under distractor objects. 
\item OOD SR instruction:
the per-rater mean SR under the substituted instruction.
\end{itemize}

\subsection{Pixel fidelity and scene geometry}
\label{app:pixel-metrics}
Action faithfulness is a necessary but not sufficient condition for world modeling, \textit{e.g.}, a  model that steers correctly yet generates physically implausible scenes will still mislead a policy that relies on visual observations. We thus evaluate decoded rollout quality across three categories — visual quality, content consistency, and motion quality — each containing \emph{reference-based} metrics that compare generated frames $\hat{o}_t$ to paired ground-truth frames $o_t^*$, and \emph{reference-free} perceptual metrics~\citep{shang2026worldarena} that score generated clips without a ground-truth counterpart.
All metrics are computed over 1{,}000 test episodes.

\paragraph{A) Visual quality.} Reference-based metrics include:
\begin{itemize}[leftmargin=*, itemsep=2pt]
    \item \textbf{PSNR}$\uparrow$ measures the
    peak signal-to-noise ratio $10\log_{10}(1/\mathrm{MSE}(\hat{o}_t, o_t^*))$, averaged over frames and episodes.
    This helps quantify pixel-level reconstruction accuracy but not the perceptual structure.

    \item \textbf{SSIM}$\uparrow$ measures
    structural similarity~\citep{wang2004image} between $\hat{o}_t$ and $o_t^*$, computed on luminance with a local window.
    Captures structural and contrast coherence that PSNR misses.

    \item \textbf{LPIPS}$\downarrow$ measures the learned Perceptual Image Patch Similarity~\citep{zhang2018unreasonable} using AlexNet \citep{krizhevsky2012imagenet} features.
    LPIPS correlates better with human perceptual judgments than pixel-level metrics, penalizing blurry or structurally incorrect generations even when MSE is low.

    \item \textbf{FID}$\downarrow$ quantifies the Fréchet Inception Distance~\citep{heusel2017gans} between the distribution of generated and ground-truth frames, computed from InceptionV3 2048-D features \citep{szegedy2016rethinking}.
    FID measures the population-level gap between generated and real frame distributions, capturing systematic biases that per-frame metrics average away.
\end{itemize}

Reference-free metrics are borrowed from \citet{shang2026worldarena} and include: 
\begin{itemize}[leftmargin=*, itemsep=2pt]
    \item \textbf{Image quality}$\uparrow$ measures the 
    MUSIQ~\citep{ke2021musiq} multi-scale image quality score, normalized to $[0,1]$.
    This helps quantify the perceptual quality of individual frames using a model trained on human quality ratings, without requiring a ground-truth reference.

    \item \textbf{Aesthetic quality}$\uparrow$ uses the
    LAION aesthetic predictor score~\citep{schuhmann2022laion}, normalized to $[0,1]$ from a raw $[0,10]$ scale.
    This helps capture the compositional and stylistic appeals of generated frames independently of content accuracy.

    \item \textbf{JEPA similarity $\uparrow$}  measures the
maximum mean discrepancy (MMD) between  feature distributions extracted from JEPA \citep{assran2023self} to provide evaluation results that better align with human
perception.
\end{itemize}

\paragraph{B) Motion quality.} Reference-based metrics include:
\begin{itemize}[leftmargin=*, itemsep=2pt]
    \item \textbf{FVD}$\downarrow$ measures the
    Fréchet Video Distance~\citep{unterthiner2018towards} computed from ResNet-3D features on 16-frame clips.
    FVD helps extend FID to the temporal domain, thus capturing spatiotemporal distribution quality of full video clips rather than individual frames.

    \item \textbf{t-LPIPS}$\downarrow$ uses
    RAFT~\citep{teed2020raft} to estimate the optical flow $\mathbf{u}_{t-1 \to t}$ on the ground-truth frames. Both generated and ground truth (GT) frames are then warped with this shared flow.
    t-LPIPS is the mean absolute difference between the per-step LPIPS of the flow-warped generated video and the flow-warped GT video.
    Using GT flow as a shared reference decouples temporal dynamics quality from content. Here, a low score signifies the model's frame-to-frame motion pattern matches ground truth.

    \item \textbf{PCK coverage}$\uparrow$ uses 
    CoTracker~\citep{karaev23cotracker} to track a $16{\times}16$ grid of query points placed on the first context frame through the generated video.
    PCK coverage is the mean fraction of these query points that remain visible (tracked with high confidence) at each rollout step.
    A drop across steps indicates that the generated video causes points to leave the frame or become untrackable, which implies geometric instability.
\end{itemize}

Reference-free metrics are borrowed from \citet{shang2026worldarena} and include:\begin{itemize}[leftmargin=*, itemsep=2pt]
    \item \textbf{Dynamic degree}$\uparrow$ measures the
    fraction of inter-frame pairs in a generated clip where RAFT-estimated optical flow magnitude exceeds a threshold $\tau{=}6$ pixels.
    A near-zero value indicates a nearly static rollout, which is unlikely to be action-faithful regardless of pixel quality.

    \item \textbf{Flow score}$\uparrow$ quantifies the
    mean magnitude of the top-5\% of optical flow vectors across all inter-frame pairs in a generated clip.
    This helps capture the strength of dominant motion events, complementing dynamic degree which only measures their frequency.
\end{itemize}

\paragraph{C) Reconstruction ceiling.}
For each encoder, all reference-based metrics are additionally computed on \emph{reconstructed} frames, \textit{i.e.}, real observations encoded and decoded without any DiT. This gives us a per-encoder upper bound.
The  gap $\Delta$  is the difference between the world model score and this ceiling, isolating the quality loss attributable to the transition model rather than the decoder.
A large gap indicates that the DiT struggles to generate in-distribution latents while a small gap implies that the encoder–decoder path is not the bottleneck.

\subsection{Latent representation quality}

\paragraph{A) Action Recoverability.} A world model can score well on PSNR/SSIM yet  use an encoder that never encoded action information to begin with, or use a good encoder but a DiT that overlooks the action-discriminative geometry during denoising. Action recoverability metrics seek to address these and include the following reference-based measures:
\begin{itemize}[leftmargin=*, itemsep=2pt]

  \item \textbf{IDM Pearson $r$ (Encoder)} uses
  an Inverse Dynamics Model (IDM) head \citep{tian2025predictive}  trained on consecutive frozen encoder latent
  pairs $(z_t,\,z_{t+k})$ from Bridge~V2  to predict an action chunk
  $\hat{a}_{t:t+k-1} \in \mathbb{R}^{k \times 7}$ for horizon $k\in\{1,4\}$.
  Pearson~$r$ is then computed by averaging over the six continuous action dimensions on held-out
  real encoded frames, establishing the maximum step-level action information
  linearly accessible from each encoder space.

  \item \textbf{IDM Pearson $r$ (WM)} applies
  the same frozen IDM (trained on real latents)  to world model
  generated latent pairs $(\hat{z}_t,\,\hat{z}_{t+k})$ from DiT rollouts of the
  same episodes.
  A small Real--WM $r$ difference confirms the transition model faithfully preserves
  action-relevant latent geometry during generation while a large difference exposes
  degradation invisible to pixel metrics.
  A large gap between Real and WM $r$ indicates generation-induced erasure of action-distinguishing
  structure even when decoded pixels look faithful.

\end{itemize}

\paragraph{B) Success classifier Accuracy or Success Separability.}
We seek to measure whether the world model's generated latent trajectories retain enough task-outcome structure for a frozen success classifier to distinguish successful from failed episodes, \textit{i.e.}, the DiT preserves semantic meaning over a full rollout and not just local action geometry. Semantic fidelity includes the following reference-based metrics that require ground-truth success/failure labels:
\begin{itemize}[leftmargin=*, itemsep=2pt]

  \item \textbf{Enc.~Acc} signify the encoder ceiling, where
  a factored spatial--temporal attention probe $g_\phi$, conditioned on frozen
  SigLIP~2 text tokens, is trained on real encoder latent trajectories
  $z_{0:T}$ from SOAR~\citep{zhou2024autonomous} to classify task success given
  the language instruction.
  Balanced accuracy  on held-out real-encoded trajectories establish
  the probe ceiling, \textit{i.e.}, the maximum task-success information preserved in each
  encoder space.

  \item \textbf{WM Acc.} applies
  the frozen probe $g_\phi$ is applied without retraining to full world model generated
  latent rollouts of the same episodes.
  Lower WM Acc relative to Enc.~Acc reveals \emph{semantic drift}: the
  generated trajectory has lost task-outcome separability even when per-step
  action signals remain partially intact.

\end{itemize}

\subsection{VLA-based evaluations}
\label{app:vla-eval-ood}
\begin{table}[H]
\centering
\caption{\textbf{OOD-instruction evaluation pairs.} Each original instruction is paired with a single semantically-related but behaviorally distinct instruction. Variations span four types: action reversal (same scene, opposite action), action\,+\,target change, spatial relation change, and target location change.}
\label{tab:ood-instruction-pairs}
\resizebox{\textwidth}{!}{%
\begin{tabular}{p{6.5cm}p{6cm}l}
\toprule
Original instruction & OOD instruction & Variation \\
\midrule
close oven & open the oven & \textit{Action reversal} \\
\rowcolor{wmGroupBg}open the drawer & close the drawer & \textit{Action reversal} \\
fold the cloth from the bottom to the top & unfold the cloth flat & \textit{Action reversal} \\
\rowcolor{wmGroupBg}sweep into pile & scatter the pile across the table & \textit{Action reversal} \\
pick up sponge and wipe plate & drop the sponge into the sink & \textit{Action + target change} \\
\rowcolor{wmGroupBg}Move the can behind the blue fork & place the can on top of the blue fork & \textit{Spatial relation} \\
pick up blue towel from the grey thing and placed it to the right of the white basket & put the blue towel inside the white basket & \textit{Spatial relation} \\
\rowcolor{wmGroupBg}put the covering lid on top of the silver pot & put the lid inside the silver pot & \textit{Spatial relation} \\
moved the blue scrubber onto the lower right burner & move the blue scrubber to the upper left burner & \textit{Spatial location} \\
\rowcolor{wmGroupBg}place the silver pot in the middle of the table & place the silver pot in the sink & \textit{Target location} \\
\bottomrule
\end{tabular}
}
\vspace{-1.2em}
\end{table}
We manually pick the set of 20 tasks present in Table \ref{tab:vla-per-instruction-ditl} to have a good mix of task difficulties as well as task diversities from the Bridge V2 test set. The tasks involve instructions like pick and place, opening/closing, interacting with non-rigid objects like clothes, and tasks that require precise arm and gripper control. We use Claude Opus 4.7 to generate the OOD instructions given the original task instruction in Table \ref{tab:ood-instruction-pairs}. These OOD instructions also span several  variations. 

\subsection{VLM prompts}
\label{app:vlm-judge}

We list the exact prompts we used to create the out of distribution distractor images, score the VLA policy trajectories, and to score the interaction quality and related metrics. We provide a summary of the full prompt from \citet{shang2026worldarena} for the latter here. We chose a subset of 10 tasks equally sampled from the difficulty levels in Table. \ref{tab:vla-per-instruction-ditl} and use ChatGPT Images 2.0 model with the distractor prompt given below to generate the initial frame with OOD objects added to the scene. 

\begin{wmPromptBox}
  {Distractor Image Editing}
  {Text-guided distractor insertion for OOD distractor objects}
Exact prompt template:
\begin{quote}\ttfamily\small
This is an initial observation for a robotics task \{task\_instruction\}.\\
Modify this image by adding distraction objects in the scene in a natural way
without moving or changing any objects in the original scene.\\[1mm]
Requirements:\\
- The robotic arm should be visible.\\
- With the distractors, the task \{task\_instruction\} should remain achievable.
\end{quote}

\end{wmPromptBox}

\begin{wmPromptBox}
  {Episode Success/Failure Scoring}
  {Online rollout scoring used by policy-in-the-loop evaluation}
Prompt structure:
\begin{quote}\ttfamily\small
Here is a sequence of frames from a robot policy which has been rolled out in a
video-generation-based world model. I need your help determining whether the
policy is successful. How successfully does the robot complete the following
task?\\[1mm]
Instruction: \{instruction\}\\[1mm]
Score rubric:\\
0 = Failure\\
0.5 = Partial (optional, when partial criteria are provided)\\
1 = Success\\[1mm]
Provide brief reasoning (2--3 sentences). Then output exactly one final line:
Final Score: X
\end{quote}

The binary version uses only \texttt{0/1}. The partial-credit version adds
\texttt{0.5} when a \texttt{partial\_criteria} string is present.
\end{wmPromptBox}

\begin{wmPromptBox}
  {Interaction Quality / Perspectivity / Instruction Following}
  {Multi-dimensional VLM judge rubric for paper-style interaction-quality metrics.}
Prompt summary:
\begin{itemize}
\item Evaluates \textbf{three} dimensions on a 1--5 Likert scale:
Interaction Quality, Perspectivity, and Instruction Following.
\item Scene prior is explicit: tabletop or counter-top \emph{robotic arm}
manipulation, not human-hand videos.
\item Includes a hard hallucination check: if the video shows human hands
instead of robotic arms, Instruction Following should be scored at most 2.
\item Requires the model to base judgments only on visible evidence in the
sampled frames and to consider temporal coherence.
\item Output is forced to a single JSON object with exactly three top-level
keys:
\begin{quote}\ttfamily\small
\{"Interaction\_Quality": \{"score": 1--5, "reason": "..."\},\\
"Perspectivity": \{"score": 1--5, "reason": "..."\},\\
"Instruction\_Following": \{"score": 1--5, "reason": "..."\}\}
\end{quote}
\end{itemize}

\end{wmPromptBox}

\begin{figure}[h]
    \centering
    \includegraphics[width=0.5\textwidth]{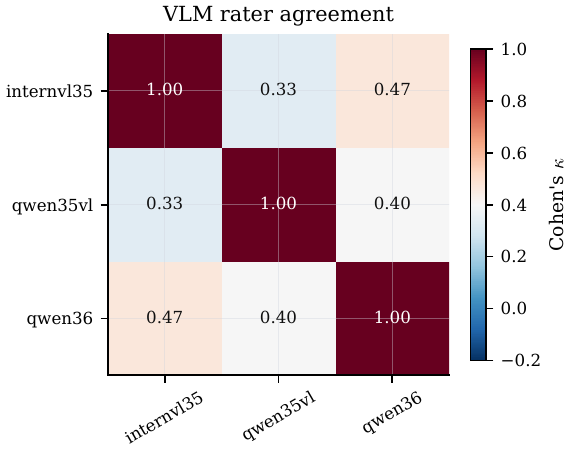}
    \caption{\textbf{The Cohen's kappa} for inter-VLM rater agreement. Given the higher agreement between InternVL 3.5 and Qwen 3.6, we choose these as our VLM judges for policy-in-the-loop task success experiments.}
    \label{fig:vlm-kappa}
\end{figure}

We rated all trajectories using three open-source, but strong Vision Language Models (VLMs), InternVL3.5-14B \citep{wang2025internvl3}, Qwen3.6-27B \citep{qwen3.6-27b}, and Qwen3.5-9B \citep{qwen3.5} with the same scoring prompt and sampled frames. We sampled 16 frames from each episodes, with 10 sampled uniformly throughout the video and 6 sampled uniformly from the second half of the episode. We did this since the ending of a trajectory often has more task success relevant information. We then calculate Cohen's kappa $\kappa$ to measures the agreement between each of the VLM raters (Fig. \ref{fig:vlm-kappa}), and find that  InternVL3.5-14B and Qwen3.6-27B are in moderate agreement. Thus we chose the consensus rating from these two VLMs for our success rate figures. We also verify that the main trend is supported by non-VLM metrics: CEM, IDM, success probes, and visual/geometric metrics.

\section{Additional Results}
\label{app:additional-res}
\subsection{Visual performance across DiT backbone sizes}

\begin{figure}[t]
    \centering
\includegraphics[width=\textwidth]{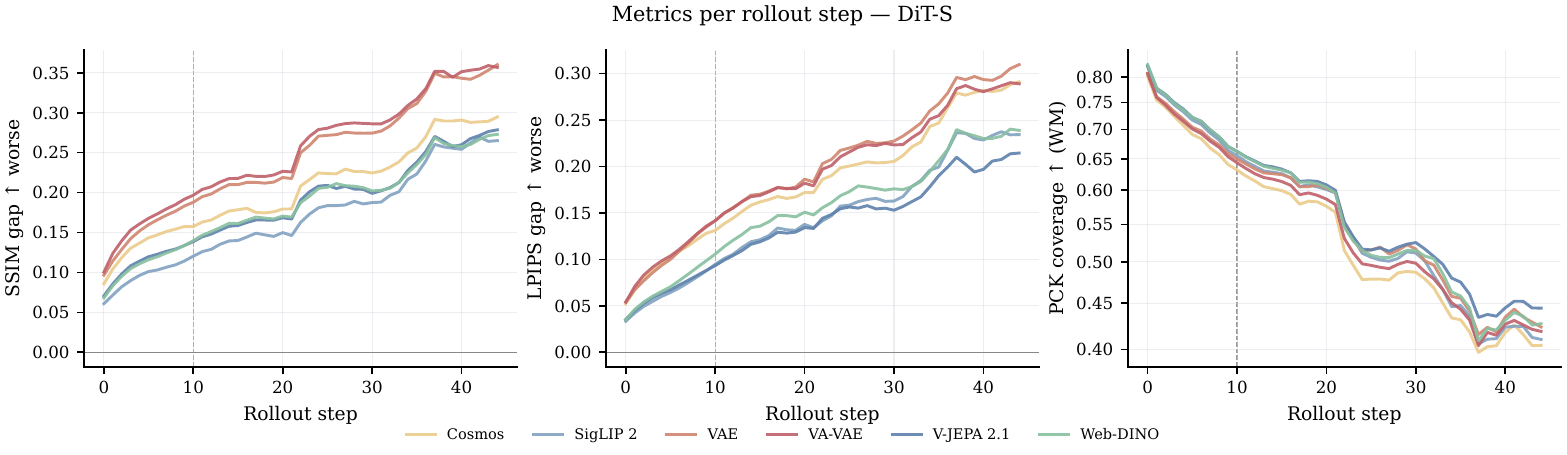}
    \caption{\textbf{SSIM gap, LPIPS gap, and PCK coverage} over 45 rollout steps. While all encoders show a strictly increasing SSIM/LPIPS gap over the full rollout due to compounding errors (each autoregressive step feeds back slightly corrupted predictions as context), semantic latent spaces from SigLIP2, V-JEPA 2.1 and Web-DINO remain particularly competitive when  forced to extrapolate beyond the 10-frame horizon length seen during training. Conversely, PCK coverage remains the highest for semantic encoders.}
    \label{fig:rollout-gap-full}
\end{figure}

\begin{table}[H]
\centering
\caption{\textbf{Reconstruction quality across DiT sizes: S, B, and L.} Each cell for PSNR, SSIM, LPIPS, t-LPIPS, FID, and FVD shows the WM value with the gap to its encoder's reconstruction ceiling in parentheses (lower is closer to the ceiling). \best{Best} and \second{runner-up} within each size group; the WM value and gap are highlighted independently.}
\label{tab:reconstruction-quality}
\resizebox{\textwidth}{!}{%
\begin{tabular}{l|ccccc|cc}
\toprule
  & \multicolumn{5}{c|}{Reconstruction fidelity} & \multicolumn{2}{c}{Generative quality} \\
\cmidrule(lr){2-6}\cmidrule(lr){7-8}
Encoder & PSNR & SSIM & LPIPS & \shortstack{PCK\\coverage} & t-LPIPS & FID & FVD \\
  & $\uparrow$ & $\uparrow$ & $\downarrow$ & $\uparrow$ & $\downarrow$ & $\downarrow$ & $\downarrow$ \\
\midrule
\multicolumn{8}{c}{DiT-S} \\
\midrule
\famtag{VAE}{$\bullet$}~VAE & 17.43~\gap{16.65} & 0.688~\gap{0.251} & 0.218~\gap{0.207} & 0.565 & 0.0264~\gap{0.0252} & 17.43~\gap{16.63} & 6.8~\gap{6.5} \\
\rowcolor{wmGroupBg}\famtag{VAVAE}{$\bullet$}~VA-VAE & 16.98~\gap{13.24} & 0.633~\gap{0.256} & 0.226~\gap{0.201} & 0.559 & 0.0253~\gap{0.0228} & 15.49~\gap{13.24} & 8.5~\gap{7.0} \\
\famtag{COSMOS}{$\bullet$}~Cosmos & 16.97~\gap{10.06} & 0.608~\gap{0.210} & 0.245~\gap{0.197} & 0.544 & 0.0223~\gap{0.0184} & 16.95~\gap{11.37} & 8.2~\gap{4.5} \\
% \cdashline{1-8}[1pt/2pt]
\cmidrule{1-8}
\rowcolor{wmGroupBg}\famtag{VJEPA2}{$\bullet$}~V-JEPA 2.1 & 18.10~\gap{11.09} & 0.725~\gap{0.190} & \best{0.176}~\gap{0.141} & \best{0.580} & 0.0197~\gap{0.0161} & 6.77~\gap{2.75} & \second{5.5}~\gap{2.3} \\
\famtag{VJEPA2}{$\bullet$}~V-JEPA 2.1$_{96}$ & \second{18.20}~\gap{10.99} & 0.729~\gap{0.186} & \second{0.179}~\gap{0.143} & 0.575 & 0.0212~\gap{0.0176} & \second{6.30}~\gap{2.27} & \best{5.2}~\gap{2.1} \\
\rowcolor{wmGroupBg}\famtag{WEBSSL}{$\bullet$}~Web-DINO & 17.42~\gap{10.87} & 0.722~\gap{0.188} & 0.199~\gap{0.160} & 0.575 & 0.0234~\gap{0.0190} & 7.63~\gap{3.37} & 6.7~\gap{3.1} \\
\famtag{WEBSSL}{$\bullet$}~Web-DINO$_{16}$ & 17.82~\gap{\best{8.54}} & 0.711~\gap{0.171} & 0.196~\gap{0.139} & 0.563 & 0.0200~\gap{\second{0.0144}} & 8.37~\gap{2.55} & 7.6~\gap{2.4} \\
\rowcolor{wmGroupBg}\famtag{WEBSSL}{$\bullet$}~Web-DINO$_{64}$ & \best{18.26}~\gap{9.54} & \second{0.738}~\gap{\second{0.168}} & 0.196~\gap{\best{0.124}} & 0.575 & \best{0.0185}~\gap{\best{0.0118}} & 14.18~\gap{\best{1.39}} & 10.9~\gap{\best{1.5}} \\
\famtag{WEBSSL}{$\bullet$}~Web-DINO$_{96}$ & 17.99~\gap{10.26} & 0.728~\gap{0.180} & 0.181~\gap{0.142} & 0.572 & \second{0.0195}~\gap{0.0151} & \best{6.00}~\gap{\second{2.16}} & 5.5~\gap{2.2} \\
\rowcolor{wmGroupBg}\famtag{WEBSSL}{$\bullet$}~Web-DINO$_{256}$ & 17.63~\gap{10.74} & 0.725~\gap{0.187} & 0.214~\gap{0.148} & 0.574 & 0.0231~\gap{0.0170} & 14.25~\gap{2.16} & 10.4~\gap{2.1} \\
\famtag{SIGLIP}{$\bullet$}~SigLIP 2 & 17.48~\gap{9.29} & 0.713~\gap{0.181} & 0.205~\gap{0.156} & 0.555 & 0.0228~\gap{0.0173} & 7.86~\gap{2.89} & 6.9~\gap{2.4} \\
\rowcolor{wmGroupBg}\famtag{SIGLIP}{$\bullet$}~SigLIP 2$_{96}$ & 18.06~\gap{\second{8.69}} & \best{0.738}~\gap{\best{0.152}} & 0.179~\gap{\second{0.131}} & \second{0.578} & 0.0223~\gap{0.0168} & 6.88~\gap{2.40} & 6.0~\gap{\second{1.8}} \\
\midrule
\multicolumn{8}{c}{DiT-B} \\
\midrule
\famtag{VAE}{$\bullet$}~VAE & 17.47~\gap{16.60} & 0.682~\gap{0.257} & 0.206~\gap{0.196} & 0.565 & 0.0236~\gap{0.0226} & 10.64~\gap{9.84} & \second{5.1}~\gap{4.7} \\
% \cdashline{1-8}[1pt/2pt]
\cmidrule{1-8}
\rowcolor{wmGroupBg}\famtag{VJEPA2}{$\bullet$}~V-JEPA 2.1 & \best{18.43}~\gap{\second{10.81}} & \best{0.740}~\gap{\best{0.178}} & \best{0.171}~\gap{\best{0.135}} & \best{0.584} & \best{0.0205}~\gap{\best{0.0169}} & 7.31~\gap{2.81} & 5.9~\gap{\second{2.5}} \\
\famtag{VJEPA2}{$\bullet$}~V-JEPA 2.1$_{96}$ & \second{18.06}~\gap{11.07} & \second{0.726}~\gap{0.186} & \second{0.185}~\gap{\second{0.150}} & \second{0.573} & \second{0.0206}~\gap{\second{0.0171}} & \second{5.99}~\gap{\best{2.40}} & \best{5.0}~\gap{\best{2.1}} \\
\rowcolor{wmGroupBg}\famtag{WEBSSL}{$\bullet$}~Web-DINO & 17.76~\gap{\best{10.40}} & 0.716~\gap{\second{0.185}} & 0.190~\gap{0.151} & 0.571 & 0.0233~\gap{0.0189} & \best{5.96}~\gap{\second{2.81}} & 5.5~\gap{2.7} \\
\midrule
\multicolumn{8}{c}{DiT-L} \\
\midrule
\famtag{VAE}{$\bullet$}~VAE & 18.44~\gap{15.63} & 0.729~\gap{0.210} & \second{0.168}~\gap{0.157} & 0.575 & 0.0202~\gap{0.0191} & \best{5.35}~\gap{4.56} & \best{3.5}~\gap{3.1} \\
\rowcolor{wmGroupBg}\famtag{COSMOS}{$\bullet$}~Cosmos & 18.01~\gap{9.02} & 0.657~\gap{0.162} & 0.186~\gap{0.138} & 0.563 & 0.0199~\gap{0.0159} & 9.23~\gap{3.65} & 6.5~\gap{2.9} \\
% \cdashline{1-8}[1pt/2pt]
\cmidrule{1-8}
\famtag{VJEPA2}{$\bullet$}~V-JEPA 2.1 & 18.53~\gap{10.70} & 0.741~\gap{0.177} & 0.172~\gap{0.136} & \second{0.583} & 0.0195~\gap{0.0159} & 6.94~\gap{2.44} & 5.4~\gap{1.9} \\
\rowcolor{wmGroupBg}\famtag{VJEPA2}{$\bullet$}~V-JEPA 2.1$_{96}$ & \best{18.65}~\gap{10.54} & \best{0.743}~\gap{0.171} & \best{0.165}~\gap{0.130} & \best{0.584} & 0.0201~\gap{0.0166} & \second{6.19}~\gap{\second{2.16}} & \second{5.2}~\gap{2.1} \\
\famtag{WEBSSL}{$\bullet$}~Web-DINO & 17.72~\gap{10.56} & 0.729~\gap{0.181} & 0.192~\gap{0.153} & 0.581 & 0.0219~\gap{0.0176} & 6.92~\gap{2.65} & 6.0~\gap{2.5} \\
\rowcolor{wmGroupBg}\famtag{WEBSSL}{$\bullet$}~Web-DINO$_{96}$ & \second{18.62}~\gap{9.62} & 0.741~\gap{\second{0.154}} & 0.189~\gap{\best{0.117}} & 0.577 & \best{0.0189}~\gap{\best{0.0123}} & 14.26~\gap{\best{1.33}} & 13.1~\gap{\best{1.3}} \\
\famtag{SIGLIP}{$\bullet$}~SigLIP 2 & 17.94~\gap{\second{8.83}} & 0.730~\gap{0.163} & 0.188~\gap{0.140} & 0.581 & 0.0207~\gap{0.0153} & 7.57~\gap{2.60} & 6.7~\gap{2.2} \\
\rowcolor{wmGroupBg}\famtag{SIGLIP}{$\bullet$}~SigLIP 2$_{96}$ & 18.30~\gap{\best{8.45}} & \second{0.743}~\gap{\best{0.147}} & 0.171~\gap{\second{0.123}} & 0.580 & \second{0.0193}~\gap{\second{0.0138}} & 6.74~\gap{2.24} & 5.8~\gap{\second{1.6}} \\
\bottomrule
\end{tabular}
}
\end{table}

\begin{table}[H]
\centering
\caption{\textbf{World Arena perceptual metrics across DiT sizes: S, B, and L.} \best{Best} and \second{runner-up} within each size group.}
\label{tab:world-arena-full}
\resizebox{\textwidth}{!}{%
\begin{tabular}{l|cc|ccc|cc|cc}
\toprule
  & \multicolumn{2}{c|}{Quality} & \multicolumn{3}{c|}{Frame consistency} & \multicolumn{2}{c|}{Motion} & \multicolumn{2}{c}{Reference-based} \\
\cmidrule(lr){2-3}\cmidrule(lr){4-6}\cmidrule(lr){7-8}\cmidrule(lr){9-10}
Encoder & \shortstack{Image\\quality} & \shortstack{Aesthetic\\quality} & \shortstack{Subject\\consist.} & \shortstack{Background\\consist.} & \shortstack{Photometric\\consist.} & \shortstack{Dyn.\\degree} & \shortstack{Flow\\score} & \shortstack{Depth\\AbsRel} & \shortstack{JEPA\\sim.} \\
  & $\uparrow$ & $\uparrow$ & $\uparrow$ & $\uparrow$ & $\downarrow$ & $\uparrow$ & $\uparrow$ & $\downarrow$ & $\uparrow$ \\
\midrule
\multicolumn{10}{c}{DiT-S} \\
\midrule
\famtag{VAE}{$\bullet$}~VAE & \best{0.592} & 0.467 & 0.810 & 0.950 & 96.26 & 0.767 & 1.186 & 0.390 & 0.871 \\
\rowcolor{wmGroupBg}\famtag{VAVAE}{$\bullet$}~VA-VAE & \second{0.585} & 0.464 & 0.817 & 0.949 & 94.93 & 0.765 & 1.204 & 0.455 & 0.783 \\
\famtag{COSMOS}{$\bullet$}~Cosmos & 0.558 & 0.463 & 0.793 & 0.946 & \best{73.29} & 0.813 & 1.511 & 0.638 & 0.517 \\
% \cdashline{1-10}[1pt/2pt]
\cmidrule{1-10}
\rowcolor{wmGroupBg}\famtag{VJEPA2}{$\bullet$}~V-JEPA 2.1 & 0.578 & \second{0.473} & 0.841 & 0.955 & 80.49 & 0.832 & 1.587 & 0.404 & 0.929 \\
\famtag{VJEPA2}{$\bullet$}~V-JEPA 2.1$_{96}$ & 0.579 & \best{0.474} & 0.841 & 0.955 & 76.30 & \best{0.843} & \best{1.653} & 0.363 & 0.928 \\
\rowcolor{wmGroupBg}\famtag{WEBSSL}{$\bullet$}~Web-DINO & 0.576 & 0.472 & 0.849 & 0.957 & 94.03 & 0.794 & 1.408 & \best{0.350} & \second{0.938} \\
\famtag{WEBSSL}{$\bullet$}~Web-DINO$_{16}$ & 0.546 & 0.469 & 0.838 & 0.952 & 76.85 & 0.824 & 1.532 & 0.399 & 0.905 \\
\rowcolor{wmGroupBg}\famtag{WEBSSL}{$\bullet$}~Web-DINO$_{64}$ & 0.575 & 0.466 & \second{0.854} & \second{0.960} & 84.95 & 0.823 & 1.532 & 0.358 & 0.774 \\
\famtag{WEBSSL}{$\bullet$}~Web-DINO$_{96}$ & 0.574 & 0.473 & 0.841 & 0.955 & 76.82 & \second{0.835} & \second{1.634} & 0.375 & \best{0.944} \\
\rowcolor{wmGroupBg}\famtag{WEBSSL}{$\bullet$}~Web-DINO$_{256}$ & 0.581 & 0.467 & \best{0.861} & \best{0.961} & 103.86 & 0.782 & 1.325 & \second{0.357} & 0.785 \\
\famtag{SIGLIP}{$\bullet$}~SigLIP 2 & 0.566 & 0.471 & 0.839 & 0.953 & \second{74.93} & 0.827 & 1.602 & 0.394 & 0.931 \\
\rowcolor{wmGroupBg}\famtag{SIGLIP}{$\bullet$}~SigLIP 2$_{96}$ & 0.573 & 0.472 & 0.843 & 0.955 & 77.30 & 0.827 & 1.547 & 0.372 & 0.938 \\
\midrule
\multicolumn{10}{c}{DiT-B} \\
\midrule
\famtag{VAE}{$\bullet$}~VAE & \best{0.591} & 0.471 & 0.813 & 0.951 & \best{79.77} & \best{0.824} & \second{1.520} & 0.434 & 0.915 \\
% \cdashline{1-10}[1pt/2pt]
\cmidrule{1-10}
\rowcolor{wmGroupBg}\famtag{VJEPA2}{$\bullet$}~V-JEPA 2.1 & \second{0.582} & \best{0.474} & \second{0.847} & \best{0.958} & 87.57 & 0.812 & 1.454 & \best{0.324} & 0.923 \\
\famtag{VJEPA2}{$\bullet$}~V-JEPA 2.1$_{96}$ & 0.577 & \second{0.474} & 0.845 & 0.957 & \second{82.82} & \second{0.823} & \best{1.521} & 0.381 & \second{0.928} \\
\rowcolor{wmGroupBg}\famtag{WEBSSL}{$\bullet$}~Web-DINO & 0.577 & 0.473 & \best{0.847} & \second{0.957} & 86.57 & 0.815 & 1.493 & \second{0.342} & \best{0.939} \\
\midrule
\multicolumn{10}{c}{DiT-L} \\
\midrule
\famtag{VAE}{$\bullet$}~VAE & \best{0.598} & \best{0.475} & 0.827 & 0.952 & \second{75.16} & \best{0.844} & \second{1.635} & \best{0.281} & \best{0.980} \\
\rowcolor{wmGroupBg}\famtag{COSMOS}{$\bullet$}~Cosmos & 0.578 & 0.469 & 0.817 & 0.952 & \best{71.22} & \second{0.843} & \best{1.650} & 0.465 & 0.760 \\
% \cdashline{1-10}[1pt/2pt]
\cmidrule{1-10}
\famtag{VJEPA2}{$\bullet$}~V-JEPA 2.1 & 0.578 & 0.474 & 0.844 & 0.956 & 80.05 & 0.832 & 1.573 & 0.330 & 0.926 \\
\rowcolor{wmGroupBg}\famtag{VJEPA2}{$\bullet$}~V-JEPA 2.1$_{96}$ & \second{0.581} & \second{0.474} & 0.842 & 0.956 & 81.48 & 0.831 & 1.558 & 0.346 & 0.929 \\
\famtag{WEBSSL}{$\bullet$}~Web-DINO & 0.573 & 0.472 & \second{0.847} & \second{0.957} & 84.57 & 0.823 & 1.557 & 0.343 & \second{0.945} \\
\rowcolor{wmGroupBg}\famtag{WEBSSL}{$\bullet$}~Web-DINO$_{96}$ & 0.578 & 0.466 & \best{0.852} & \best{0.959} & 79.79 & 0.833 & 1.568 & 0.352 & 0.709 \\
\famtag{SIGLIP}{$\bullet$}~SigLIP 2 & 0.569 & 0.472 & 0.845 & 0.956 & 79.16 & 0.822 & 1.562 & 0.344 & 0.937 \\
\rowcolor{wmGroupBg}\famtag{SIGLIP}{$\bullet$}~SigLIP 2$_{96}$ & 0.573 & 0.472 & 0.844 & 0.956 & 76.98 & 0.830 & 1.580 & \second{0.326} & 0.937 \\
\bottomrule
\end{tabular}
}
\end{table}

\subsection{Policy performance across DiT backbone sizes}
\label{app:policy-perf}

\begin{table}[H]
\centering
\caption{ \textbf{Policy and behavioral metrics for different DiT sizes:} small (S), base (B), and large (L). \best{Best} and \second{runner-up} within each size group. In-distribution (ID) SR: InternVL3.5 on the 10 episodes shared with OOD evaluations. OOD SR: InternVL3.5 only. Borda rank (lower~$=$~better) aggregates InternVL3.5-14B, and Qwen3.6-27B rankings.  Muted $\pm$ terms show one standard deviation averaged over episode for SR and CEM metrics.}
\label{tab:policy-full}
\resizebox{\textwidth}{!}{%
\begin{tabular}{l|cc|cc|c|ccc|cc}
\toprule
  & \multicolumn{2}{c|}{VLA SR} & \multicolumn{2}{c|}{Interaction quality} & \multicolumn{1}{c|}{PCK} & \multicolumn{3}{c|}{OOD robustness} & \multicolumn{2}{c}{CEM error} \\
\cmidrule(lr){2-3}\cmidrule(lr){4-5}\cmidrule(lr){6-6}\cmidrule(lr){7-9}\cmidrule(lr){10-11}
Encoder & \shortstack{Cons.\\SR} & \shortstack{Borda\\rank} & \shortstack{IQ\\score} & \shortstack{Instruction\\follow} & \shortstack{PCK\\coverage} & \shortstack{ID\\SR} & \shortstack{OOD SR\\distractor} & \shortstack{OOD SR\\instruction} & k=1 & k=4 \\
  & $\uparrow$ & $\downarrow$ & $\uparrow$ & $\uparrow$ & $\uparrow$ & $\uparrow$ & $\uparrow$ & $\uparrow$ & $\downarrow$ & $\downarrow$ \\
\midrule
\multicolumn{11}{c}{DiT-S} \\
\midrule
\famtag{VAE}{$\bullet$}~VAE & 0.169~\uncertainty{$\pm$\,0.030} & 31 & 3.26 & 3.48 & 0.719 & 0.375~\uncertainty{$\pm$\,0.054} & 0.287~\uncertainty{$\pm$\,0.051} & 0.200~\uncertainty{$\pm$\,0.045} & 0.111~\uncertainty{$\pm$\,0.009} & 0.612~\uncertainty{$\pm$\,0.023} \\
\rowcolor{wmGroupBg}\famtag{VAVAE}{$\bullet$}~VA-VAE & 0.175~\uncertainty{$\pm$\,0.030} & 28 & 3.22 & 3.42 & 0.715 & 0.350~\uncertainty{$\pm$\,0.053} & 0.250~\uncertainty{$\pm$\,0.048} & 0.200~\uncertainty{$\pm$\,0.045} & 0.097~\uncertainty{$\pm$\,0.005} & 0.543~\uncertainty{$\pm$\,0.023} \\
\famtag{COSMOS}{$\bullet$}~Cosmos & 0.244~\uncertainty{$\pm$\,0.034} & 20 & 3.32 & 3.51 & 0.707 & 0.425~\uncertainty{$\pm$\,0.055} & 0.362~\uncertainty{$\pm$\,0.054} & 0.275~\uncertainty{$\pm$\,0.050} & 0.112~\uncertainty{$\pm$\,0.009} & 0.661~\uncertainty{$\pm$\,0.033} \\
% \cdashline{1-11}[1pt/2pt]
\cmidrule{1-11}
\rowcolor{wmGroupBg}\famtag{VJEPA2}{$\bullet$}~V-JEPA 2.1 & \second{0.344}~\uncertainty{$\pm$\,0.038} & \best{7} & 3.43 & 3.78 & \second{0.735} & \second{0.600}~\uncertainty{$\pm$\,0.055} & \second{0.575}~\uncertainty{$\pm$\,0.055} & \best{0.400}~\uncertainty{$\pm$\,0.055} & \second{0.084}~\uncertainty{$\pm$\,0.008} & \best{0.424}~\uncertainty{$\pm$\,0.014} \\
\famtag{VJEPA2}{$\bullet$}~V-JEPA 2.1$_{96}$ & \best{0.362}~\uncertainty{$\pm$\,0.038} & \second{9} & \best{3.52} & \second{3.84} & 0.735 & 0.600~\uncertainty{$\pm$\,0.055} & 0.537~\uncertainty{$\pm$\,0.056} & 0.250~\uncertainty{$\pm$\,0.048} & 0.089~\uncertainty{$\pm$\,0.007} & 0.548~\uncertainty{$\pm$\,0.017} \\
\rowcolor{wmGroupBg}\famtag{WEBSSL}{$\bullet$}~Web-DINO & 0.212~\uncertainty{$\pm$\,0.032} & 26 & 3.34 & 3.58 & 0.735 & 0.550~\uncertainty{$\pm$\,0.056} & 0.512~\uncertainty{$\pm$\,0.056} & 0.250~\uncertainty{$\pm$\,0.048} & 0.090~\uncertainty{$\pm$\,0.007} & \second{0.474}~\uncertainty{$\pm$\,0.026} \\
\famtag{WEBSSL}{$\bullet$}~Web-DINO$_{16}$ & 0.256~\uncertainty{$\pm$\,0.035} & 13 & \second{3.51} & \best{3.85} & 0.721 & 0.500~\uncertainty{$\pm$\,0.056} & 0.500~\uncertainty{$\pm$\,0.056} & 0.300~\uncertainty{$\pm$\,0.051} & 0.104~\uncertainty{$\pm$\,0.008} & 0.555~\uncertainty{$\pm$\,0.020} \\
\rowcolor{wmGroupBg}\famtag{WEBSSL}{$\bullet$}~Web-DINO$_{64}$ & 0.281~\uncertainty{$\pm$\,0.036} & 16 & 3.34 & 3.50 & 0.734 & 0.550~\uncertainty{$\pm$\,0.056} & 0.487~\uncertainty{$\pm$\,0.056} & \second{0.325}~\uncertainty{$\pm$\,0.052} & --- & --- \\
\famtag{WEBSSL}{$\bullet$}~Web-DINO$_{96}$ & 0.300~\uncertainty{$\pm$\,0.036} & 13 & 3.44 & 3.77 & 0.732 & 0.600~\uncertainty{$\pm$\,0.055} & 0.512~\uncertainty{$\pm$\,0.056} & 0.275~\uncertainty{$\pm$\,0.050} & 0.090~\uncertainty{$\pm$\,0.007} & 0.531~\uncertainty{$\pm$\,0.025} \\
\rowcolor{wmGroupBg}\famtag{WEBSSL}{$\bullet$}~Web-DINO$_{256}$ & 0.194~\uncertainty{$\pm$\,0.031} & 21 & 3.31 & 3.56 & \best{0.735} & 0.512~\uncertainty{$\pm$\,0.056} & 0.500~\uncertainty{$\pm$\,0.056} & 0.287~\uncertainty{$\pm$\,0.051} & --- & --- \\
\famtag{SIGLIP}{$\bullet$}~SigLIP 2 & 0.325~\uncertainty{$\pm$\,0.037} & 10 & 3.43 & 3.58 & 0.730 & 0.537~\uncertainty{$\pm$\,0.056} & 0.500~\uncertainty{$\pm$\,0.056} & 0.263~\uncertainty{$\pm$\,0.049} & \best{0.082}~\uncertainty{$\pm$\,0.006} & 0.523~\uncertainty{$\pm$\,0.030} \\
\rowcolor{wmGroupBg}\famtag{SIGLIP}{$\bullet$}~SigLIP 2$_{96}$ & 0.331~\uncertainty{$\pm$\,0.037} & 16 & 3.42 & 3.71 & 0.731 & \best{0.625}~\uncertainty{$\pm$\,0.054} & \best{0.588}~\uncertainty{$\pm$\,0.055} & 0.312~\uncertainty{$\pm$\,0.052} & 0.086~\uncertainty{$\pm$\,0.005} & 0.537~\uncertainty{$\pm$\,0.026} \\
\midrule
\multicolumn{11}{c}{DiT-B} \\
\midrule
\famtag{VAE}{$\bullet$}~VAE & 0.256~\uncertainty{$\pm$\,0.035} & 11 & 3.31 & 3.62 & 0.723 & 0.463~\uncertainty{$\pm$\,0.056} & 0.438~\uncertainty{$\pm$\,0.055} & 0.225~\uncertainty{$\pm$\,0.047} & 0.113~\uncertainty{$\pm$\,0.010} & --- \\
% \cdashline{1-11}[1pt/2pt]
\cmidrule{1-11}
\rowcolor{wmGroupBg}\famtag{VJEPA2}{$\bullet$}~V-JEPA 2.1 & \second{0.319}~\uncertainty{$\pm$\,0.037} & \best{4} & \second{3.51} & \best{3.77} & \best{0.739} & \best{0.625}~\uncertainty{$\pm$\,0.054} & 0.475~\uncertainty{$\pm$\,0.056} & \best{0.325}~\uncertainty{$\pm$\,0.052} & \second{0.096}~\uncertainty{$\pm$\,0.008} & --- \\
\famtag{VJEPA2}{$\bullet$}~V-JEPA 2.1$_{96}$ & \best{0.325}~\uncertainty{$\pm$\,0.037} & \second{6} & \best{3.52} & 3.69 & \second{0.737} & 0.575~\uncertainty{$\pm$\,0.055} & \second{0.525}~\uncertainty{$\pm$\,0.056} & 0.200~\uncertainty{$\pm$\,0.045} & 0.097~\uncertainty{$\pm$\,0.009} & --- \\
\rowcolor{wmGroupBg}\famtag{WEBSSL}{$\bullet$}~Web-DINO & 0.287~\uncertainty{$\pm$\,0.036} & 8 & 3.44 & \second{3.75} & 0.736 & \second{0.600}~\uncertainty{$\pm$\,0.055} & \best{0.550}~\uncertainty{$\pm$\,0.056} & \second{0.300}~\uncertainty{$\pm$\,0.051} & \best{0.084}~\uncertainty{$\pm$\,0.006} & --- \\
\midrule
\multicolumn{11}{c}{DiT-L} \\
\midrule
\famtag{VAE}{$\bullet$}~VAE & 0.350~\uncertainty{$\pm$\,0.038} & 11 & \best{3.60} & \second{3.95} & \second{0.737} & \best{0.688}~\uncertainty{$\pm$\,0.052} & \best{0.675}~\uncertainty{$\pm$\,0.052} & 0.350~\uncertainty{$\pm$\,0.053} & 0.120~\uncertainty{$\pm$\,0.009} & --- \\
\rowcolor{wmGroupBg}\famtag{COSMOS}{$\bullet$}~Cosmos & \best{0.406}~\uncertainty{$\pm$\,0.039} & 9 & 3.57 & \best{4.01} & 0.722 & 0.637~\uncertainty{$\pm$\,0.054} & 0.500~\uncertainty{$\pm$\,0.056} & \best{0.438}~\uncertainty{$\pm$\,0.055} & 0.132~\uncertainty{$\pm$\,0.011} & --- \\
% \cdashline{1-11}[1pt/2pt]
\cmidrule{1-11}
\famtag{VJEPA2}{$\bullet$}~V-JEPA 2.1 & 0.350~\uncertainty{$\pm$\,0.038} & 21 & 3.52 & 3.84 & \best{0.740} & 0.575~\uncertainty{$\pm$\,0.055} & 0.562~\uncertainty{$\pm$\,0.055} & 0.312~\uncertainty{$\pm$\,0.052} & 0.093~\uncertainty{$\pm$\,0.008} & --- \\
\rowcolor{wmGroupBg}\famtag{VJEPA2}{$\bullet$}~V-JEPA 2.1$_{96}$ & \second{0.388}~\uncertainty{$\pm$\,0.039} & \best{8} & 3.44 & 3.80 & 0.737 & \second{0.688}~\uncertainty{$\pm$\,0.052} & 0.550~\uncertainty{$\pm$\,0.056} & 0.287~\uncertainty{$\pm$\,0.051} & 0.106~\uncertainty{$\pm$\,0.008} & --- \\
\famtag{WEBSSL}{$\bullet$}~Web-DINO & 0.325~\uncertainty{$\pm$\,0.037} & 14 & 3.39 & 3.69 & 0.737 & 0.588~\uncertainty{$\pm$\,0.055} & 0.500~\uncertainty{$\pm$\,0.056} & 0.325~\uncertainty{$\pm$\,0.052} & \best{0.087}~\uncertainty{$\pm$\,0.007} & --- \\
\rowcolor{wmGroupBg}\famtag{WEBSSL}{$\bullet$}~Web-DINO$_{96}$ & 0.344~\uncertainty{$\pm$\,0.038} & 14 & \second{3.59} & 3.90 & 0.735 & 0.550~\uncertainty{$\pm$\,0.056} & 0.588~\uncertainty{$\pm$\,0.055} & 0.225~\uncertainty{$\pm$\,0.047} & 0.091~\uncertainty{$\pm$\,0.006} & --- \\
\famtag{SIGLIP}{$\bullet$}~SigLIP 2 & 0.356~\uncertainty{$\pm$\,0.038} & \second{8} & 3.42 & 3.75 & 0.734 & 0.625~\uncertainty{$\pm$\,0.054} & 0.588~\uncertainty{$\pm$\,0.055} & 0.300~\uncertainty{$\pm$\,0.051} & \second{0.088}~\uncertainty{$\pm$\,0.007} & --- \\
\rowcolor{wmGroupBg}\famtag{SIGLIP}{$\bullet$}~SigLIP 2$_{96}$ & 0.381~\uncertainty{$\pm$\,0.038} & 11 & 3.43 & 3.75 & 0.733 & 0.575~\uncertainty{$\pm$\,0.055} & \second{0.613}~\uncertainty{$\pm$\,0.054} & \second{0.388}~\uncertainty{$\pm$\,0.054} & 0.092~\uncertainty{$\pm$\,0.007} & --- \\
\bottomrule
\end{tabular}
}
\end{table}

\begin{table}[H]
\centering
\caption{\textbf{Per-instruction VLA task success for DiT-L encoders} (each cell: successes out of 8 trials). Columns: \textcolor{cVAE}{$\bullet$}~VAE,\; \textcolor{cCOSMOS}{$\bullet$}~Cosmos,\; \textcolor{cVJEPA2}{$\bullet$}~V-JEPA\,2.1$_{96}$,\; \textcolor{cWEBSSL}{$\bullet$}~Web-DINO$_{96}$,\; \textcolor{cSIGLIP}{$\bullet$}~SigLIP\,2$_{96}$. Thin gray rule separates reconstruction encoders from semantic encoders within each rater group. Instructions are ranked into four difficulty levels for a tabletop robotic arm: L1\ basic single-step actions; L2\ pick-and-place with relative/area positioning; L3\ precise placement, specific orientation, or force control; L4\ deformable-object manipulation and structural stacking. \best{Bold}: highest count per instruction and rater (zero-only rows not highlighted).}
\label{tab:vla-per-instruction-ditl}
\resizebox{\textwidth}{!}{%
\begin{tabular}{p{9cm}cc!{\color{gray!40}\vline}ccc|cc!{\color{gray!40}\vline}ccc}
\toprule
 & \multicolumn{5}{c|}{Qwen\,3.6} & \multicolumn{5}{c}{InternVL\,3.5} \\
\cmidrule(lr){2-6}\cmidrule(lr){7-11}
Instruction & \textcolor{cVAE}{$\bullet$} & \textcolor{cCOSMOS}{$\bullet$} & \textcolor{cVJEPA2}{$\bullet$} & \textcolor{cWEBSSL}{$\bullet$} & \textcolor{cSIGLIP}{$\bullet$} & \textcolor{cVAE}{$\bullet$} & \textcolor{cCOSMOS}{$\bullet$} & \textcolor{cVJEPA2}{$\bullet$} & \textcolor{cWEBSSL}{$\bullet$} & \textcolor{cSIGLIP}{$\bullet$} \\
\midrule
\multicolumn{11}{c}{\small\textit{Level 1 --- Basic single-step actions (open, close, sweep)}} \\
\midrule
close oven & 6 & 6 & 7 & \best{8} & 6 & \best{8} & \best{8} & \best{8} & \best{8} & \best{8} \\
\rowcolor{wmGroupBg}open the drawer & \best{8} & \best{8} & \best{8} & \best{8} & \best{8} & \best{8} & \best{8} & \best{8} & \best{8} & \best{8} \\
pick up sponge and wipe plate & \best{8} & 7 & \best{8} & \best{8} & \best{8} & \best{8} & \best{8} & 7 & 5 & \best{8} \\
\rowcolor{wmGroupBg}sweep into pile & 7 & 7 & 7 & \best{8} & 6 & \best{8} & \best{8} & 7 & 4 & 5 \\
\arrayrulecolor{gray!50}\midrule\arrayrulecolor{black}
\rowcolor{wmHeaderBg}\textit{mean} & 7.2 & 7.0 & 7.5 & \best{8.0} & 7.0 & \best{8.0} & \best{8.0} & 7.5 & 6.2 & 7.2 \\
\midrule
\multicolumn{11}{c}{\small\textit{Level 2 --- Pick-and-place with relative or area-level positioning}} \\
\midrule
Move the can behind the blue fork & 2 & 3 & 2 & 6 & \best{7} & \best{7} & 4 & \best{7} & \best{7} & 6 \\
\rowcolor{wmGroupBg}Move the red spoon to the left of the pot & \best{7} & 6 & 5 & 6 & \best{7} & 6 & \best{8} & 7 & \best{8} & \best{8} \\
close brown1fbox flap & 1 & \best{2} & \best{2} & 0 & \best{2} & 7 & 4 & \best{8} & 2 & 6 \\
\rowcolor{wmGroupBg}moved the blue scrubber onto the lower right burner & 4 & 5 & 5 & \best{6} & 3 & 2 & 3 & \best{4} & \best{4} & 2 \\
pick up the green object above the drawer and place it on the table & 0 & 0 & 1 & \best{2} & 1 & 1 & 1 & 1 & \best{3} & 1 \\
\rowcolor{wmGroupBg}place the silver pot in the middle of the table & 1 & \best{4} & \best{4} & 3 & 0 & \best{2} & 1 & \best{2} & 0 & 0 \\
put banana in pot or pan & 5 & \best{6} & 5 & 3 & 5 & \best{4} & \best{4} & 1 & 0 & 0 \\
\arrayrulecolor{gray!50}\midrule\arrayrulecolor{black}
\rowcolor{wmHeaderBg}\textit{mean} & 2.9 & \best{3.7} & 3.4 & \best{3.7} & 3.6 & 4.1 & 3.6 & \best{4.3} & 3.4 & 3.3 \\
\midrule
\multicolumn{11}{c}{\small\textit{Level 3 --- Precise placement, specific orientation, or force control}} \\
\midrule
pick up blue towel from the grey thing and placed it to the right of the white basket & 4 & 5 & 5 & 5 & \best{8} & 6 & 5 & \best{7} & 6 & 6 \\
\rowcolor{wmGroupBg}pour almonds in pot & 4 & 2 & 1 & 1 & \best{8} & 0 & 1 & 0 & 0 & \best{2} \\
put cucumber in cup & 4 & 3 & 4 & \best{7} & 3 & 1 & 1 & \best{3} & \best{3} & \best{3} \\
\rowcolor{wmGroupBg}put the covering lid on top of the silver pot & 3 & \best{4} & 1 & 3 & 3 & 1 & 0 & 0 & \best{2} & 1 \\
turn lever vertical to front & \best{4} & 1 & 2 & 1 & 1 & 0 & 0 & 0 & 0 & 0 \\
\arrayrulecolor{gray!50}\midrule\arrayrulecolor{black}
\rowcolor{wmHeaderBg}\textit{mean} & 3.8 & 3.0 & 2.6 & 3.4 & \best{4.6} & 1.6 & 1.4 & 2.0 & 2.2 & \best{2.4} \\
\midrule
\multicolumn{11}{c}{\small\textit{Level 4 --- Deformable-object manipulation and structural stacking}} \\
\midrule
fold the cloth from the bottom to the top & 7 & \best{8} & 7 & \best{8} & 7 & 5 & \best{6} & 5 & 0 & 2 \\
\rowcolor{wmGroupBg}move the red rectangle from one tower to another & 0 & 0 & 0 & 0 & 0 & 1 & 0 & 0 & \best{2} & 1 \\
put the rectangular block on top of the yellow and blue cubes & 0 & 0 & 0 & 0 & 0 & 0 & 0 & 0 & 0 & 0 \\
\rowcolor{wmGroupBg}unfold the cloth from bottom right to top left & 6 & \best{8} & 7 & 6 & 5 & \best{8} & \best{8} & 7 & 6 & 5 \\
\arrayrulecolor{gray!50}\midrule\arrayrulecolor{black}
\rowcolor{wmHeaderBg}\textit{mean} & 3.2 & \best{4.0} & 3.5 & 3.5 & 3.0 & \best{3.5} & \best{3.5} & 3.0 & 2.0 & 2.0 \\
\bottomrule
\end{tabular}
}
\end{table}

\subsection{Statistical Analyses}
\label{app:stat-analysis}
\begin{table}[H]
\centering
\caption{\textbf{Uncertainty estimates for policy-facing metrics.} Cells show means with 95\% bootstrap confidence intervals. VLA SR uses consensus VLM success; OOD SR pools distractor and instruction shifts; CEM is one-step controllability error. Family-level rows compare semantic encoders against reconstruction encoders. \best{Best} and \second{runner-up} are scoped per column.}

\resizebox{\textwidth}{!}{%
\begin{tabular}{lccc}
\toprule
Encoder & VLA SR$\uparrow$ & OOD SR$\uparrow$ & CEM error$\downarrow$ \\
\midrule
\famtag{VAE}{$\bullet$}~VAE & 0.169~\uncertainty{[0.113, 0.231]} & 0.303~\uncertainty{[0.244, 0.366]} & 0.111~\uncertainty{[0.096, 0.129]} \\
\rowcolor{wmGroupBg}\famtag{VAVAE}{$\bullet$}~VA-VAE & 0.175~\uncertainty{[0.119, 0.237]} & 0.225~\uncertainty{[0.163, 0.294]} & 0.097~\uncertainty{[0.095, 0.120]} \\
\famtag{COSMOS}{$\bullet$}~Cosmos & 0.244~\uncertainty{[0.181, 0.312]} & 0.319~\uncertainty{[0.250, 0.388]} & 0.112~\uncertainty{[0.096, 0.130]} \\
\midrule
\rowcolor{wmGroupBg}\famtag{VJEPA2}{$\bullet$}~V-JEPA 2.1 & \best{0.344}~\uncertainty{[0.269, 0.419]} & \best{0.487}~\uncertainty{[0.412, 0.569]} & \second{0.084}~\uncertainty{[0.070, 0.100]} \\
\famtag{WEBSSL}{$\bullet$}~Web-DINO & 0.212~\uncertainty{[0.150, 0.275]} & \second{0.388}~\uncertainty{[0.319, 0.456]} & 0.090~\uncertainty{[0.078, 0.103]} \\
\rowcolor{wmGroupBg}\famtag{SIGLIP}{$\bullet$}~SigLIP 2 & \second{0.325}~\uncertainty{[0.256, 0.400]} & 0.381~\uncertainty{[0.306, 0.456]} & \best{0.082}~\uncertainty{[0.071, 0.094]} \\
\midrule
\multicolumn{4}{l}{\rowhead{Family-level semantic vs. reconstruction tests}} \\
Semantic $-$ reconstruction & +0.098~\uncertainty{[0.025, 0.177]} & +0.136~\uncertainty{[0.088, 0.184]} & -0.0266~\uncertainty{[-0.0412, -0.0122]} \\
One-sided test & $p=0.0129$ & $p<5{\times}10^{-5}$ & $p=0.00015$ \\
\bottomrule
\end{tabular}
}
\label{tab:dits-bootstrap-policy-cem}
\end{table}

\paragraph{Uncertainty over policy-facing metrics.} The results show the same simple pattern across the policy-facing metrics: semantic latent spaces are better for task-relevant behavior than reconstruction latent spaces. For in-distribution VLA rollouts, semantic encoders exceed reconstruction encoders by 9.8 percentage points, with a 95\% paired bootstrap interval of [2.5, 17.7] points and an exact one-sided sign-flip test of $p=0.0129$ over the 20 shared task episodes. The OOD result is also positive: when pooling distractor and instruction shifts, semantic encoders exceed reconstruction encoders by 13.6 percentage points, with a 95\% bootstrap interval of [8.8, 18.4] points and $p<5{\times}10^{-5}$. For CEM action recovery, lower error is better; semantic encoders reduce one-step controllability error by 0.0266, with a 95\% bootstrap interval of [0.0122, 0.0412] lower error and $p=0.00015$. Thus, the semantic-family advantage is statistically supported for VLA success, OOD success, and CEM action recovery.

\subsection{Latent representation quality}
\label{app:latent-rep}
\begin{table}[H]
\centering
\caption{\textbf{Inverse Dynamics Model action-recovery} (Pearson $r$ averaged over action dimensions) for horizons $k{=}1$ and $k{=}4$. \emph{Real} = on encoded GT latents (the encoder ceiling); \emph{WM} = on world-model rollouts. \best{Best} and \second{runner-up} per column.}
\resizebox{\textwidth}{!}{%
\begin{tabular}{lcccc|cccc|cccc}
\toprule
  & \multicolumn{4}{c|}{DiT-S} & \multicolumn{4}{c|}{DiT-B} & \multicolumn{4}{c}{DiT-L} \\
\cmidrule(lr){2-5}\cmidrule(lr){6-9}\cmidrule(lr){10-13}
  & \multicolumn{2}{c}{$k{=}1$} & \multicolumn{2}{c|}{$k{=}4$}
  & \multicolumn{2}{c}{$k{=}1$} & \multicolumn{2}{c|}{$k{=}4$}
  & \multicolumn{2}{c}{$k{=}1$} & \multicolumn{2}{c}{$k{=}4$} \\
\cmidrule(lr){2-3}\cmidrule(lr){4-5}
\cmidrule(lr){6-7}\cmidrule(lr){8-9}
\cmidrule(lr){10-11}\cmidrule(lr){12-13}
Encoder
& Real$\uparrow$ & WM$\uparrow$ & Real$\uparrow$ & WM$\uparrow$
& Real$\uparrow$ & WM$\uparrow$ & Real$\uparrow$ & WM$\uparrow$
& Real$\uparrow$ & WM$\uparrow$ & Real$\uparrow$ & WM$\uparrow$ \\
\midrule
\famtag{VAE}{$\bullet$}~VAE
& 0.507 & 0.476 & 0.478 & 0.464
& 0.507 & 0.495 & 0.478 & 0.470
& 0.507 & 0.510 & 0.478 & 0.483 \\
\rowcolor{wmGroupBg}\famtag{VAVAE}{$\bullet$}~VA-VAE
& 0.549 & 0.545 & 0.744 & 0.719
& --- & --- & --- & ---
& --- & --- & --- & --- \\
\famtag{COSMOS}{$\bullet$}~Cosmos
& 0.626 & 0.581 & 0.673 & 0.651
& --- & --- & --- & ---
& 0.626 & 0.617 & 0.673 & 0.671 \\
% \groupsep
\cmidrule{1-13}
\rowcolor{wmGroupBg}\famtag{VJEPA2}{$\bullet$}~V-JEPA 2.1
& \best{0.829} & \best{0.781} & \best{0.865} & \best{0.840}
& \best{0.829} & \best{0.779} & \best{0.865} & \best{0.834}
& \best{0.829} & \best{0.797} & \best{0.865} & \best{0.848} \\
\famtag{WEBSSL}{$\bullet$}~Web-DINO
& \second{0.820} & \second{0.729} & \second{0.845} & \second{0.794}
& \second{0.820} & \second{0.778} & \second{0.845} & \second{0.824}
& \second{0.820} & 0.705 & \second{0.845} & \second{0.785} \\
\rowcolor{wmGroupBg}\famtag{SIGLIP}{$\bullet$}~SigLIP 2
& 0.772 & 0.697 & 0.793 & 0.757
& --- & --- & --- & ---
& 0.772 & \second{0.705} & 0.793 & 0.762 \\
\bottomrule
\end{tabular}
}
\label{tab:idm-pearson-k1-k4}
\end{table}
\begin{table}[H]
\centering
\caption{\textbf{Trajectory success-probe accuracy across DiT sizes.} \emph{Enc.\,Acc/AUC} is computed on encoded ground-truth latents (the probe ceiling); per-DiT columns are accuracy on world-model rollouts and the absolute \emph{Drop} from the encoder ceiling (lower is better). \best{Best} and \second{runner-up} per column. Dashed rule separates VAE-like and SSL encoders.}
\resizebox{0.7\textwidth}{!}{%
\begin{tabular}{lcc|cc|cc|cc}
\toprule
  &   &   & \multicolumn{2}{c|}{DiT-S} & \multicolumn{2}{c|}{DiT-B} & \multicolumn{2}{c}{DiT-L} \\
\cmidrule(lr){4-5}\cmidrule(lr){6-7}\cmidrule(lr){8-9}
Encoder & Enc.\,Acc & Enc.\,AUC & Acc & Drop$\downarrow$ & Acc & Drop$\downarrow$ & Acc & Drop$\downarrow$ \\
\midrule
\famtag{VAE}{$\bullet$}~VAE & 0.835 & 0.917 & 0.716 & 0.119 & 0.716 & 0.119 & 0.685 & 0.150 \\
\rowcolor{wmGroupBg}\famtag{VAVAE}{$\bullet$}~VA-VAE & 0.868 & 0.938 & 0.744 & 0.124 & --- & --- & --- & --- \\
\famtag{COSMOS}{$\bullet$}~Cosmos & 0.851 & 0.925 & 0.723 & 0.128 & --- & --- & 0.732 & 0.119 \\
% \groupsep
\cmidrule{1-9}
\rowcolor{wmGroupBg}\famtag{VJEPA2}{$\bullet$}~V-JEPA 2.1 & \second{0.905} & \second{0.963} & \second{0.789} & \second{0.116} & \best{0.791} & \best{0.114} & 0.796 & \second{0.109} \\
\famtag{WEBSSL}{$\bullet$}~Web-DINO & \best{0.906} & \best{0.963} & 0.788 & 0.118 & \second{0.789} & \second{0.117} & \second{0.797} & 0.109 \\
\rowcolor{wmGroupBg}\famtag{SIGLIP}{$\bullet$}~SigLIP 2 & 0.903 & 0.961 & \best{0.823} & \best{0.080} & --- & --- & \best{0.827} & \best{0.076} \\
\bottomrule
\end{tabular}
}

\label{tab:success-probe-full}
\end{table}
\begin{figure}[H]
    \centering
\includegraphics[width=0.85\textwidth]{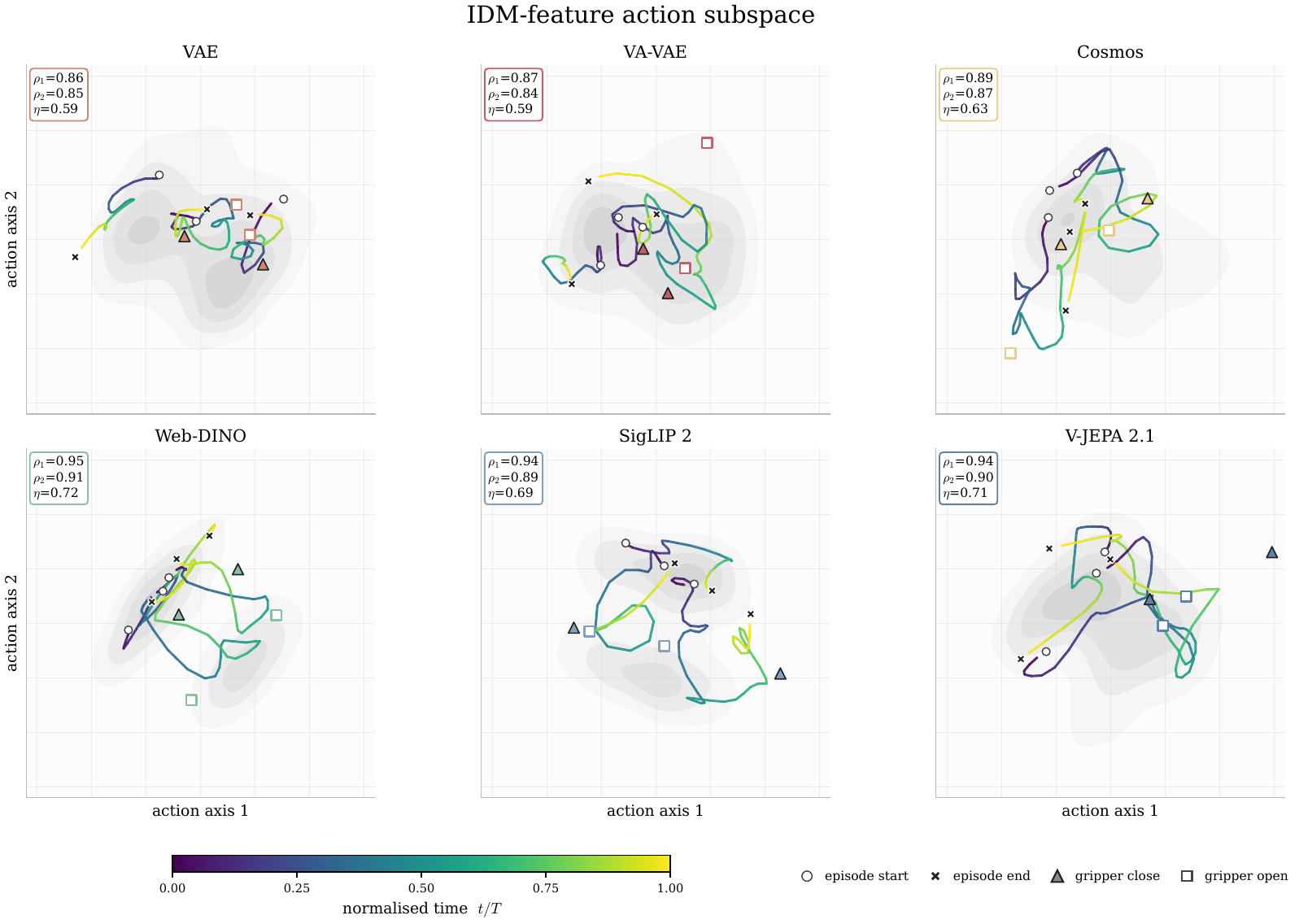}
   \caption{\textbf{Action trajectories induced by encoder spaces:} episode rollouts projected onto the top-2 canonical-correlation directions between IDM features and ground-truth actions. $(\rho_1, \rho_2)$ are the leading canonical correlations, $\eta$ summarizes the aggregate action alignment. Colored curves are episodes.}
    \label{fig:action-traj-full}
    \vspace{-1.7em}
\end{figure}

\subsection{Multi-view transfer learning}
\begin{table}[H]
\centering
\vspace{-1.5em}
\caption{\textbf{DiT-S single-view vs multi-view.} Each cell for PSNR and LPIPS shows the WM value with the gap to its encoder's reconstruction ceiling in parentheses (smaller = closer to ceiling). \best{Best} and \second{runner-up} per column across all rows; the WM value and gap are highlighted independently. The two adapter pairs (V-JEPA~2$_{96}$, Web-DINO$_{96}$) only have multi-view data for CEM. \best{Best}  within each column.}
\label{tab:single-vs-multi-view}
\resizebox{0.7\textwidth}{!}{%
\begin{tabular}{l|cc|cc|c}
\toprule
  & \multicolumn{2}{c|}{Reconstruction fidelity} & \multicolumn{2}{c|}{Generative quality} & \multicolumn{1}{c}{Controllability} \\
\cmidrule(lr){2-3}\cmidrule(lr){4-5}\cmidrule(lr){6-6}
Encoder & PSNR & LPIPS & FID & FVD & \shortstack{CEM\\L2} \\
  & $\uparrow$ & $\downarrow$ & $\downarrow$ & $\downarrow$ & $\downarrow$ \\
\midrule
\famtag{VAE}{$\bullet$}~VAE & 17.43~\gap{16.65} & 0.218~\gap{0.207} & 17.43 & 6.8 & 0.111 \\
\rowcolor{wmGroupBg}\famtag{VAE}{$\bullet$}~VAE~(multi) & 16.67~\gap{18.90} & 0.234~\gap{0.226} & 22.03 & 12.9 & \best{0.047} \\
\cdashline{1-6}[1pt/2pt]
\famtag{COSMOS}{$\bullet$}~Cosmos & 16.97~\gap{\second{10.06}} & 0.245~\gap{0.197} & 16.95 & 8.2 & 0.112 \\
\rowcolor{wmGroupBg}\famtag{COSMOS}{$\bullet$}~Cosmos~(multi) & 16.07~\gap{12.02} & 0.266~\gap{0.223} & 27.65 & 13.8 & \second{0.050} \\
\cdashline{1-6}[1pt/2pt]
\famtag{VJEPA2}{$\bullet$}~V-JEPA 2.1 & \best{18.10}~\gap{11.09} & \best{0.176}~\gap{\best{0.141}} & 6.77 & \second{5.5} & 0.084 \\
\rowcolor{wmGroupBg}\famtag{VJEPA2}{$\bullet$}~V-JEPA 2.1~(multi) & \second{17.50}~\gap{11.10} & 0.186~\gap{0.145} & 9.18 & 6.2 & 0.056 \\
\cdashline{1-6}[1pt/2pt]
\famtag{WEBSSL}{$\bullet$}~Web-DINO & 17.42~\gap{10.87} & 0.199~\gap{0.160} & 7.63 & 6.7 & 0.090 \\
\rowcolor{wmGroupBg}\famtag{WEBSSL}{$\bullet$}~Web-DINO~(multi) & 17.43~\gap{\best{9.77}} & 0.191~\gap{\second{0.141}} & 10.12 & 7.3 & 0.052 \\
\bottomrule
\end{tabular}
}
\end{table}

% --------

\subsection{Effect of adapter dimension}

\begin{wrapfigure}{r}{0.4\textwidth}
  \centering
  \vspace{-2.3em}
  \captionof{table}{\textbf{Adapter dim. $d$ ablation} for Web-DINO DiT-S. \best{Best} and \second{runner-up} highlighted per row.}
  \label{tab:adapter-dim-ablation}

  \resizebox{\linewidth}{!}{%
  \begin{tabular}{lccc}
  \toprule
   & \multicolumn{3}{c}{\famtag{WEBSSL}{$\bullet$}~Web-DINO (DiT-S) latent dim} \\
  \cmidrule(lr){2-4}
  Metric & $d_{16}$ & $d_{96}$ & $D_{1024}$ \\
  \midrule
  VLA SR$\uparrow$ & \second{0.256} & \best{0.269} & 0.181 \\

  \rowcolor{wmGroupBg}SSIM$\uparrow$ & 0.711 & \best{0.728} & \second{0.722} \\
  LPIPS$\downarrow$ & \second{0.196} & \best{0.181} & 0.199 \\

  \rowcolor{wmGroupBg}FID$\downarrow$ & \second{8.37} & \best{6.00} & 7.63 \\
  FVD$\downarrow$ & \second{7.65} & \best{5.51} & 6.66 \\
  \bottomrule
  \end{tabular}
  }
  \vspace{-1.2em}
\end{wrapfigure}
We observe that adapter dimension has a non-monotonic sweet spot. Table~\ref{tab:adapter-dim-ablation} shows that the adapter bottleneck dimension has a non-monotonic effect on performance. For Web-DINO with DiT-S, the intermediate $d_{96}$ setting gives the best overall tradeoff, achieving the highest VLA success rate and the best LPIPS, FID, and FVD. Smaller bottlenecks such as $d_{16}$ remain competitive for policy performance but lose visual quality, while using the full $D_{1024}$ encoder output is worse than the compact $d_{96}$ adapter.

%--------------

\section{Additional Rollouts}
\label{app:rollouts}
\begin{figure}[h]
    \centering
\includegraphics[width=\textwidth]{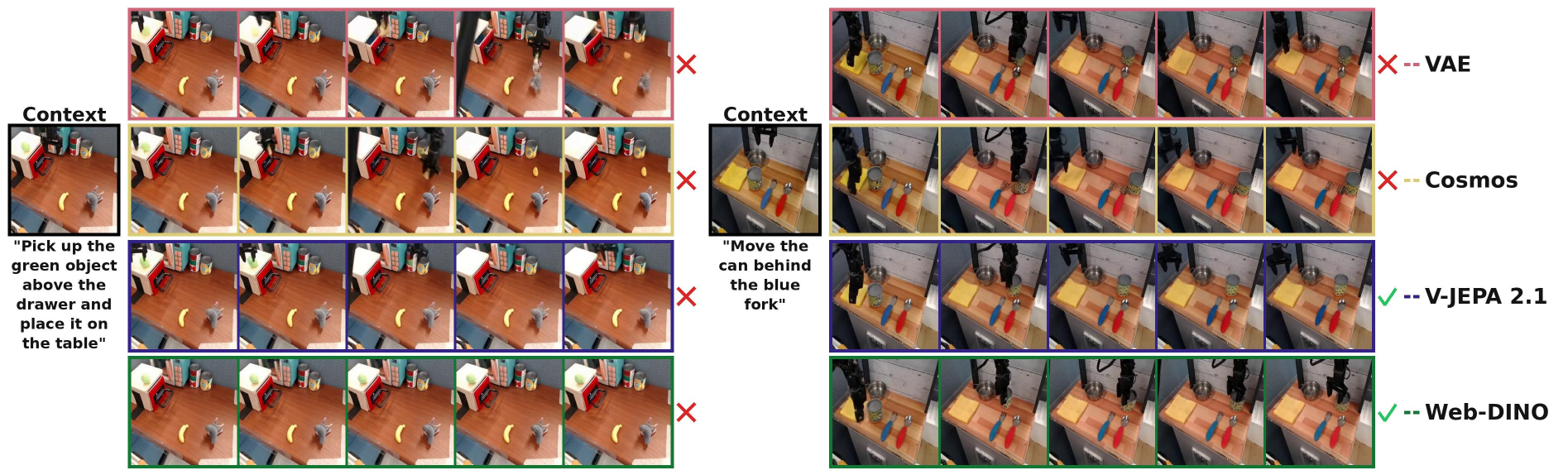}
  \caption{\textbf{Open-VLA success rate comparison on two random episodes:} four frames are sampled at even intervals. \textcolor{green}{\cmark} and \textcolor{red}{\xmark}\;
 show trajectories marked as success and failure by InternVL 3.5 VLM.}
    \label{fig:success-rate-comp-2}
\end{figure}

\begin{figure}[H]
    \centering
\includegraphics[width=\textwidth]{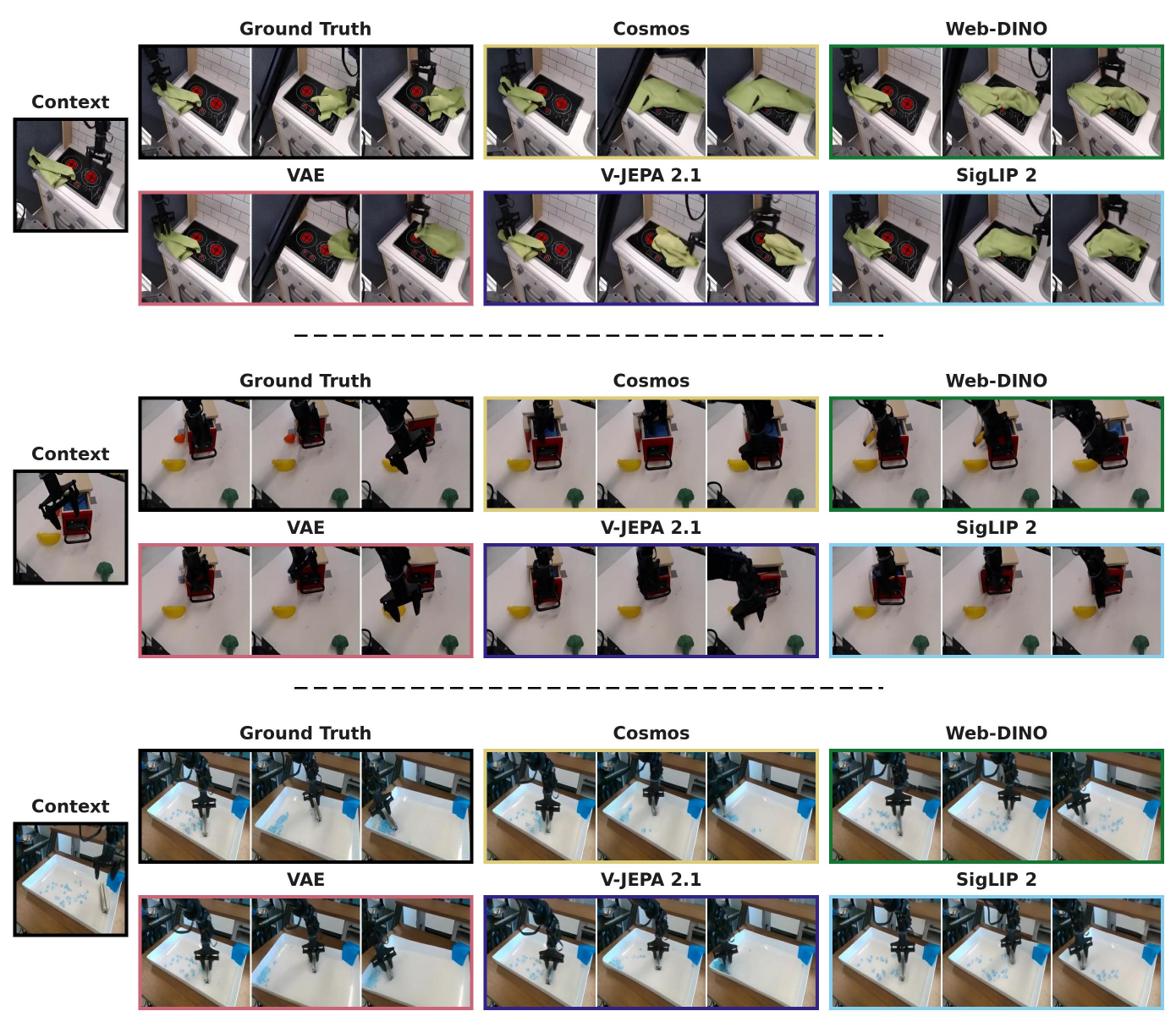}
    \caption{\textbf{Pixel rollout comparison across models on diverse episodes:} the first frame is fed as context and the rest 3 frames are sampled at even intervals from the generated world model rollout.}
    \label{fig:rollout-pixels2}
\end{figure}

\begin{figure}[H]
    \centering
\includegraphics[width=\textwidth]{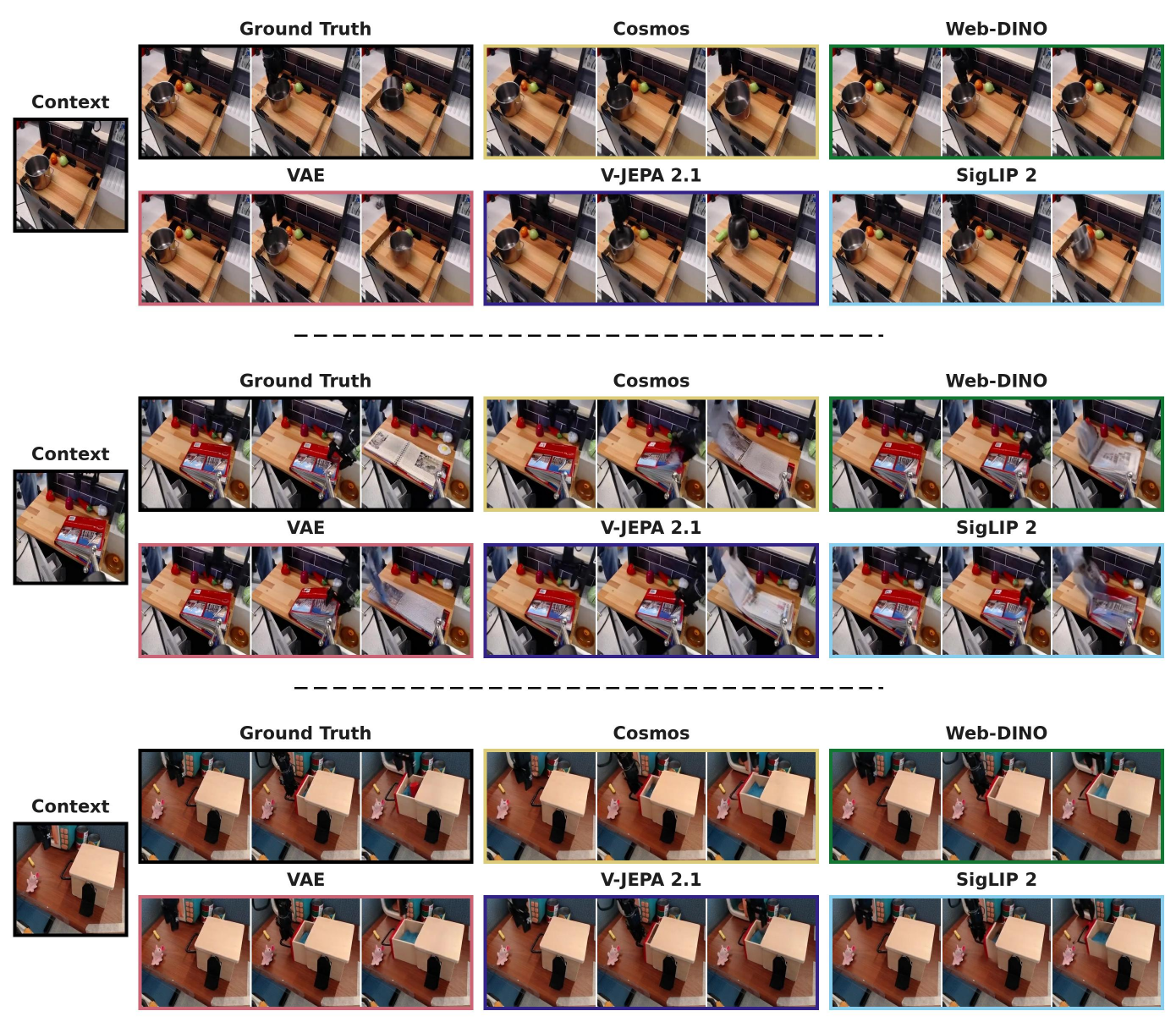}
    \caption{\textbf{Hallucinated pixel rollout comparison across models on diverse episodes:} the first frame is fed as context and the rest 3 frames are sampled at even intervals from the generated world model rollout. \textbf{Top:} flipping the pot consistently causes distortions for all models; \textbf{Middle:}  turning the book pages causes the models to only partially follow the motion  with the book/page appearances becoming smeared and inconsistent, the page edges and cover boundaries drifting; \textbf{Bottom:} while all models predict the appearance of an opening drawer, some clearly under-predict the opening (e.g. VJEPA 2.1) while others show unstable drawer boundary and front panel (e.g. Cosmos).  }
    \label{fig:hallucinated-rollout-pixels2}
\end{figure}

\begin{figure}[H]
    \centering
\includegraphics[width=\textwidth]{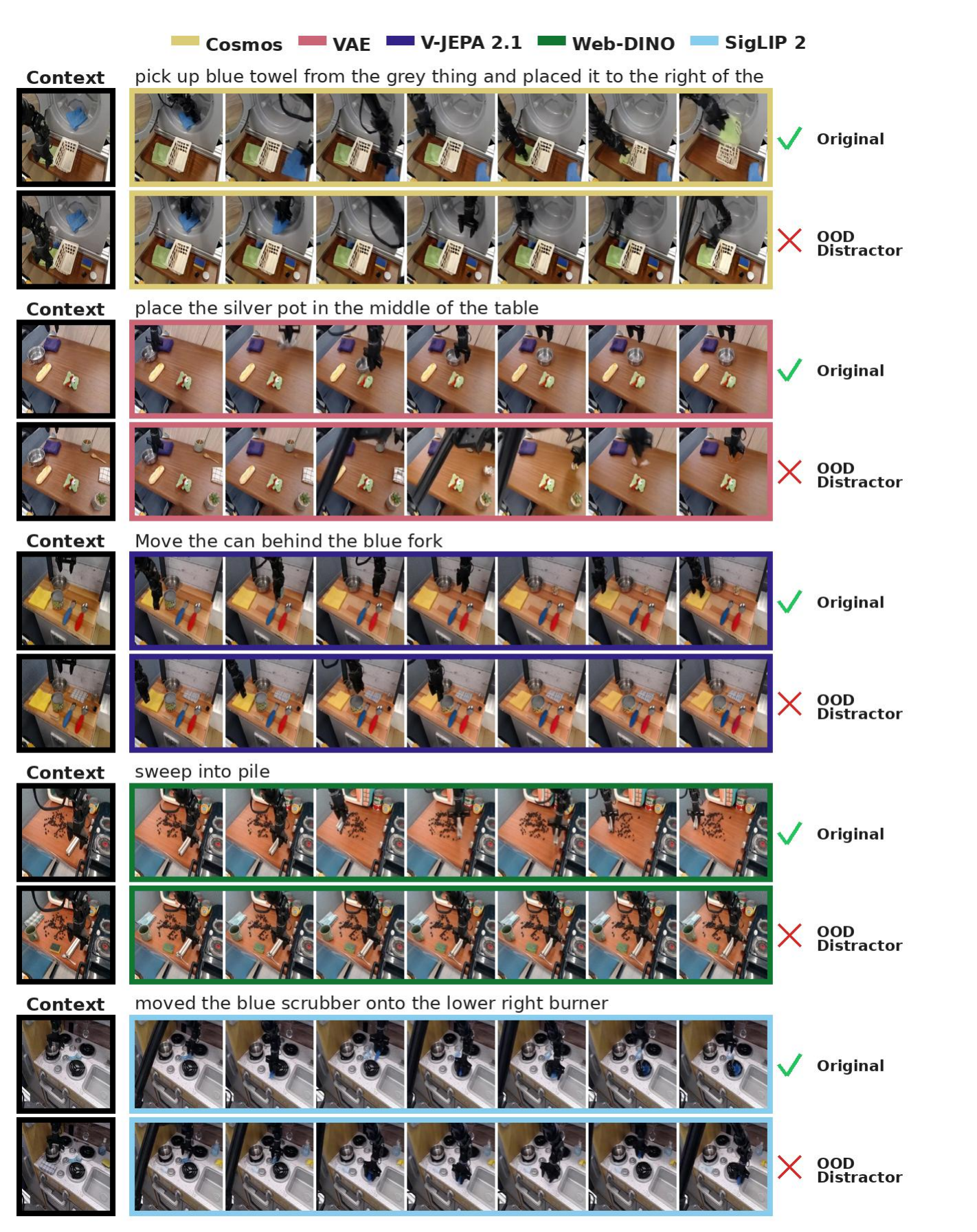}
    \caption{\textbf{OOD Distractor comparison showing failure episodes per model:} OOD objects break task-object binding and action-conditioned state tracking across all models. \textcolor{green}{\cmark} and \textcolor{red}{\xmark}\;
 show trajectories marked as success and failure by InternVL 3.5. In their respective trajectories: Cosmos generates  less stable towel/object state; VAE fails at task-relevant placement of the silver pot; V-JEPA 2.1 loses stable binding between the can, the blue fork, and the instruction, with the can failing to end up reliably behind the fork; Web-DINO fails to maintain the pile-forming interaction; SigLIP 2 keeps the stove layout recognizable, but it does not preserve the precise relation between the scrubber and the target-burner. }
    \label{fig:ood-comparison}
\end{figure}

\begin{figure}[H]
    \centering
\includegraphics[width=\textwidth]{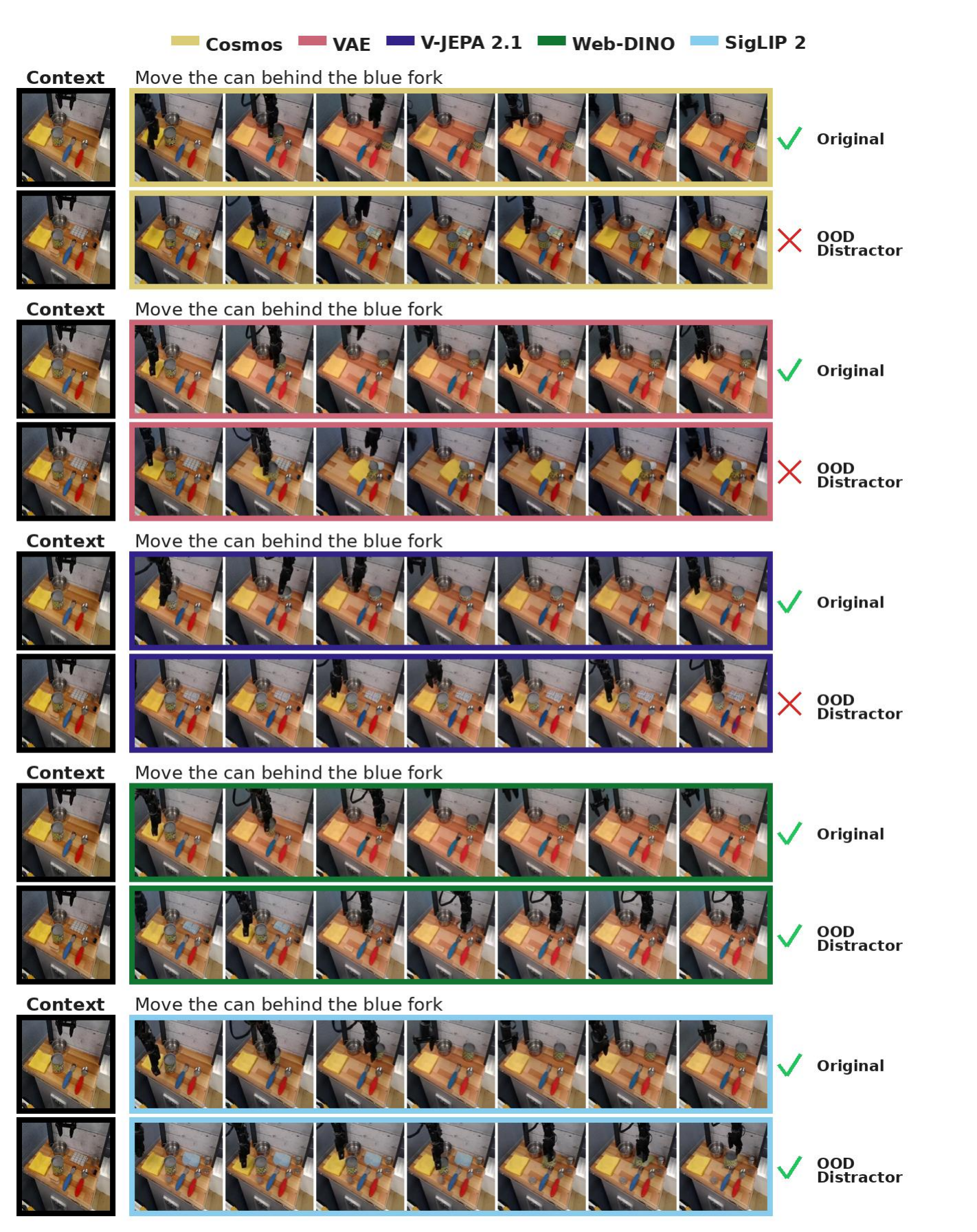}
    \caption{\textbf{OOD Distractor comparison for the same episode:} OOD distractor competes with the target objects and exposes whether a model can keep the target objects bound to the instruction. \textcolor{green}{\cmark} and \textcolor{red}{\xmark}\;
 show trajectories marked as success and failure by InternVL 3.5. Here, irrespective of the task success, the added object visually changes the predicted interaction for all models: the robot/can motion becomes less task-directed, the can’s position is less consistently moved behind the blue fork, and the models appear to let the distractor alter the scene dynamics. }
    \label{fig:ood-comparison-same-ep}
\end{figure}

\begin{figure}[H]
    \centering
\includegraphics[width=\textwidth]{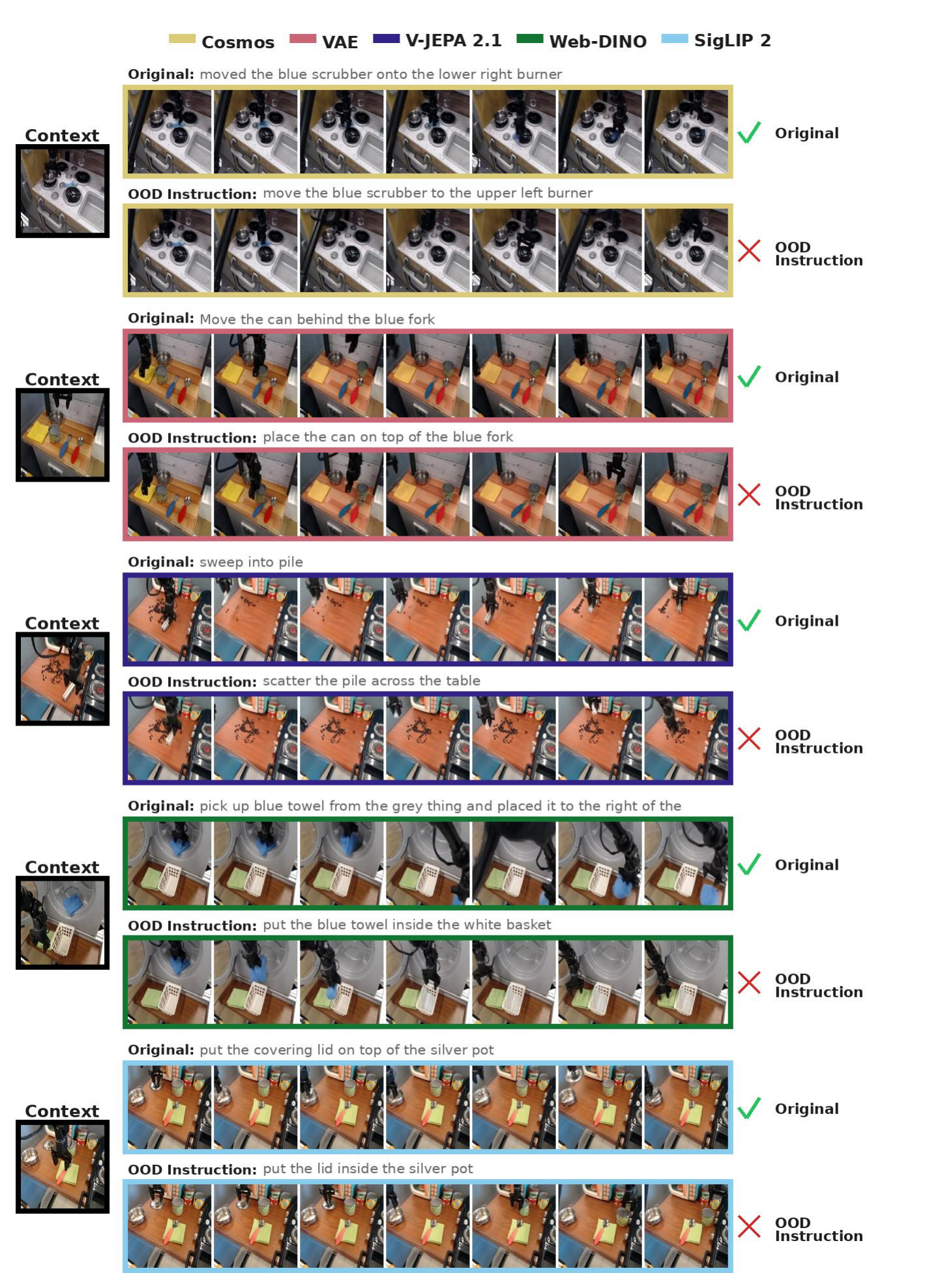}
    \caption{\textbf{OOD Instruction comparison showing failure cases per model:} for each model, the same initial context is rolled out with the original instruction, which succeeds, and then with a OOD instruction, which fails. \textcolor{green}{\cmark} and \textcolor{red}{\xmark}\;
 show trajectories marked as success and failure by InternVL 3.5. Model-specific OOD instruction trajectories show: Cosmos rollout still moves the blue scrubber around the stove, but does not reliably bind it to the new target burner; VAE preserves the table scene, but fails to understand the spatial relation "on top of"; VJEPA 2.1 rollout continues to look like sweeping/piling behavior rather than reversing the task into scattering; Web-DINO keeps the towel manipulation plausible, but misses the new container-based goal; SigLIP 2 rollout shows the lid disappear off the frame. }
    \label{fig:ood-instruction}
\end{figure}

\begin{figure}[H]
    \centering
\includegraphics[width=\textwidth]{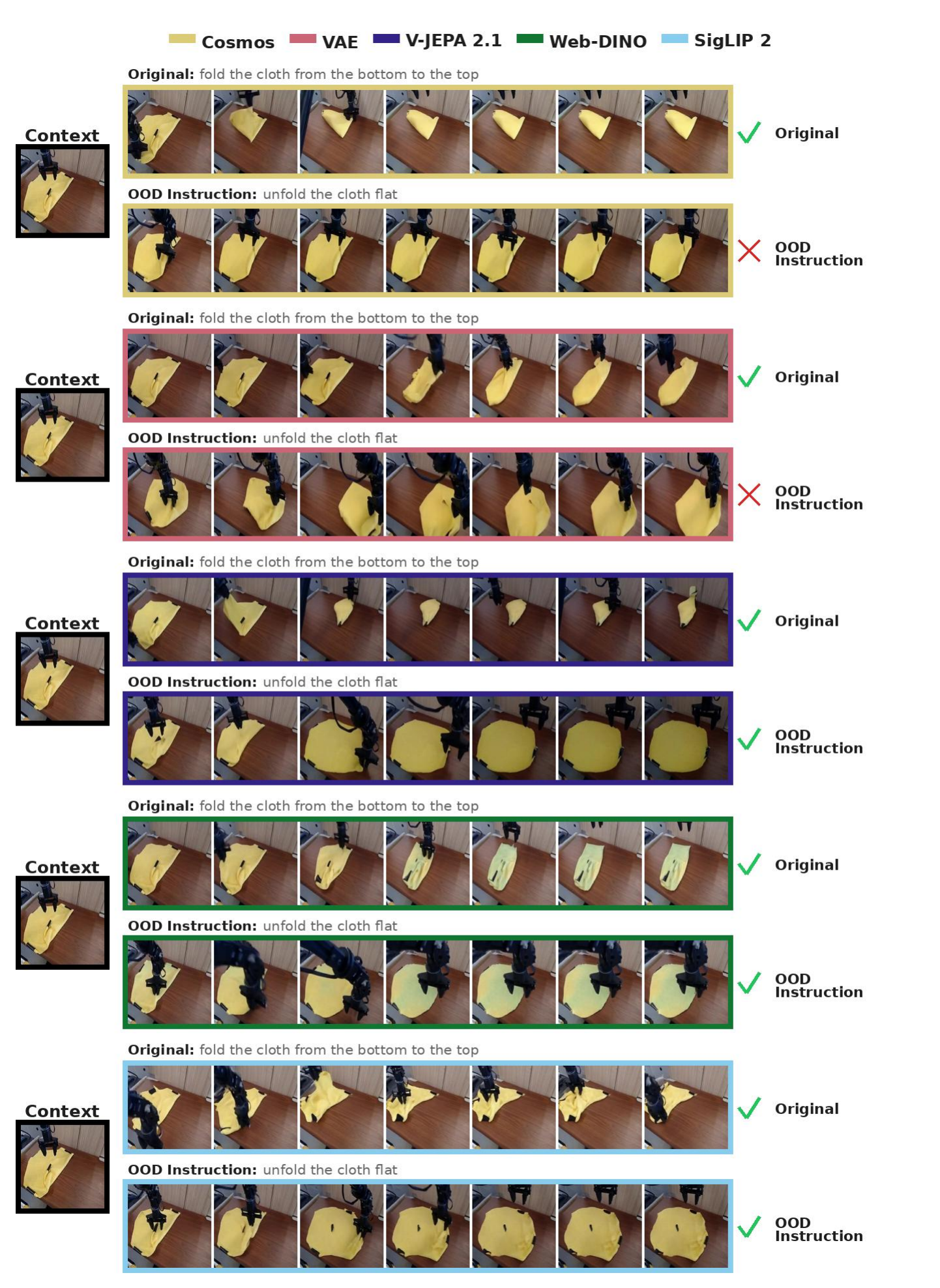}
    \caption{\textbf{OOD Instruction comparison for the same episode:} most models exhibit a common hallucination where  the original object dynamics or defaults to a familiar action pattern instead of updating the final state to match the new instruction. \textcolor{green}{\cmark} and \textcolor{red}{\xmark}\;
 show trajectories marked as success and failure by InternVL 3.5. Both Cosmos and VAE maintain the cloth in a partially folded/creased state instead of flattening it. Semantic encoders more clearly capture the semantic difference between folding and unfolding with VJEPA 2.1 most clearly producing a flatter cloth for the OOD instruction. Web-DINO spreads the cloth, but with some shape distortion and robot occlusion while for SigLIP 2, the cloth shape becomes rounded, suggesting some geometry hallucination despite correct task-level outcome.}
    \label{fig:ood-instruction-same-ep}
\end{figure}

We provide additional rollouts alongside the key observations for Open-VLA success rate comparison (Fig. \ref{fig:success-rate-comp-2}), plain pixel rollouts for comparing differences between standard model outputs (Fig. \ref{fig:rollout-pixels2}) and hallucinated model outputs (Fig. \ref{fig:hallucinated-rollout-pixels2}), rollouts under OOD distractor objects as well as under OOD instructions  for all models across diverse episodes (Fig. \ref{fig:ood-comparison}, \ref{fig:ood-instruction})  as well as on the same episode (Fig. \ref{fig:ood-comparison-same-ep}, \ref{fig:ood-instruction-same-ep}). We also provide sample rollout videos for analyses with the supplementary files. 

\end{document}